\RequirePackage{fix-cm}
\documentclass{svjour3}                     
\smartqed  
\usepackage{graphicx}
\usepackage{bm}			
\usepackage[caption=false]{subfig}
\usepackage{enumitem}
\usepackage[american]{babel}
\usepackage{microtype}
\usepackage{amssymb,amsmath,amstext}
\usepackage{color}
\usepackage{bm}			
\usepackage{url}
\usepackage{algpseudocode,algorithm}
\usepackage{chngpage}	
\usepackage[misc]{ifsym}
\usepackage{multirow}
\usepackage{array}

\begin{document}

\title{Exploiting Color Name Space for Salient Object Detection
}


\author{Jing Lou        \and
        Huan Wang			\and
        Longtao Chen		\and
        Fenglei Xu		\and
        Qingyuan Xia		\and
        Wei Zhu			\and
        Mingwu Ren
}


\institute{J. Lou (\Letter) \at
              School of Information Engineering, Changzhou Vocational Institute of Mechatronic Technology, Changzhou, Jiangsu 213164, P.R. China. \\
              \email{loujing@jsut.edu.cn}           
           \and
           H. Wang \and L. Chen \and F. Xu \and Q. Xia \and W. Zhu \and M. Ren (\Letter) \at 
              School of Computer Science and Engineering, Nanjing University of Science and Technology, Nanjing, Jiangsu 210094, P.R. China. \\
              \email{renmingwu@mail.njust.edu.cn}
           \and
           Project page: \url{http://www.loujing.com/cns-sod/}
}

\date{Received: date / Accepted: date}

\maketitle

\begin{abstract}
In this paper, we will investigate the contribution of color names for the task of salient object detection. An input image is first converted to color name space, which is consisted of 11 probabilistic channels. By exploiting a surroundedness cue, we obtain a saliency map through a linear combination of a set of sequential attention maps. To overcome the limitation of only using the surroundedness cue, two global cues with respect to color names are invoked to guide the computation of a weighted saliency map. Finally, we integrate the above two saliency maps into a unified framework to generate the final result. In addition, an improved post-processing procedure is introduced to effectively suppress image backgrounds while uniformly highlight salient objects. Experimental results show that the proposed model produces more accurate saliency maps and performs well against twenty-one saliency models in terms of three evaluation metrics on three public data sets.

\end{abstract}

\section{Introduction}
\label{sec:introduction}
Visual attention, one of intrinsic properties of human vision to extract important information from abundant visual inputs, is concerned with the understanding and modeling of biological perception systems. Psychophysical and physiological studies indicate that the selective attention mechanism, which can be directed by human visual system to gaze the most conspicuous location and then shift to the next conspicuous location, plays an important role in the early representation~\cite{HN1985/Koch}. Since these conspicuous locations might be the feature cues based salient regions, the computational visual attention aims to deal with the automatic saliency detection in images or videos. In computer vision, the main tasks of saliency research include eye fixation prediction which attempts to predict human fixation data~\cite{TPAMI1998/Itti,CVPR2007/Hou,CVPR2011/Murray,ICCV2013/Zhang,JOV2013/Erdem,TPAMI2013/Li}, and salient object detection for the localization and identification of salient regions in visual scenes~\cite{CVPR2009/Achanta,CVPR2011/Cheng,CVPR2013/Yan,CVPR2013/Yang,SPL2013/Yang}.

Over the past decades, saliency detection has been widely used in many computer vision applications, including image segmentation~\cite{NC2014/Qin}, object detection~\cite{MTA2017/JingLou}, object recognition~\cite{TCSVT2014/Ren}, visual tracking~\cite{CVPR2012/Borji}, image and video compression~\cite{TIP2010/Guo}, and video summarization~\cite{CVPR2012/Lee}. Generally, the resultant map of saliency detection is called ``saliency map'', which topographically describes the conspicuity of each location in the whole scene. From a computational point of view, saliency detection techniques can be divided into two categories: slow, top-down, task-dependent manner; and rapid, bottom-up, task-independent manner~\cite{1998/Niebur}. Although top-down manner is indispensable for guiding the attention to behaviorally relevant objects, the salient features based bottom-up attention is more closely related to an early stage of visual processing~\cite{CP1980/Treisman,HN1985/Koch} and has been investigated by numerous researchers.

In the feature integration theory of attention, a visual scene is initially coded along a number of elementary features, e.g., color, orientation, brightness, and spatial frequency~\cite{CP1980/Treisman}. The selective attention mechanism~\cite{HN1985/Koch} suggests to compute these elementary features in parallel and combine the resultant cortical topographic maps into a saliency map. Hence, a majority of bottom-up saliency models aim to investigate different visual features and apply them to define the saliency of a pixel or a region. In these models, the contrast based detection is one of the most commonly adopted techniques. As no prior knowledge regarding salient objects is provided, the contrast based detection mainly focuses on two aspects: local center-surround difference, and global rarity.

For local center-surround difference, one of the most influential bottom-up saliency models is introduced by Itti et al.~\cite{TPAMI1998/Itti}. Basing on the Koch and Ullman's early representation model~\cite{HN1985/Koch}, Itti et al. extract various features at multiple resolutions and use center-surround differences between different resolutions to form a saliency map. Ma and Zhang~\cite{AcmMM2003/Ma} regard an image as a perceive field and define the saliency by measuring differences between the stimuli perceived by different perception units. Goferman et al.~\cite{CVPR2010/Goferman} exploit four basic principles of human visual attention to detect the context-aware saliency, i.e., local low-level features, global considerations, visual organization rules, and high-level factors. Furthermore, by means of the Kullback-Leibler divergence, an information-theoretic approach is proposed to extract saliency from multi-scale center-surround feature distributions~\cite{ICCV2011/Klein}.

For another, the global rarity based saliency models tend to find rare features from an image. Achanta et al.~\cite{CVPR2009/Achanta} propose a frequency-tuned (FT) approach, which defines the pixel-wise saliency by comparing the color of each pixel with the average image color in LAB color space. In~\cite{CVPR2011/Cheng}, Cheng et al. present a histogram contrast (HC) based saliency method, which uses color statistics to compute saliency. In addition, a regional contrast (RC) based saliency method is introduced in that work, which simultaneously evaluates global contrast differences and spatial coherences. In order to reduce the complexity of calculating the color contrasts between regions, we subsequently follow the RC method and propose a regional principal color (RPC) based saliency method~\cite{PONE2014/JingLou} by only retaining the most frequently occurred color of each superpixel. Besides the widely used color features, some other visual cues are also exploited in the global contrast based saliency models, such as orientation~\cite{TIP2014/Tian}, intensity~\cite{AcmMM2006/Zhai}, spectrum~\cite{CVPR2007/Hou,TPAMI2013/Li}, and texture~\cite{TIP2015/Scharfenberger}.

In this paper, we also focus on the bottom-up and contrast-based saliency detection technique. Actually, if we review the task of salient object detection, we can see it has two clear implications: one is that the detected regions should be salient in an image, the other is that these salient regions should contain objects of any category. Gestalt psychological studies indicate that objects lying in the foreground may result in being more salient than background elements~\cite{1958/Rubin,PsyR2005/Mazza}. Since salient objects are more likely to be involved in foreground regions, two questions consequently arise: 1) How to extract foreground regions? 2) How to define the contrast-based saliency? For the first question, one answer is to employ figure-ground segregation.

Recently, a simple and effective saliency model called ``Boolean Map based Saliency'' (BMS) is proposed in~\cite{ICCV2013/Zhang}. The BMS model first demonstrates that the rarity based models sometimes ignore global structure information and falsely highlight high contrast regions. Then following the suggestion of Gestalt psychology that the surroundedness may influence figure-surround segregation~\cite{1999/Palmer}, BMS exploits a set of randomly sampled boolean maps to model the saliency of foreground regions. By using different parameter settings, BMS is suitable for both eye fixation prediction and salient object detection, and achieves the state-of-the-art performance.

\begin{figure}[b]
	\centering
	\hfill%
	\subfloat[]{\includegraphics[width=0.24\linewidth]{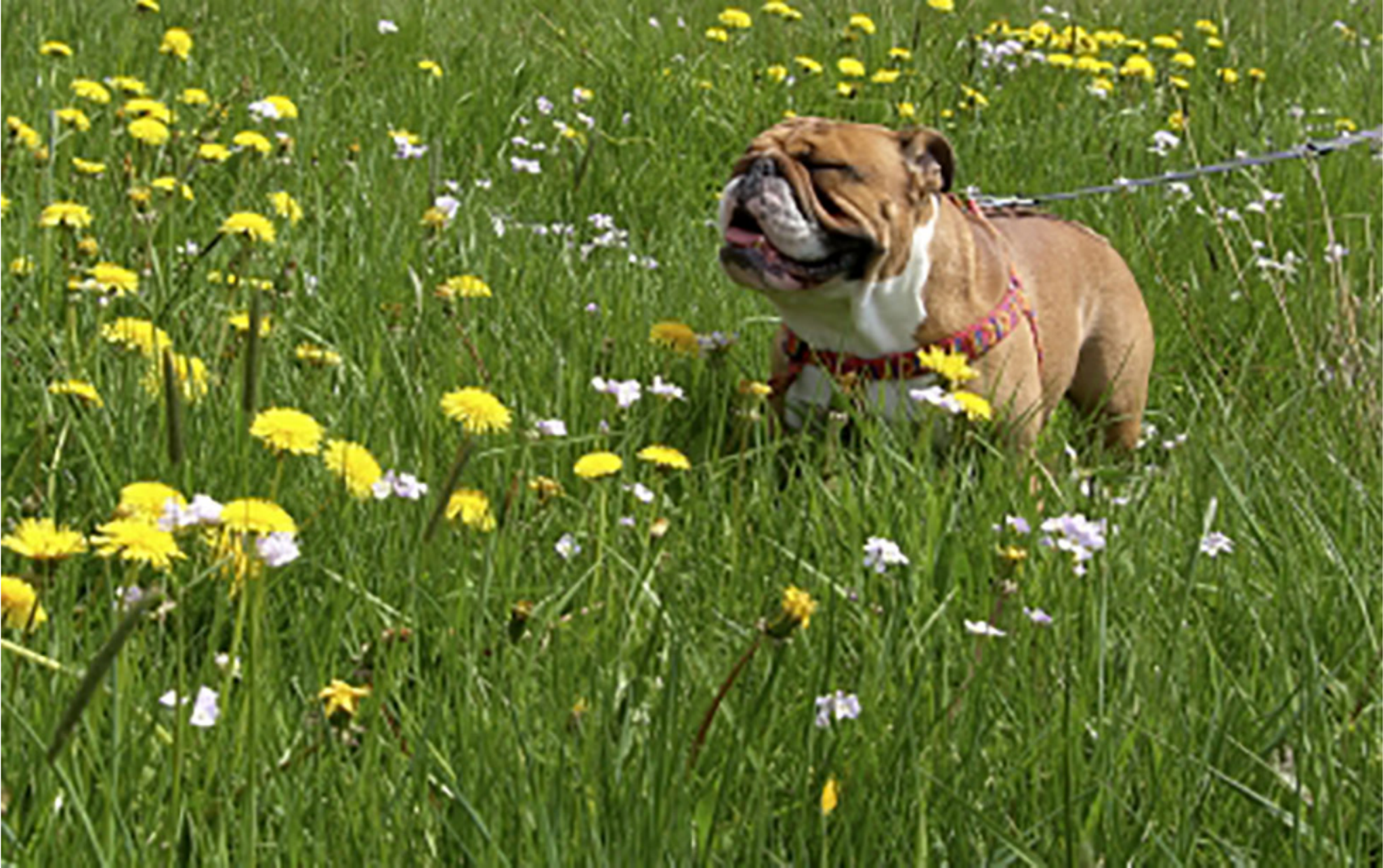}\label{fig-introduction-a}}\hspace{2pt}%
	\subfloat[]{\includegraphics[width=0.24\linewidth]{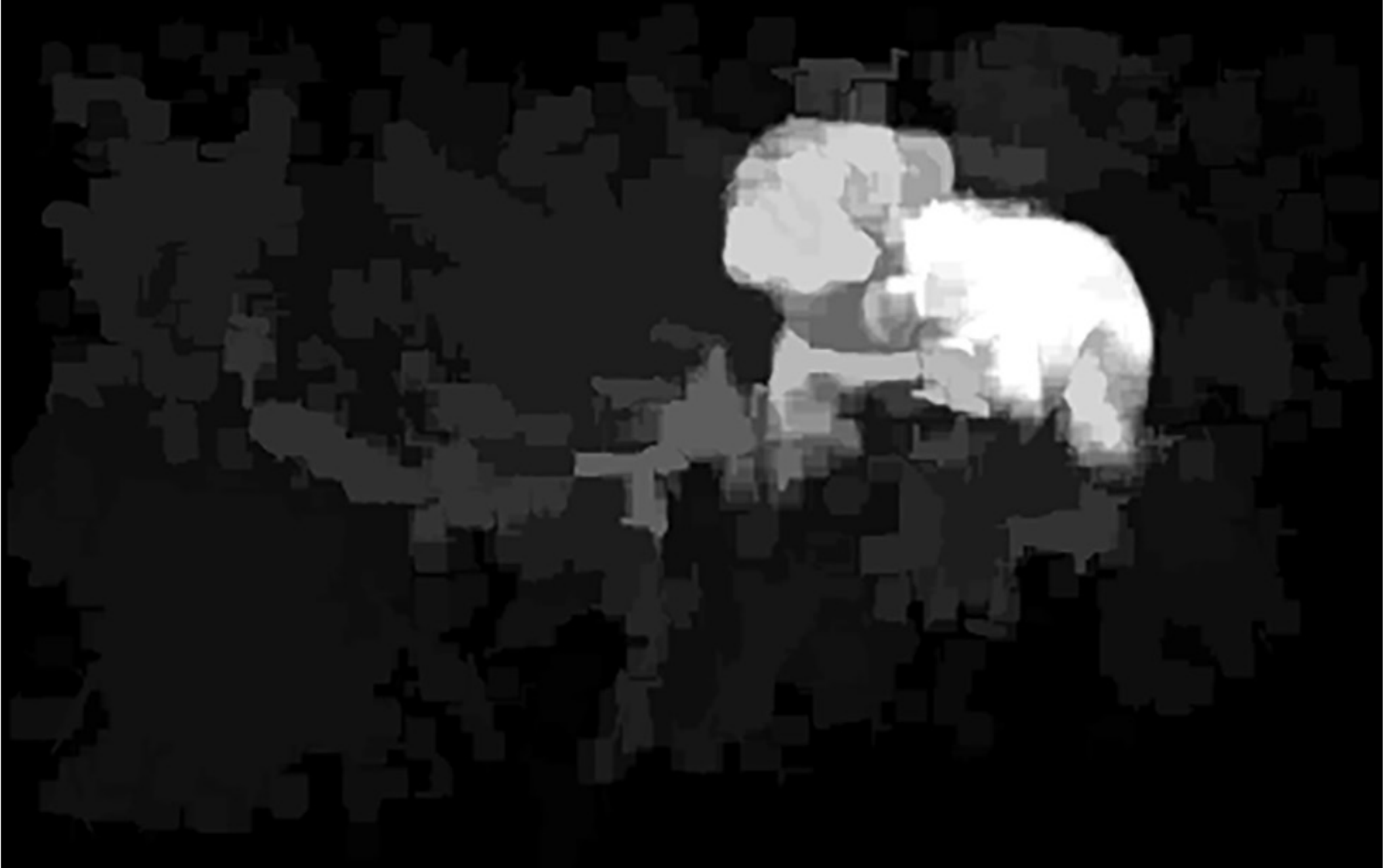}\label{fig-introduction-b}}\hspace{2pt}%
	\subfloat[]{\includegraphics[width=0.24\linewidth]{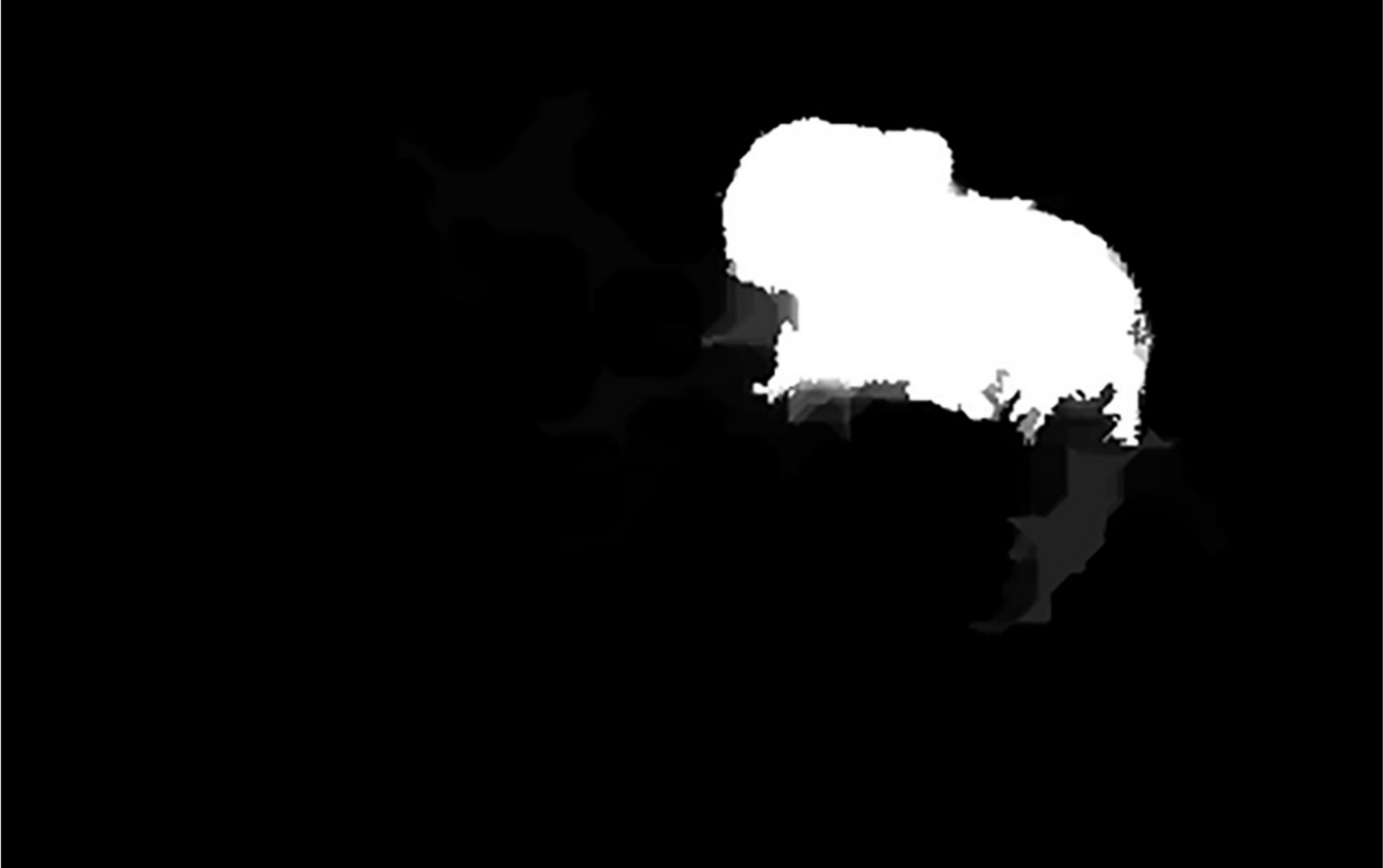}\label{fig-introduction-c}}%
	\hfill\null\\
	\hfill%
	\subfloat[]{\includegraphics[width=0.24\linewidth]{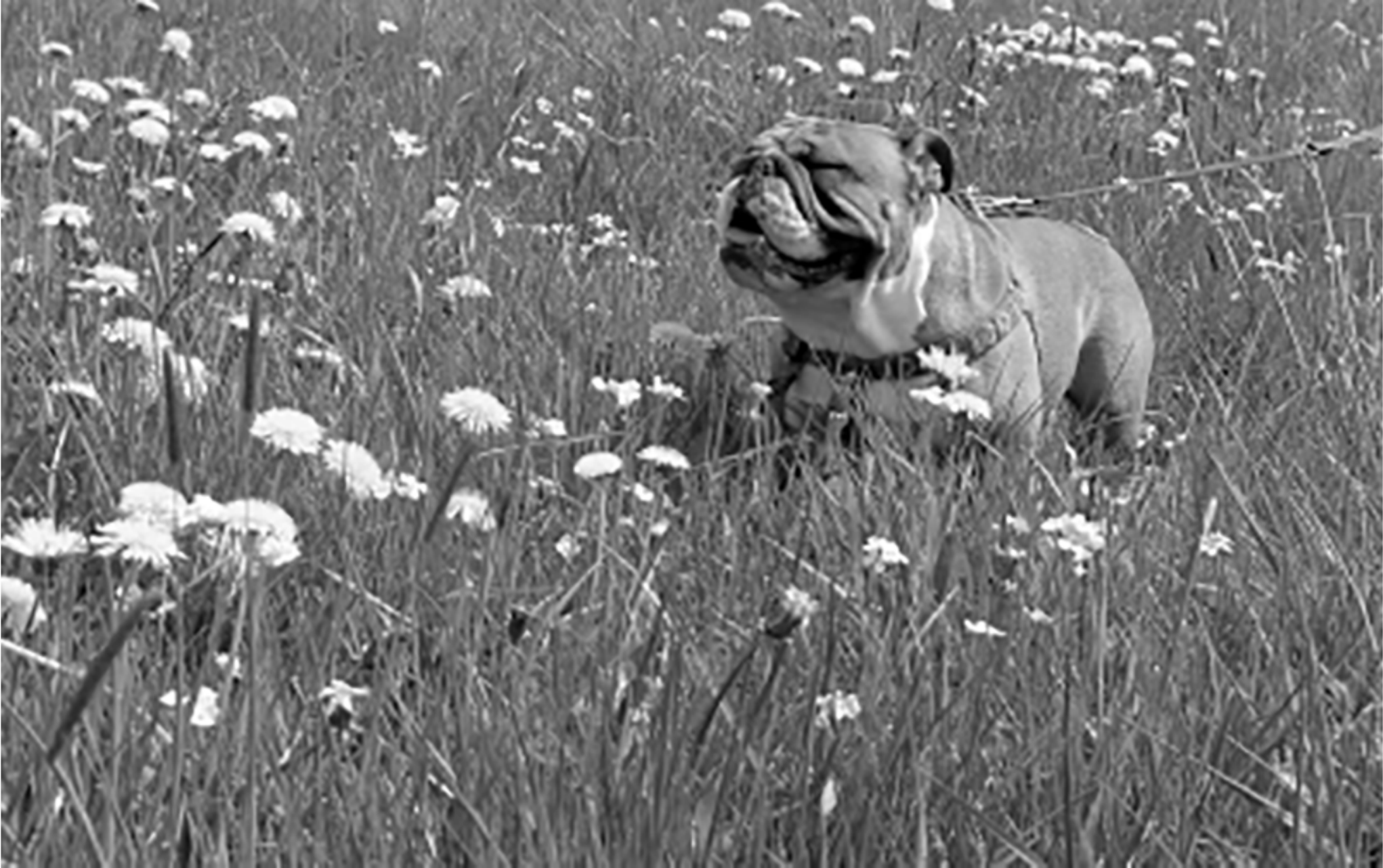}\label{fig-introduction-d}}\hspace{2pt}%
	\subfloat[]{\includegraphics[width=0.24\linewidth]{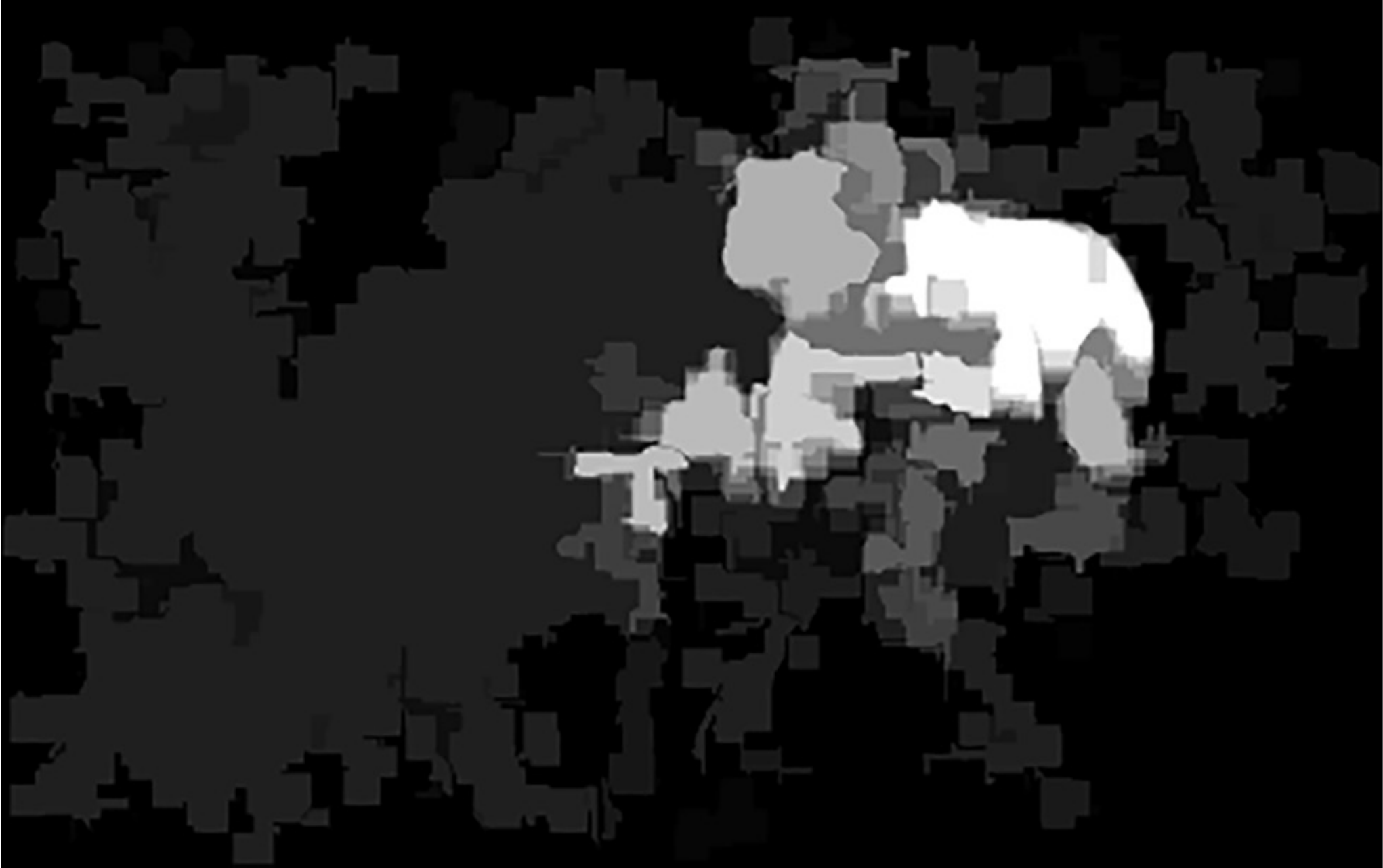}\label{fig-introduction-e}}\hspace{2pt}%
	\subfloat[]{\includegraphics[width=0.24\linewidth]{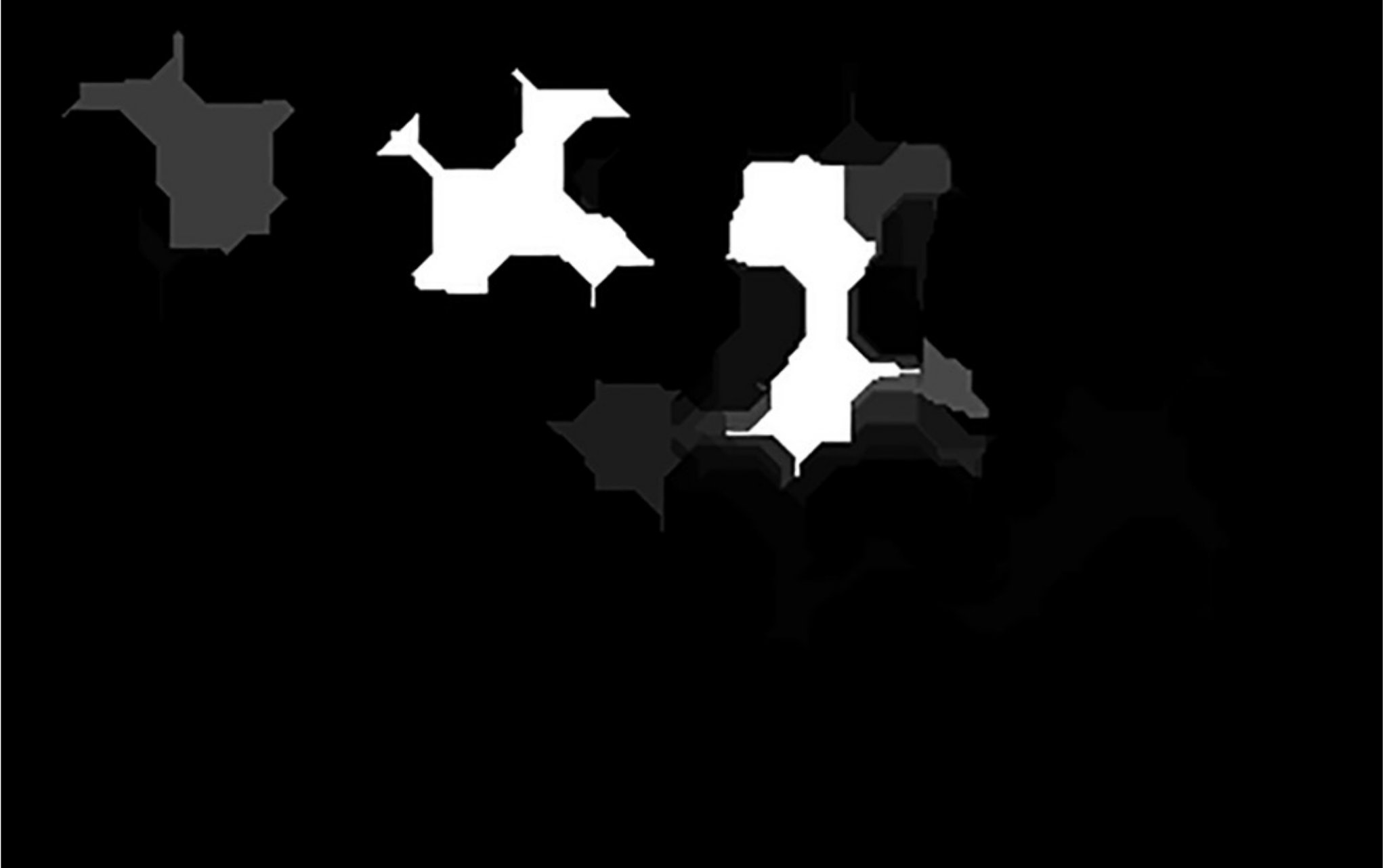}\label{fig-introduction-f}}%
	\hfill\null
\caption{(a)~RGB image from ECSSD data set~\cite{CVPR2013/Yan,TPAMI2016/Shi}, and the saliency maps generated by using (b)~BMS~\cite{ICCV2013/Zhang} and (c)~our model, respectively. (d)~The L channel obtained by converting~(a) to LAB color space, and the resultant saliency maps of (e)~BMS and (f)~our model.}
\label{fig-introduction}
\end{figure}

Here, we only discuss its results of salient object detection. Although three channels of LAB color space are chosen as the randomly sampled feature maps, the essence of BMS is the use of the closed outer contours of foreground regions. The effect of salient object detection is somewhat equivalent to applying it to a lightness image. As illustrated in Fig.~\ref{fig-introduction}, it is interesting that if we convert the input RGB image (Fig.~\ref{fig-introduction-a}) to LAB color space and apply BMS to the L channel (normalized to $[0,255]$, see Fig.~\ref{fig-introduction-d}), we obtain two similar saliency maps (cf. Figs.~\ref{fig-introduction-b} and \ref{fig-introduction-e}). The detected salient regions have similar characteristics, that is, they are enclosed by the outer boundaries and not connected to the image borders. Obviously, the color information is discarded in this case.

In this paper, we couple a surroundedness cue with two global color cues into a unified framework by extending BMS to \textit{Color Name Space}, which is obtained by using the PLSA-bg color naming model~\cite{CVPR2007/Weijer} (or called PLSA-ind in~\cite{TIP2009/Weijer}). In computer vision, color names are linguistic color labels assigned to image pixels. The linguistic study of Berlin and Kay~\cite{1969/Berlin} indicates that there are eleven basic color terms (i.e., color names) in the English language, as given in Table~\ref{table:color-names}. In the proposed model, both the probabilities and statistics of eleven color names are simultaneously incorporated to measure color differences. The topological structure information also participates in the computation of the color names based saliency, hence several weighted master attention maps are generated. Through a simple linear combination and an improved post-processing procedure, we obtain two saliency maps and then fuse them into a single map. Finally, several image processing procedures, including truncation operation, intensity mapping, and hole-filling, are invoked to infer the final result. Figures~\ref{fig-introduction-c} and~\ref{fig-introduction-f} show the saliency results produced by the proposed model. We can see that the color name space based saliency shows higher precision. It demonstrates that the color cue is of as much importance as the surroundedness cue.

\begin{table}[b]
\renewcommand{\arraystretch}{1.2}
\caption{Eleven basic color terms in the English language}
\label{table:color-names}
\centering
	\begin{tabular}{c|c|c|c|c|c}
	\hline
	$i$ & 1 & 2 & 3 & 4 & 5 \\
	\hline
	Term ($t_i$) & black & blue & brown & grey & green\\
	RGB ($c_i$)	 & [0~~0~~0] &[0~~0~~1] & [.5~~.4~~.25] & [.5~~.5~~.5] & [0~~1~~0]\\
	\hline
	\hline
	6 & 7 & 8 & 9 & 10 & 11 \\
	\hline
	orange & pink & purple & red & white & yellow\\
	\text{[1~~.8~~0]} & [1~~.5~~1] & [1~~0~~1] & [1~~0~~0] & [1~~1~~1] & [1~~1~~0]\\
	\hline
	\end{tabular}
\end{table}

In the following sections, the proposed model will be called ``CNS''. The main contributions of this paper include:
\begin{itemize}[itemsep=0cm,topsep=0cm]
\item[1)] By expoliting color name space, we propose an integrated framework to effectively compute the color based saliency.
\item[2)] A weighted global contrast mechanism is introduced to incorporate more color cues into the topological structure information of an image.
\item[3)] An improved post-processing procedure is proposed to uniformly highlight salient objects, which are easy to be segmented.
\end{itemize}

The remainder of this paper is organized as follows. Section~\ref{sec:related-work} is the review of related work. The proposed salient object detection model is presented in Section~\ref{sec:cns}. In Section~\ref{sec:experiments}, performance comparisons are made with three benchmark data sets. Conclusions and possible extensions are presented in Section~\ref{sec:conclusions}.

\section{Related Work}
\label{sec:related-work}
We base the proposed model on BMS~\cite{ICCV2013/Zhang} and PLSA-bg~\cite{CVPR2007/Weijer}. The key idea of BMS is the use of the surroundedness cue, which can be characterized by a set of boolean maps. The BMS model first converts an input RGB image $I$ to LAB color space, then scales each channel to $[0,255]$. Subsequently, BMS chooses each channel as a feature map, and uses a set of fixed thresholds to binarize each feature map to boolean maps $B_i$ as follows~\cite{ICCV2013/Zhang}:
\begin{equation}
B_i = {\textbf{\texttt{THRESH}}}\left(\phi(I), \theta\right) \,,
\end{equation}
where $\phi(I)$ is a feature map of $I$, and $\theta$ represents a fixed threshold. Based on a Gestalt principle of figure-ground segregation~\cite{1999/Palmer}, BMS performs several morphological operations to generate a set of attention maps, in which all the regions connected to the image borders are masked out since they are not surrounded by any closed outer contour. The final saliency map is simply the average of these attention maps, followed by a morphological post-processing.

The surroundedness cue is also invoked in the proposed CNS model. However, different from BMS, CNS uses color name space instead of LAB color space. In the field of document analysis, the standard PLSA model~\cite{SIGIR1999/Hofmann} computes the conditional probability of a word $w$ in a document $d$, and estimates the distributions $p(z|d)$ and $p(w|z)$ by using an Expectation-Maximization (EM) algorithm, where $z$ represents a latent topic. Considering that PLSA does not exploit the color name labels of training images, the PLSA-bg model~\cite{CVPR2007/Weijer} represents an image $d$ (i.e., document) as a LAB color histogram with a group of color bins (i.e., words), and decomposes $d$ into the foreground distribution according to a given color name label $l_d$ (i.e., topic) and the background distribution shared between all training images. By estimating the mixing proportion of foreground versus background, color name distributions, and background model, the probability of a color name for a given image pixel is represented as
\begin{equation}
p(z|w) \propto p(z) p(w|z) \,,
\end{equation}
where the prior over eleven color names is taken to be uniform.

Besides the probability information of color names, the proposed model makes use of a statistical cue. This is achieved by a \textit{Color Name Histogram}, in which eleven color name bins are involved for measuring color differences. In~\cite{CVPR2011/Cheng}, the HC method directly uses color statistics to define the saliency value of each color bin. Compared with HC, our model solely exploits the color name histogram to compute eleven weighting coefficients, and further produces eleven weighted master attention maps. The color name histogram does not participate in the generation of original attention maps, which are still determined by the surroundedness cue as used in BMS.

\section{Color Name Space Based Saliency Detection}
\label{sec:cns}
To incorporate more color information, we extend BMS~\cite{ICCV2013/Zhang} from LAB color space to color name space. Two saliency cues, i.e., surroundedness and color, are separately invoked to produce two saliency maps. They are fused into a single map for generating the final result. These steps are described in the following sections.

\subsection{General Framework}
\label{sec:framework}

\begin{figure*}[t]
	\centering
	\includegraphics[width=\linewidth]{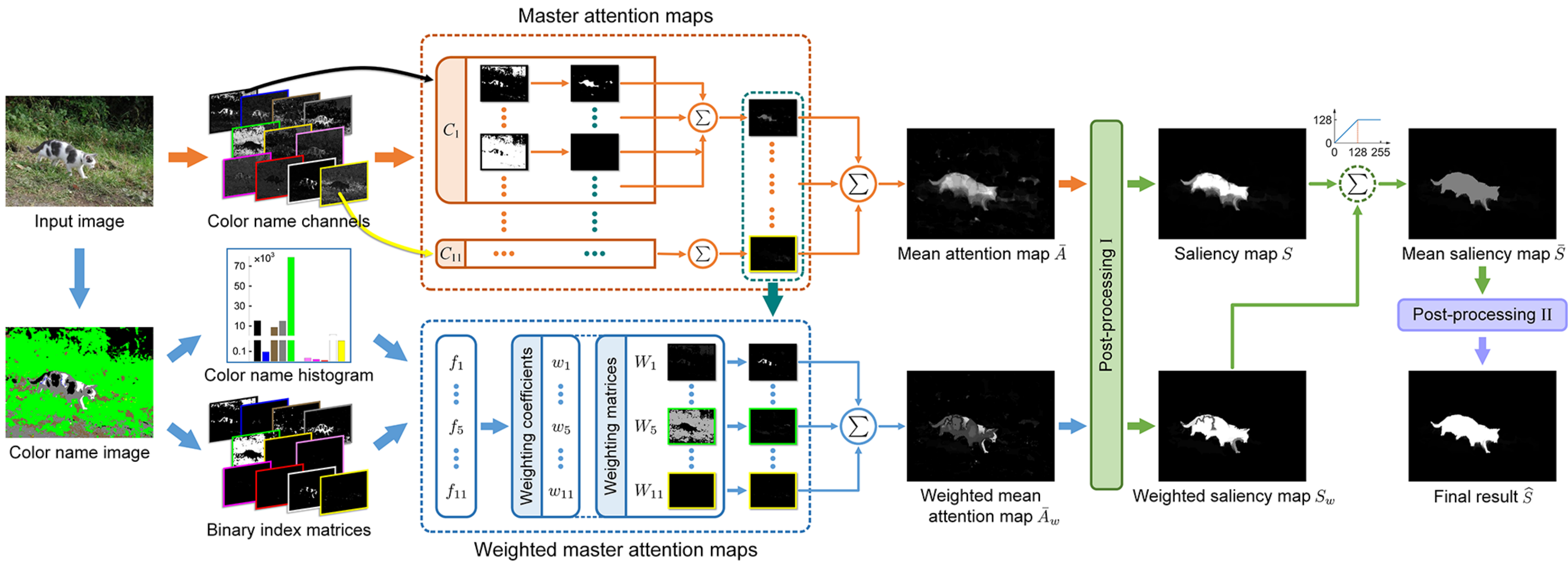}
	\caption{Framework of the proposed CNS model.}
	\label{fig-framework}
\end{figure*}

As illustrated in Fig.~\ref{fig-framework}, the integrated framework of CNS includes two computational pipelines.

\smallskip
\noindent\textbf{Pipeline I.}~~An input RGB image is first resized to 400 pixels in width and converted to color name space. The resultant space is composed of 11 monochrome intensity components, namely \textit{Color Name Channel} in this paper. Following BMS~\cite{ICCV2013/Zhang}, a set of attention maps is generated based on figure-ground segregation. The attention maps of each channel are linearly fused to produce a master attention map. Finally, the mean attention map $\bar{A}$ is obtained by combining 11 master attention maps and further post-processed to generate the saliency map $S$.

\smallskip
\noindent\textbf{Pipeline II.}~~The resized RGB image is first converted to a \textit{Color Name Image}, from which two statistical characteristics are derived: 1) a color name histogram which consists of 11 color bins, and 2) 11 binary index matrices where each of them represents the distribution of a corresponding color name. By exploiting two kinds of weighting patterns, we generate 11 weighted master attention maps. All the master attention maps obtained in Pipeline I also participate in this process. Finally, the weighted saliency map $S_w$ is obtained by using the same combination and post-processing as used in Pipeline I.

\smallskip
\noindent\textbf{Combination.}~~The saliency maps $S$ and $S_w$ are fed into a truncation operation to produce the mean saliency map $\bar{S}$, which simultaneously codes for the topological structure and color conspicuity over the entire image. In addition, we apply another post-processing procedure to generate the final saliency result $\widehat{S}$, in which the salient region is evenly highlighted and smoothed for convenience in the task of salient object segmentation.

\subsection{Color Name Channel Based Attention Map}
\label{sec:attention-map}
First, we directly use the \textbf{im2c} function provided by~\cite{CVPR2007/Weijer}\footnote{\url{http://lear.inrialpes.fr/people/vandeweijer/color_names.html}} to generate the color name space $\mathbf{C}=\{C_1,C_2,\ldots,C_{11}\}$, where each color name channel $C_i$ has the range of values $[0,1]$. Thus, for the resized RGB image $I$, the color representation of each pixel is mapped from a 3-dimensional (3-D) RGB value to a probabilistic 11-dimensional (11-D) vector which sums up to 1. Considering that the topological structure of $I$ is independent of the perceptual color coherence, each color name channel is treated equally and normalized to $[0,255]$ for the subsequent thresholding operation.

Then, we use a set of sequential thresholds from 0 to 255 with a step size of $\delta$ to binarize each color name channel $C_i\in \mathbf{C}$ to $n$ boolean maps
\begin{equation}
B_i^j = {\texttt{\textbf{THRESH}}}\left(C_i,~\theta_j\right) \,,
\end{equation}
where at each threshold $\theta_j$, the above function generates a boolean map $B_i^j$ by setting all the values above $\theta_j$ to 1s and replacing all the others with 0s. After two morphological operations on $B_i^j$, including closing and hole-filling, we 
use a clear-border algorithm~\cite{1999/Soille} to mask out all the foreground regions connected to the image borders, and obtain a corresponding attention map $A_i^j$. The same processing steps are also executed for the complement map of $B_i^j$ (denoted $\widetilde{B}_i^j$). As summarized in Algorithm~\ref{algo-attention-map}, two parameters are required in this stage: sample step $\delta$, and kernel radius $\omega_c$ of the closing operation. We will discuss the influences of them in Section~\ref{sec:parameter-analysis}.

\renewcommand{\algorithmicrequire}{\textbf{Input:}}
\renewcommand{\algorithmicensure}{\textbf{Output:}}
\begin{algorithm}[b]
\caption{attention map computation}
\label{algo-attention-map}
\begin{small}
	\begin{algorithmic}[1]
	\Require resized RGB image $I$
	\Ensure attention maps $A_i^j$ and $\widetilde{A}_i^j$
	\State {convert $I$ from RGB space to color name space $\mathbf{C}$}
	\For {each $C_i \in \mathbf{C}$}
		\For {$\theta_j = 0:\delta:255$}
		\State {$B_i^j = {\texttt{\textbf{THRESH}}}(C_i,\theta_j)$}
		\State {$B_i^j = {\texttt{\textbf{CLOSE}}}(B_i^j, \omega_c)$}
		\State {$B_i^j = {\texttt{\textbf{FILL}}}(B_i^j)$}
		\State {$A_i^j = {\texttt{\textbf{CLEAR-BORDER}}}(B_i^j)$}
		\vspace{0.5em}
		\State {$\widetilde{B}_i^j = {\texttt{\textbf{INVERT}}}(B_i^j)$}
		\State {$\widetilde{B}_i^j = {\texttt{\textbf{CLOSE}}}(\widetilde{B}_i^j, \omega_c)$}
		\State {$\widetilde{B}_i^j = {\texttt{\textbf{FILL}}}(\widetilde{B}_i^j)$}
		\State {$\widetilde{A}_i^j = {\texttt{\textbf{CLEAR-BORDER}}}(\widetilde{B}_i^j)$}
		\EndFor
	\EndFor
	\end{algorithmic}
\end{small}
\end{algorithm}

However, different from BMS which averages all the attention maps, the proposed model computes the mean attention map $A_i$ of each color name channel $C_i$ separately. All the attention maps $A_i^j$ and $\widetilde{A}_i^j$ share the same weight and are averaged to $A_i$, which is called \textit{Master Attention Map} in this paper. Then, the mean attention map $\bar{A}$ of 11 master attention mpas can be further calculated as follows:
\begin{align}
A_i &= \frac{1}{2n} \sum_{j=1}^{n}  \left(A_i^j + \widetilde{A}_i^j \right) \,, \label{eq:A_i} \\
\bar{A} &= \frac{1}{11} \sum_{i=1}^{11} A_i \,. \label{eq:barA}
\end{align}

Actually, if we merge Eqs.~\eqref{eq:A_i}~and~\eqref{eq:barA}, we can get the same computation procedure of $\bar{A}$ as introduced in the BMS model~\cite{ICCV2013/Zhang}. The slight difference lies in the usage of 11 master attention maps. In Pipeline I, the computation of $\bar{A}$ is mainly based on the surroundedness cue. To make better use of color name space, the proposed framework couples the surroundedness cue with two color cues to compute the color based saliency. In Section~\ref{sec:weighting}, we will again use the 11 master attention maps to produce a weighted mean attention map $\bar{A}_w$.

\subsection{Post-processing}
\label{sec:post-processing}
\begin{figure}[b]
	\centering
	\hfill%
	\subfloat[]{\includegraphics[width=0.24\linewidth]{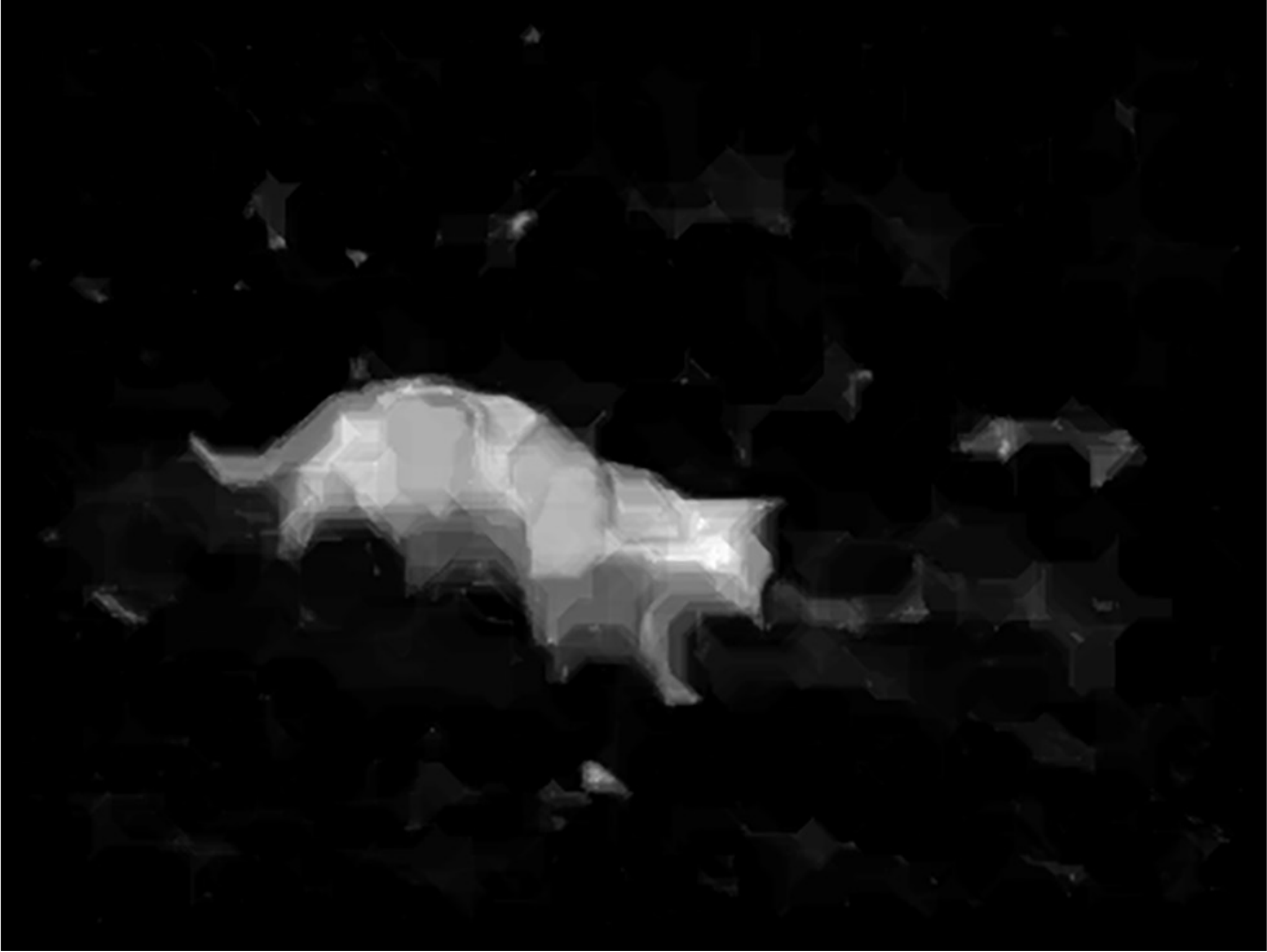}\label{fig-post-processing-I-a}}\hspace{1pt}%
	\subfloat[]{\includegraphics[width=0.24\linewidth]{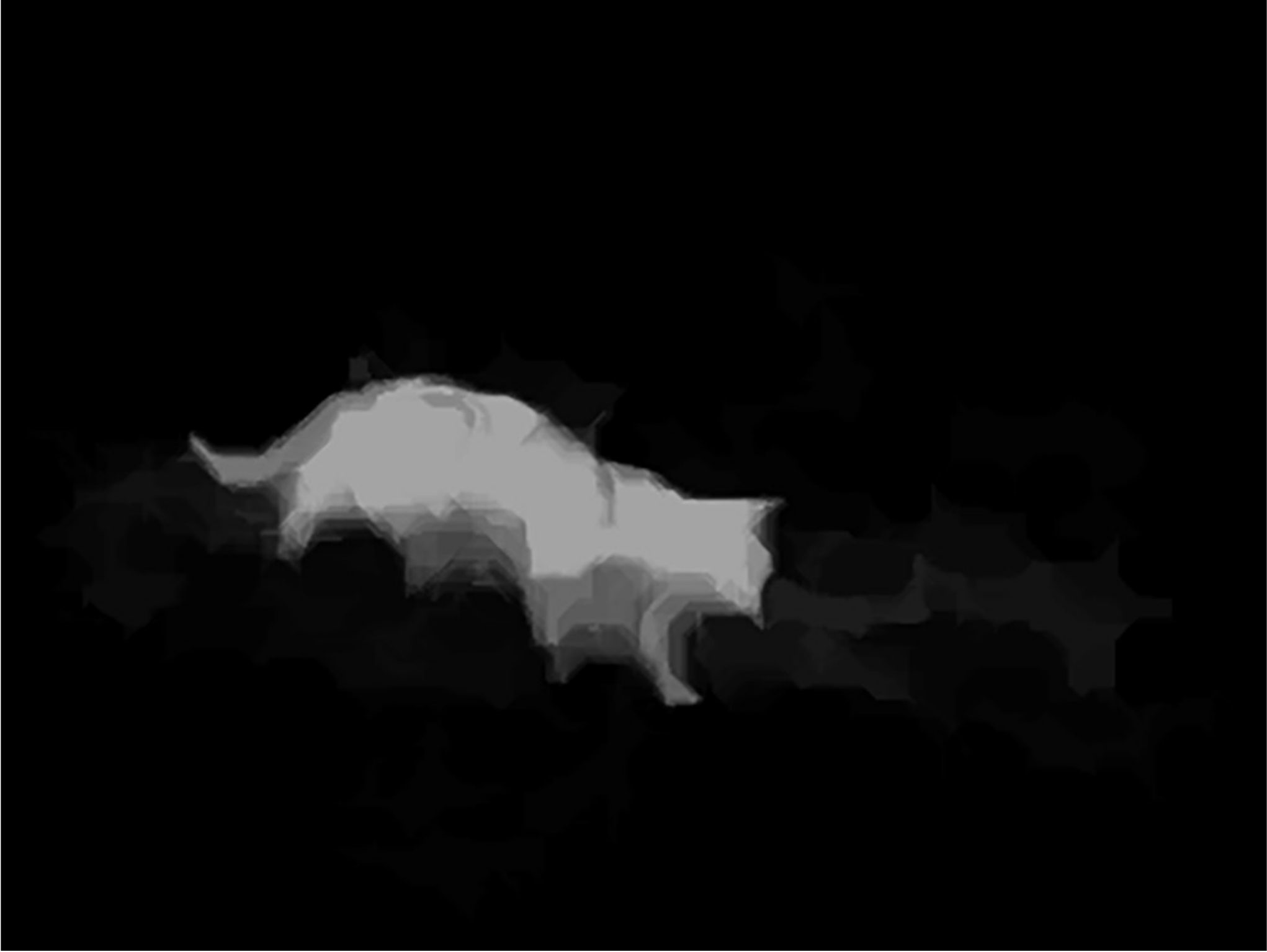}\label{fig-post-processing-I-b}}\hspace{1pt}%
	\subfloat[]{\includegraphics[width=0.24\linewidth]{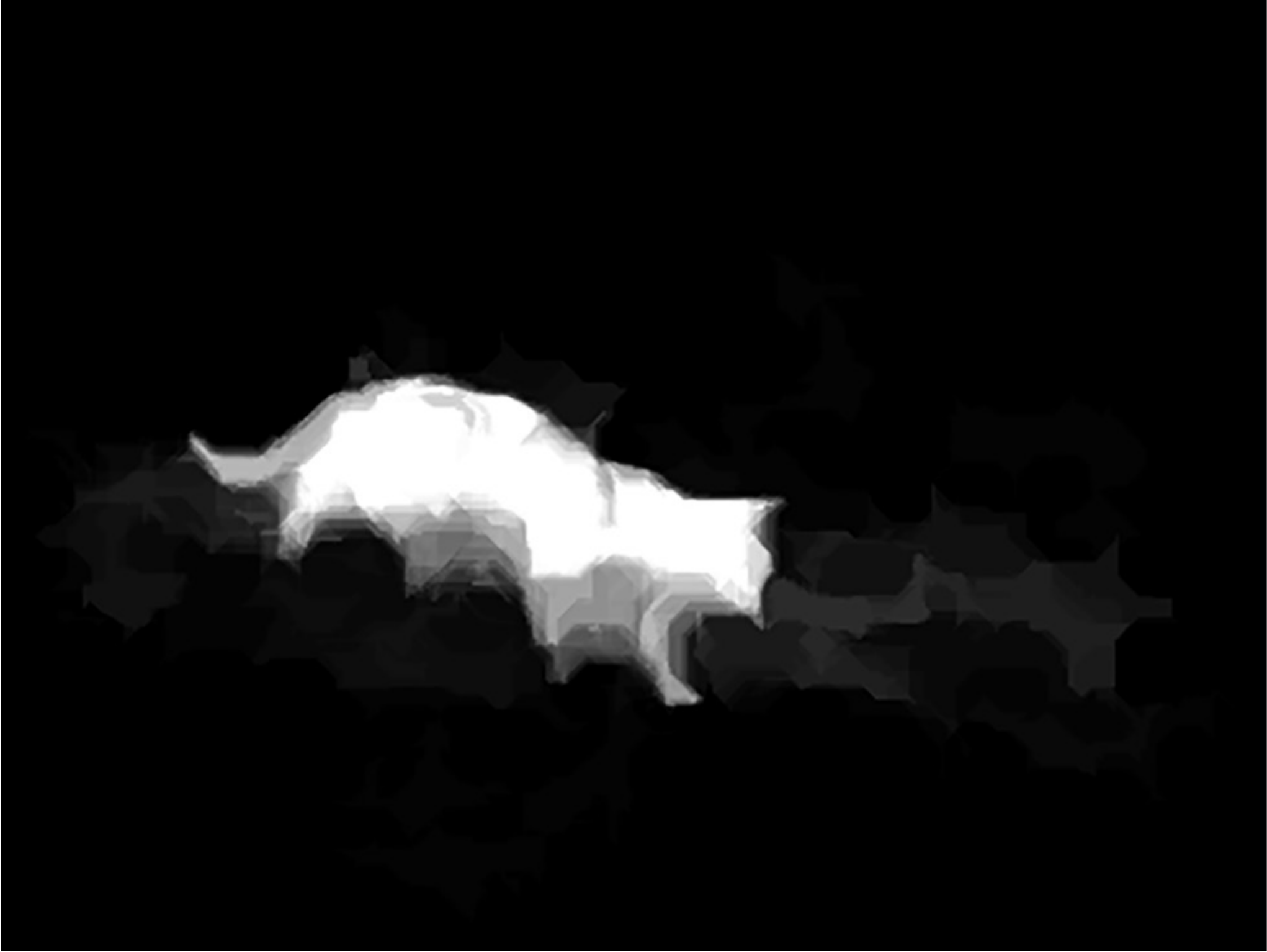}\label{fig-post-processing-I-c}}\hspace{1pt}%
	\subfloat[]{\includegraphics[width=0.24\linewidth]{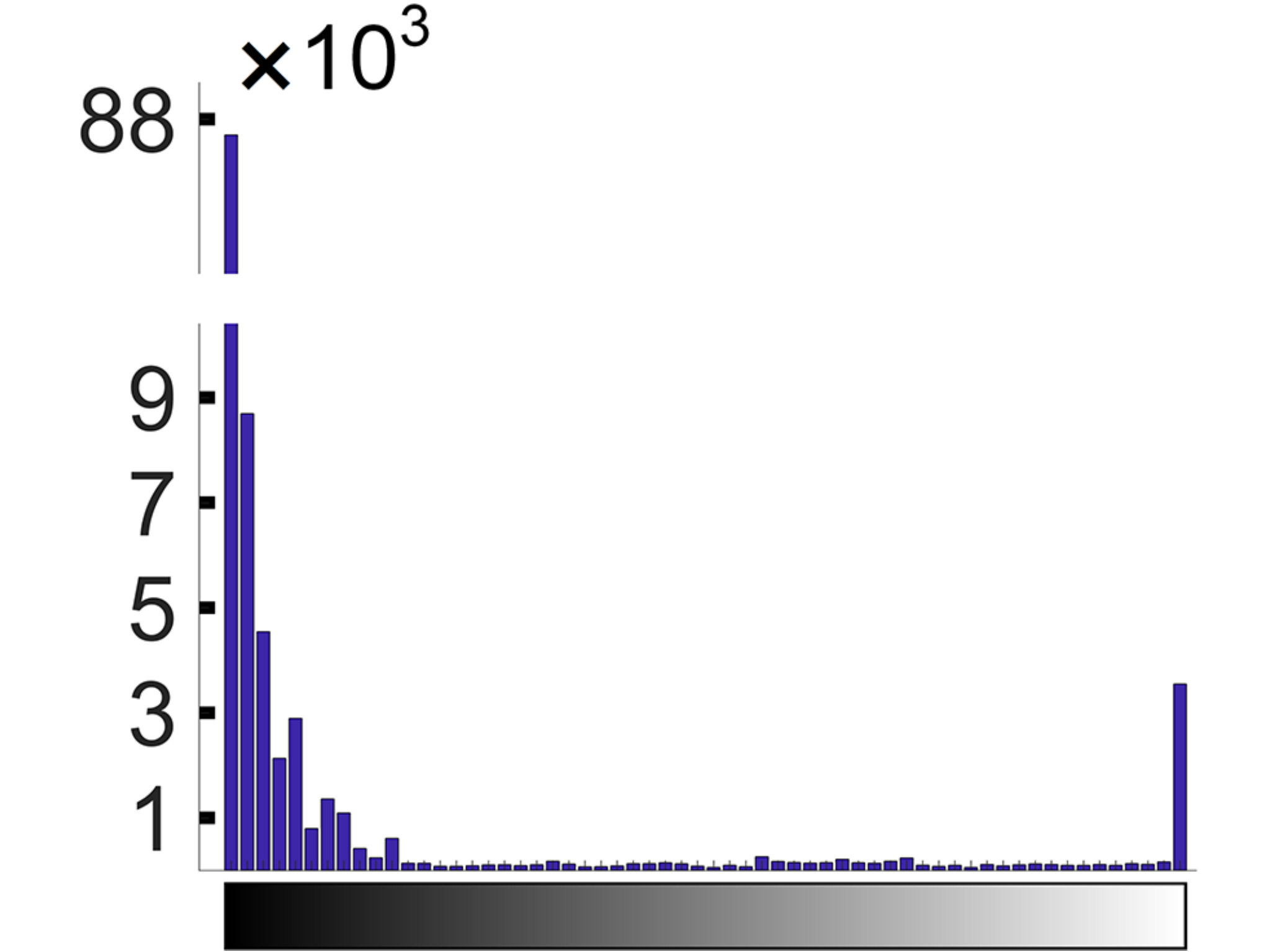}\label{fig-post-processing-I-d}}%
	\hfill\null
	\vfill
	\hfill%
	\subfloat[]{\includegraphics[width=0.24\linewidth]{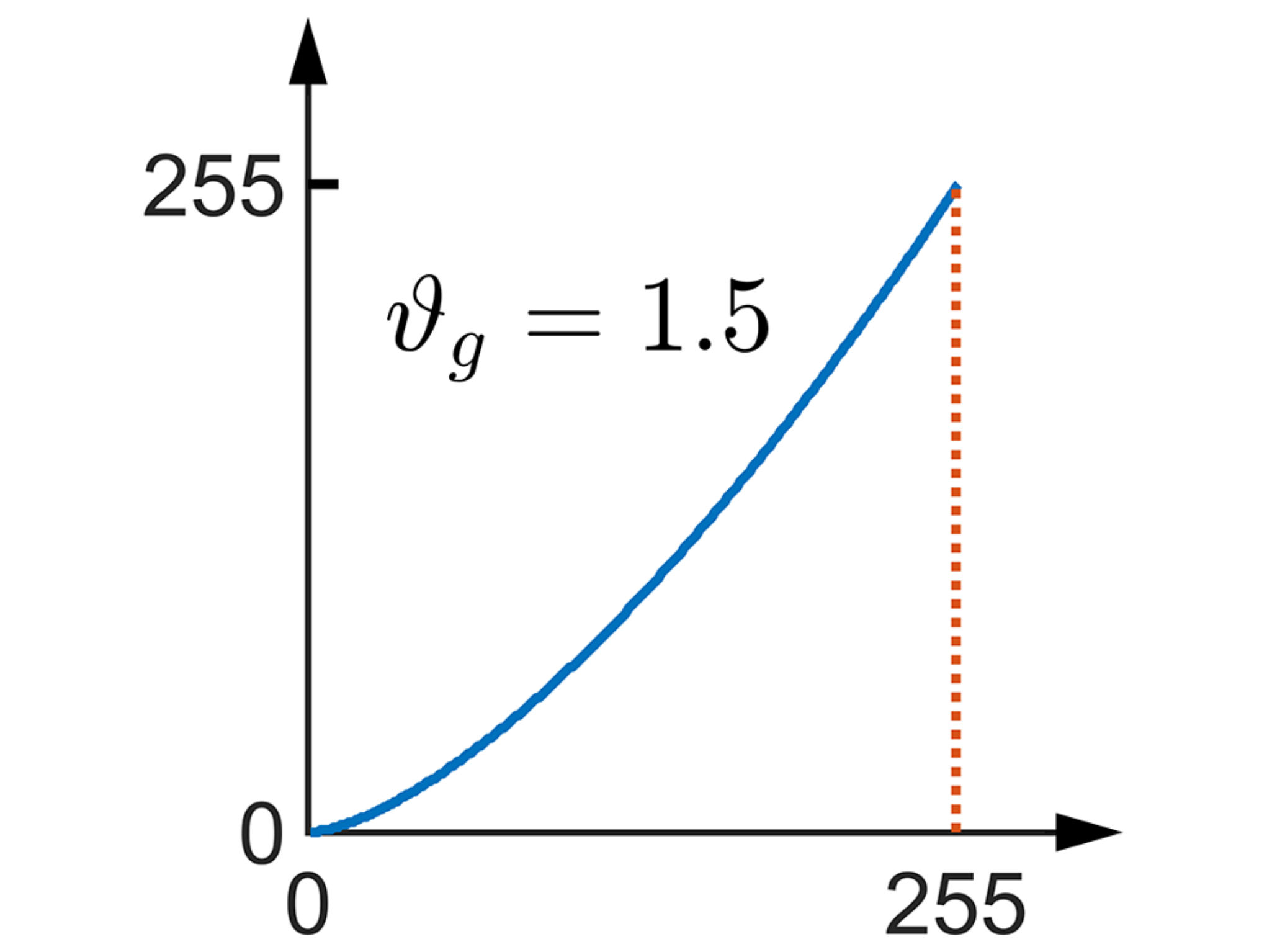}\label{fig-post-processing-I-e}}\hspace{1pt}%
	\subfloat[]{\includegraphics[width=0.24\linewidth]{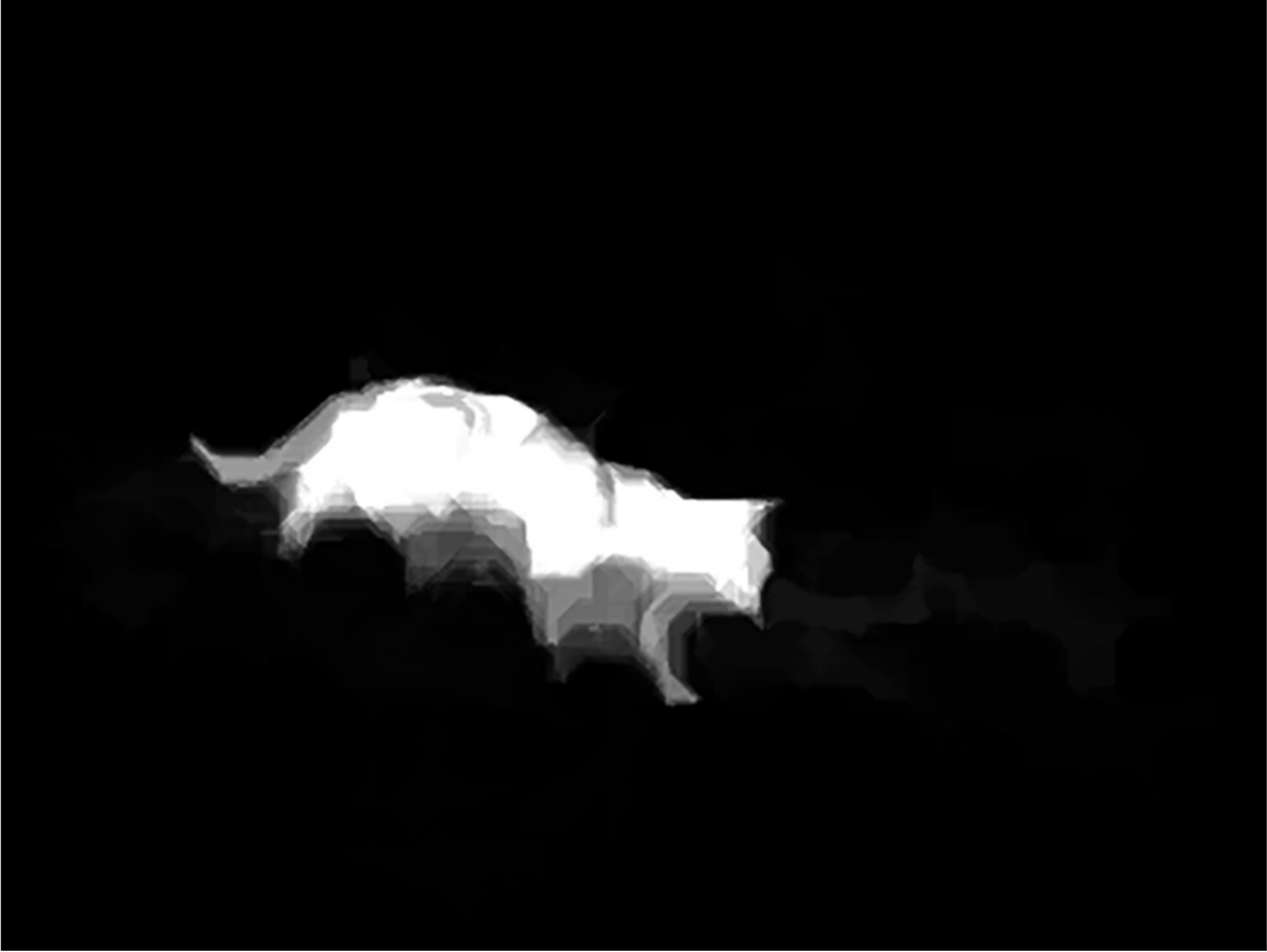}\label{fig-post-processing-I-f}}\hspace{1pt}%
	\subfloat[]{\includegraphics[width=0.24\linewidth]{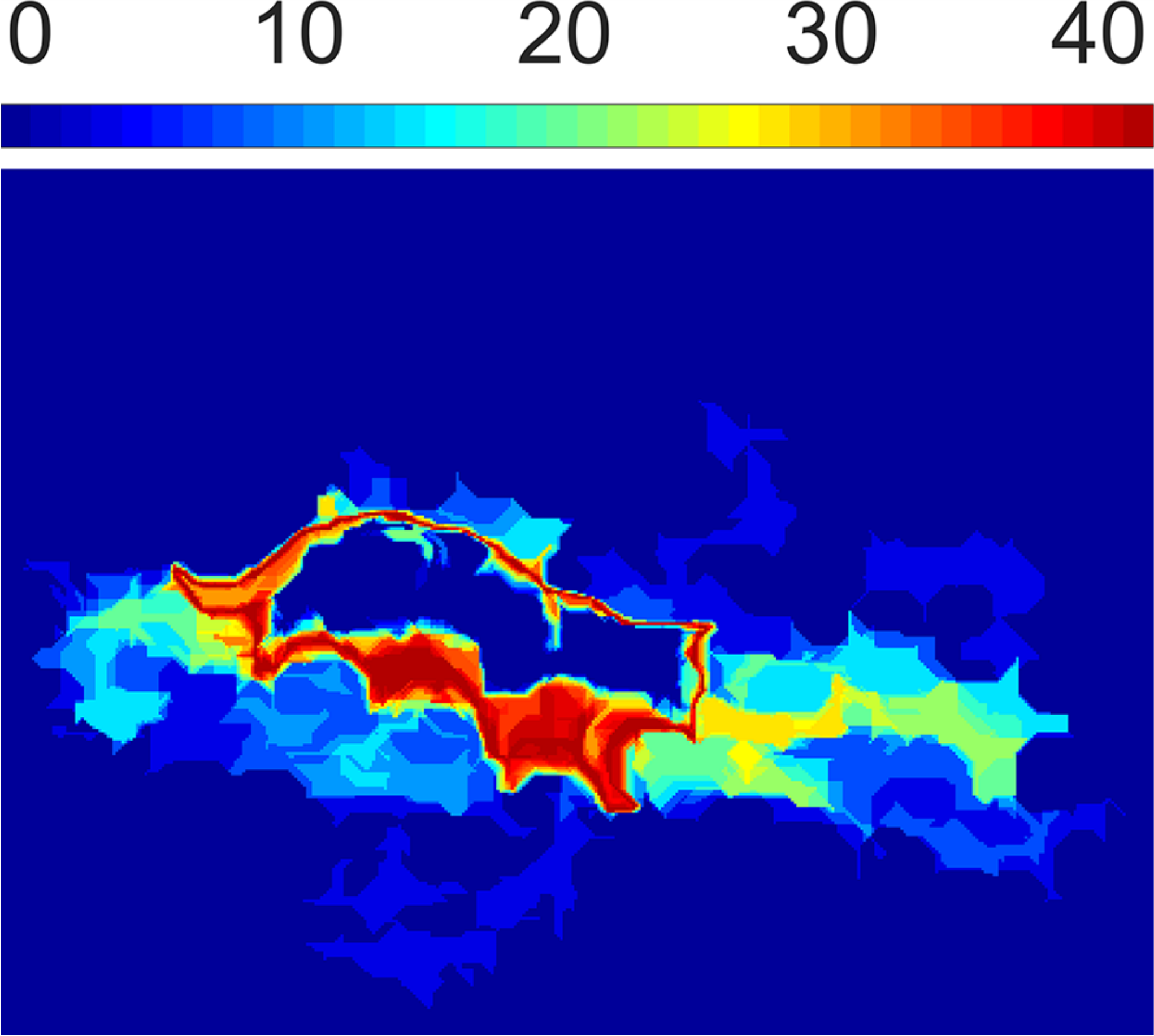}\label{fig-post-processing-I-g}}\hspace{1pt}%
	\subfloat[]{\includegraphics[width=0.24\linewidth]{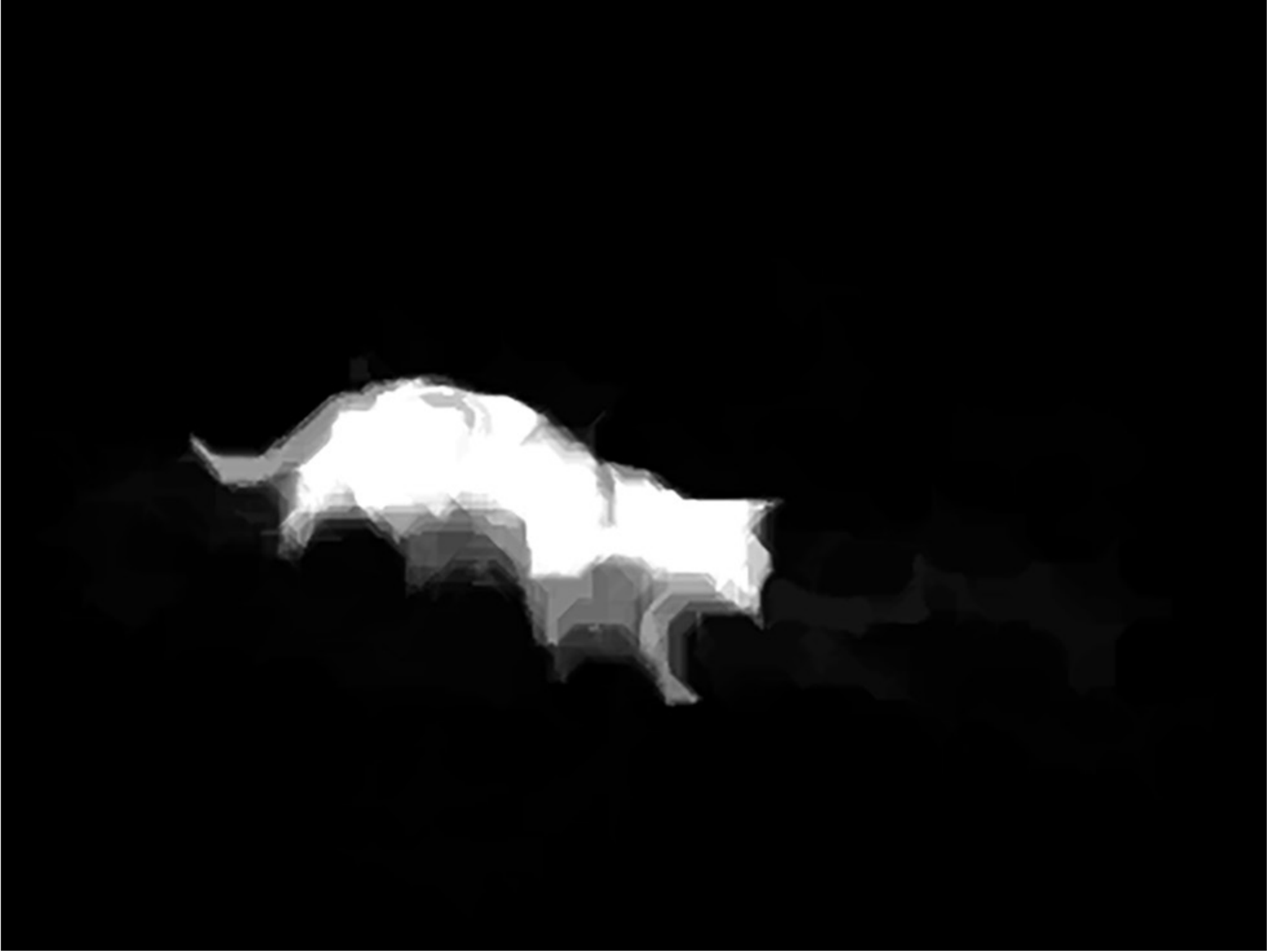}\label{fig-post-processing-I-h}}%
	\hfill\null
\caption{Post-processing I. (a)~Mean attention map $\bar{A}$. (b)~Morphological reconstruction. (c)~Normalization result, and (d)~its histogram. (e)~Intensity mapping curve. (f)~Mapping result obtained by using $\vartheta_r=0.02$ and $\vartheta_g=1.5$. (g)~Difference between (c) and (f). (h)~Saliency map $S$.}
\label{fig-post-processing-I}
\end{figure}

The mean attention map $\bar{A}$ is shown in Fig.~\ref{fig-post-processing-I-a}. Due to the existence of other surrounded objects that have clear boundaries and uniform colors (for example, the red flower below the cat), there are several small salient regions in $\bar{A}$. In order to outstand the main salient object (i.e., the cat), we also follow BMS to remove small salient regions by sequentially performing two steps of morphological reconstruction operations~\cite{TIP1993/Vincent,2004/Gonzalez}. The structuring element used here is a disk shape with the radius $\omega_r$. Figure~\ref{fig-post-processing-I-b} shows the reconstruction result. It can be observed that those small salient regions have been erased while the original shape of the salient cat is still remained.

For the task of salient object detection, the ideal output should be a binary map in which the pixel values of salient objects are 1s while the others are 0s. However, the disadvantage of the morphological reconstruction is that the high intensity values of salient pixels are suppressed simultaneously. In addition, the background of the reconstruction result contains some inconspicuous regions with non-black pixels, which would decrease the detection precision. To address the above issues, a nonlinear mapping function is introduced to transform the intensity values to a new range. Overall, we wish to weight the mapping toward the lower output values and map all the intensity values above a specific threshold to 1s. Suppose that $F$ is the reconstruction result, the intensity mapping function has the syntax form as follows:
\begin{equation}
\label{eq:Map}
G = {\texttt{\textbf{MAP}}}\left(F,~[0,T_F],~\vartheta_g \right) \,,
\end{equation}
where $T_F$ is the truncation threshold, $\vartheta_g$ determines the mapping relationship between $F$ and $G$. To suppress non-salient pixels, the lower limit of the mapping is set to 0, and $\vartheta_g$ is set to be greater than 1.

In Eq.~\eqref{eq:Map}, all the intensity values above $T_F$ (i.e., in the interval $[T_F,255]$) are clipped and mapped to 1s. For automatically obtaining $T_F$, we exploit the statistical information extracted from the histogram of $F$. After scaling $F$ to $[0,255]$ (see Fig.~\ref{fig-post-processing-I-c}), we get its histogram $H$ where $H_k,k\in[0,255]$ denotes the number of pixels at the $k$th intensity level. By summing up the number of pixels from $H_0$, we obtain the minimum intensity level $T_F$ which should satisfy the following criteria:
\begin{equation}
\label{eq:T_F}
(1-\vartheta_r)\sum_{k=0}^{255}{H_k} \;\leqslant\; \sum_{k=0}^{T_F}{H_k}  \,,
\end{equation}
that is, the non-salient pixels should cover no less than $1-\vartheta_r$ of the total number of image pixels. For convenience, we abbreviate Eq.~\eqref{eq:Map} as
\begin{equation}
G = {\texttt{\textbf{MAP}}}\left(F,~\vartheta_r,~\vartheta_g \right) \,,
\end{equation}
where $\vartheta_r$ is empirically set to be less than 10\%. 

Figure~\ref{fig-post-processing-I-e} illustrates the intensity mapping curve with $\vartheta_r=0.02$ and $\vartheta_g=1.5$. By using this mapping, the lower (darker) values in the output map (Fig.~\ref{fig-post-processing-I-f}) are further suppressed. From the difference map (Fig.~\ref{fig-post-processing-I-g}), we can see that those non-salient regions on the right side of the cat have been eliminated. Finally, we perform a hole-filling operation to generate the saliency map $S$. The whole post-processing procedure is summarized in Algorithm~\ref{algo-post-processing-I}. In Section~\ref{sec:parameter-analysis}, we will discuss the influences of the parameters $\omega_r$, $\vartheta_r$, and $\vartheta_g$.

\renewcommand{\algorithmicrequire}{\textbf{Input:}}
\renewcommand{\algorithmicensure}{\textbf{Output:}}
\begin{algorithm}[b]
\caption{post-processing I}
\label{algo-post-processing-I}
\begin{small}
	\begin{algorithmic}[1]
	\Require mean attention map $\bar{A}$
	\Ensure saliency map $S$

	\State {$S = {\texttt{\textbf{RECONSTRUCT}}}\,(\bar{A},~\omega_r)$}
	\State {$S = {\texttt{\textbf{NORMALIZE}}}\,(S,~[0,255])$}
	\State {$S = {\texttt{\textbf{MAP}}}\,(S,~\vartheta_r,~\vartheta_g)$}
	\State {$S = {\texttt{\textbf{FILL}}}\,(S)$}
	\end{algorithmic}
\end{small}
\end{algorithm}

\subsection{Global Color Cue Based Saliency}
\label{sec:weighting}
As indicated previously, we then introduce a color-based saliency algorithm to overcome the limitation of only using the surroundedness cue. In order to take advantage of color attributes, two global color cues including statistic and contrast, are inferred from color name image and employed to compute weighting coefficients and matrices. The 11 master attention maps obtained in Section~\ref{sec:attention-map} are coupled with two kinds of weights to further produce a weighted saliency map $S_w$.

First, we again use the \textbf{im2c} function~\cite{CVPR2007/Weijer} to convert each pixel value of the resized RGB image $I$ from a 3-D RGB value to a probabilistic 11-D vector. By exploiting the index number of the largest element in the vector, we construct an index map $\mathbf{M}$ where each pixel has an integer value between 1 and 11. Basing on $\mathbf{M}$, we derive two kinds of weights.

\subsubsection{Color Name Statistic Based Weights}
\begin{figure}[b]
	\centering
	\hfill%
	\subfloat[]{\includegraphics[width=0.3\linewidth]{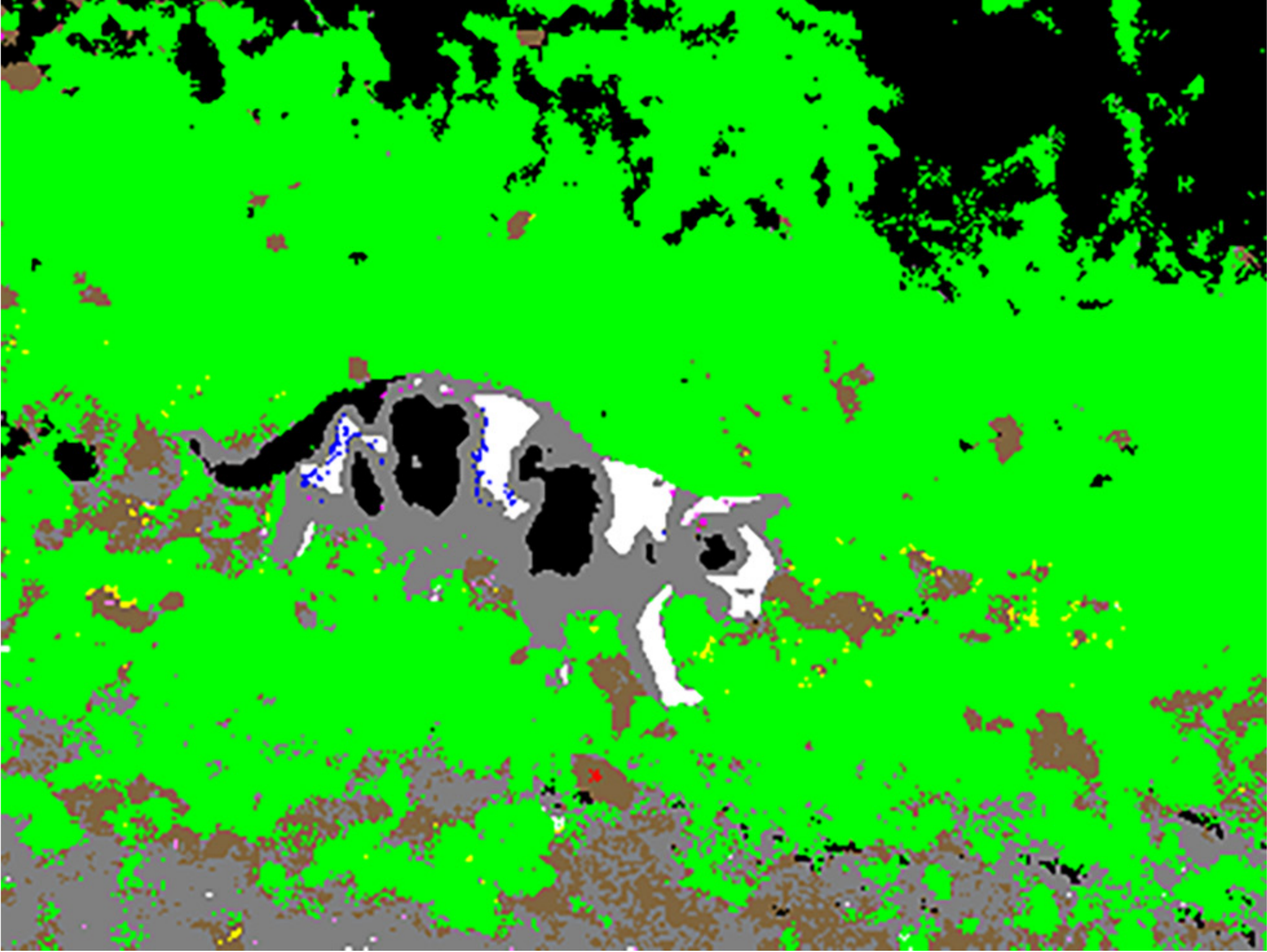}\label{fig-color-name-image-a}}\hfill%
	\subfloat[]{\includegraphics[width=0.3\linewidth]{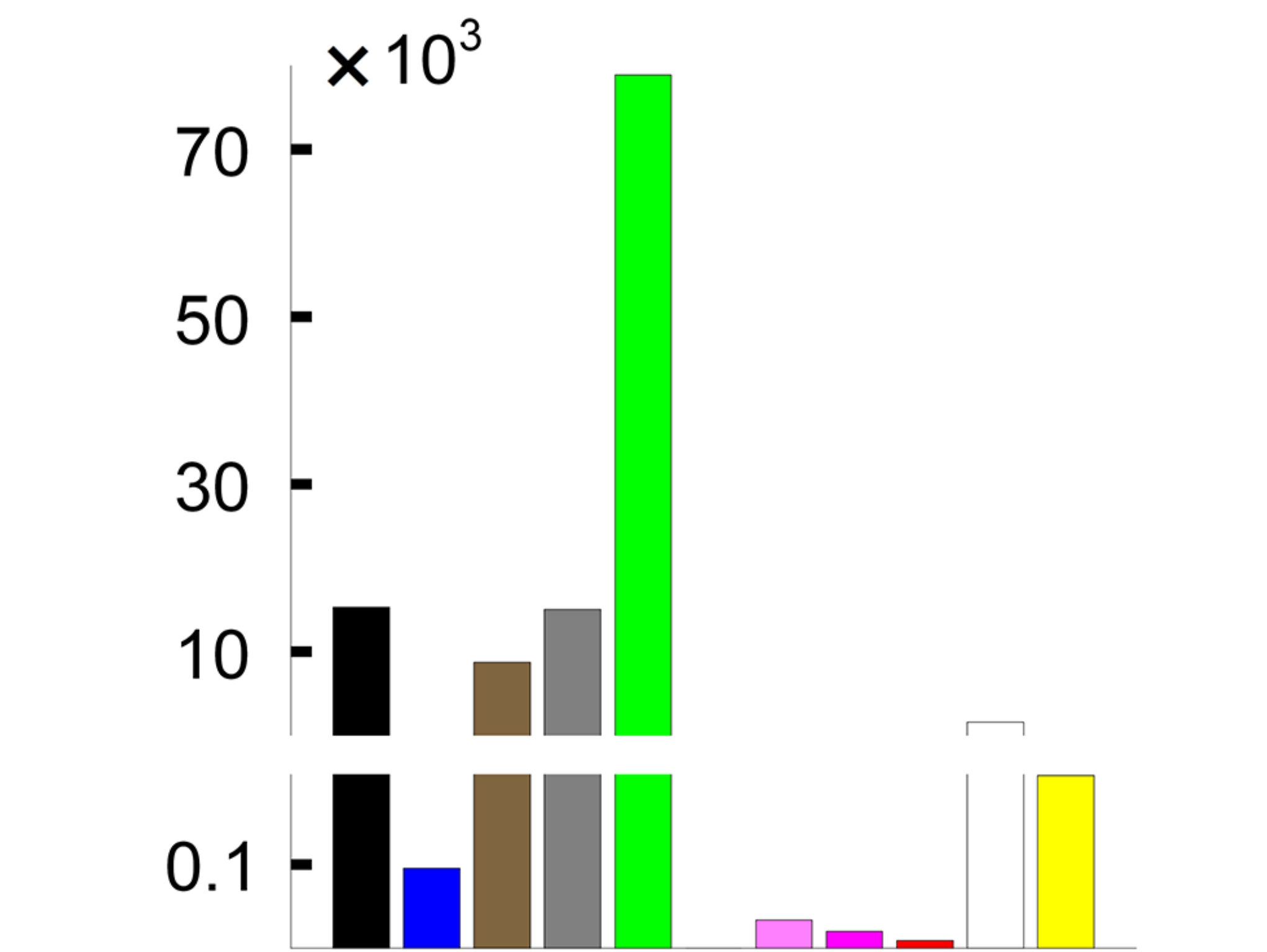}\label{fig-color-name-image-b}}%
	\hfill\null
\caption{Color name image and histogram. (a) Color name image. (b) Color name histogram.}
\label{fig-color-name-image}
\end{figure}

If we use the corresponding RGB value $c_i$ given in Table~\ref{table:color-names} to represent each pixel in $\mathbf{M}$, we get the color name image as shown in Fig.~\ref{fig-color-name-image-a}. The histogram of the color name image has totally 11 color levels, where the $i$th level corresponds to the number of pixels having the color name $t_i$. In this paper, the histogram is called \textit{Color Name Histogram}, as shown in Fig.~\ref{fig-color-name-image-b}. From the color name histogram, we can obtain 11 probability values. The probability of the $i$th color name is denoted as $f_i$.

Another cue is the distributions of the color names in $\mathbf{M}$. For the purpose of combining with 11 master attention maps, we use Eq.~\eqref{eq:M_i} to construct 11 index matrices. In the $i$th index matrix $M_i$, any element value equal to $i$ is set to 1, otherwise is set to 0:
\begin{equation}
\label{eq:M_i}
M_i(x,y) = 
\begin{cases}
	1 \,, & \textrm{if}~~\textbf{\textbf{M}}(x,y) = i  \,; \\
	0 \,, & \textrm{otherwise} \,.
\end{cases}
\end{equation}

As discussed in Section~\ref{sec:attention-map}, the attention map $A_i$ of the $i$th color name channel is computed by linearly averaging boolean maps, where all the foreground regions that connected to the image borders are abandoned. For any boolean map, all the pixels in the surrounded regions share the same weight. To jointly consider the frequencies and distributions of different color names, we simply combine $f_i$ and $M_i$ to obtain the first kind of weights, i.e., weighting matrices
\begin{equation}
W_i = f_i M_i\,.
\end{equation}

\subsubsection{Color Name Contrast Based Weights}
Mainly inspired by~\cite{AcmMM2006/Zhai} and \cite{CVPR2011/Cheng}, we calculate the second kind of weights, i.e., the contrast based weighting coefficients. The weight of each color name is defined as its color contrast to all the other color names. All the pixels having the same color name share the same weight. For the color distance metric, we directly use the RGB values of 11 color names given in Table~\ref{table:color-names}. Specifically, the weighting coefficient $w_i$ of the color name $t_i$ is defined as
\begin{equation} 
w_i = \sum\limits_{j=1}^{11}{f_j \left\| c_i-c_j \right\|_2^2} \,,
\end{equation}
where $\left\| c_i-c_j \right\|_2$ is the $\ell_2$-norm of the color difference between the color names $t_i$ and $t_j$.

\begin{figure}[t]
	\centering
	\hfill%
	\subfloat[]{\includegraphics[width=0.24\linewidth]{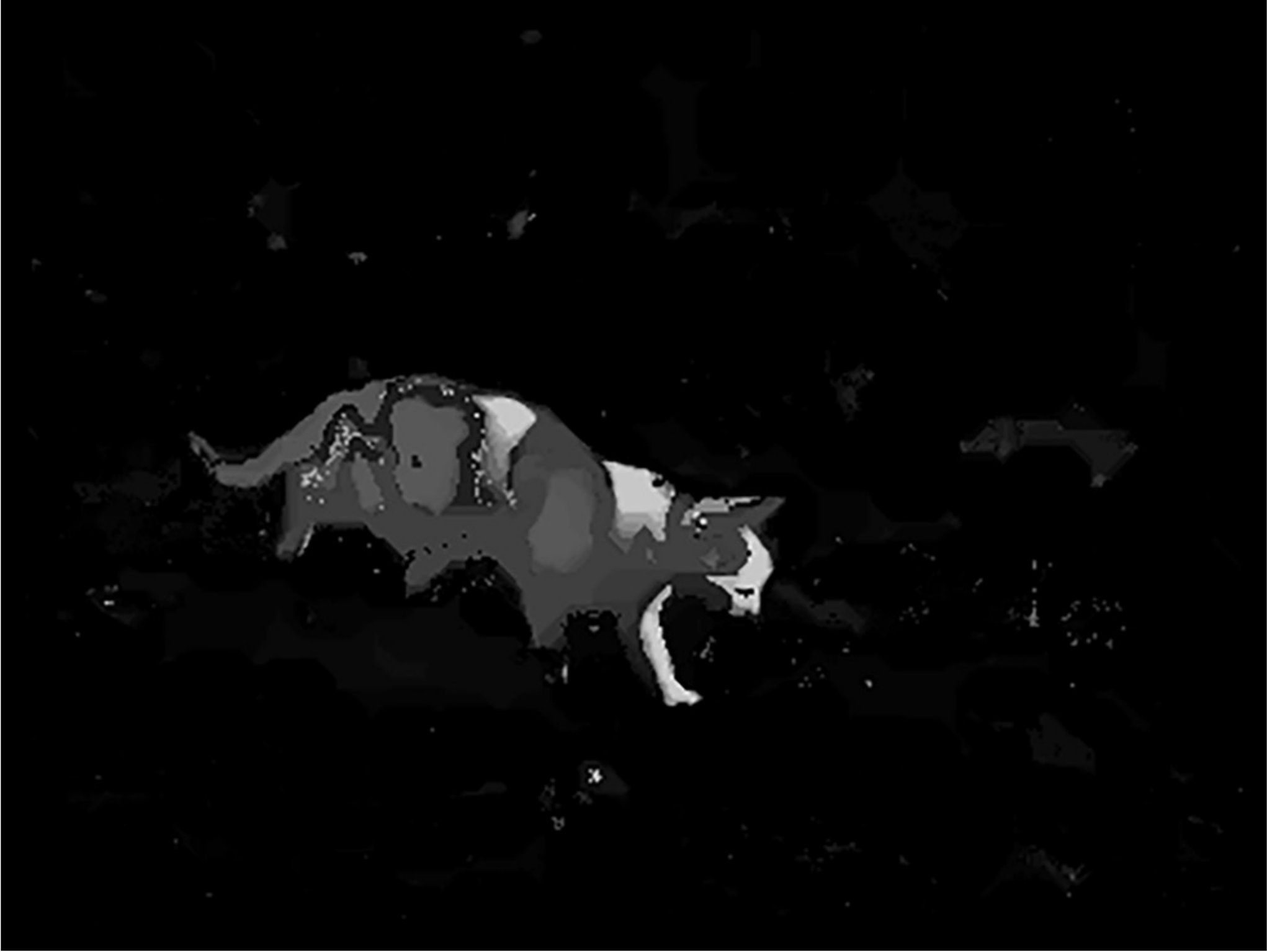}\label{fig-weighted-saliency-map-a}}\hspace{1pt}%
	\subfloat[]{\includegraphics[width=0.24\linewidth]{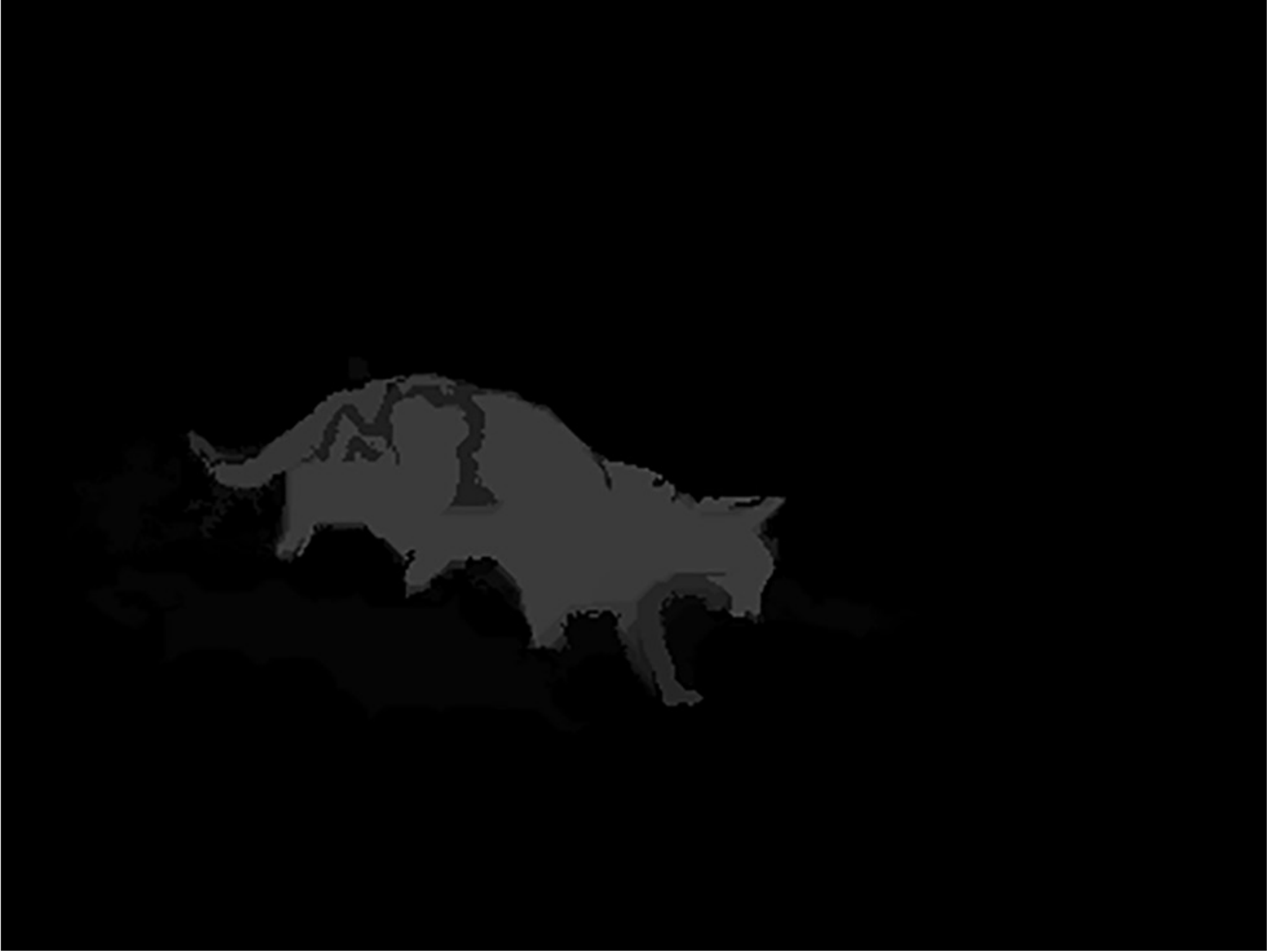}\label{fig-weighted-saliency-map-b}}\hspace{1pt}%
	\subfloat[]{\includegraphics[width=0.24\linewidth]{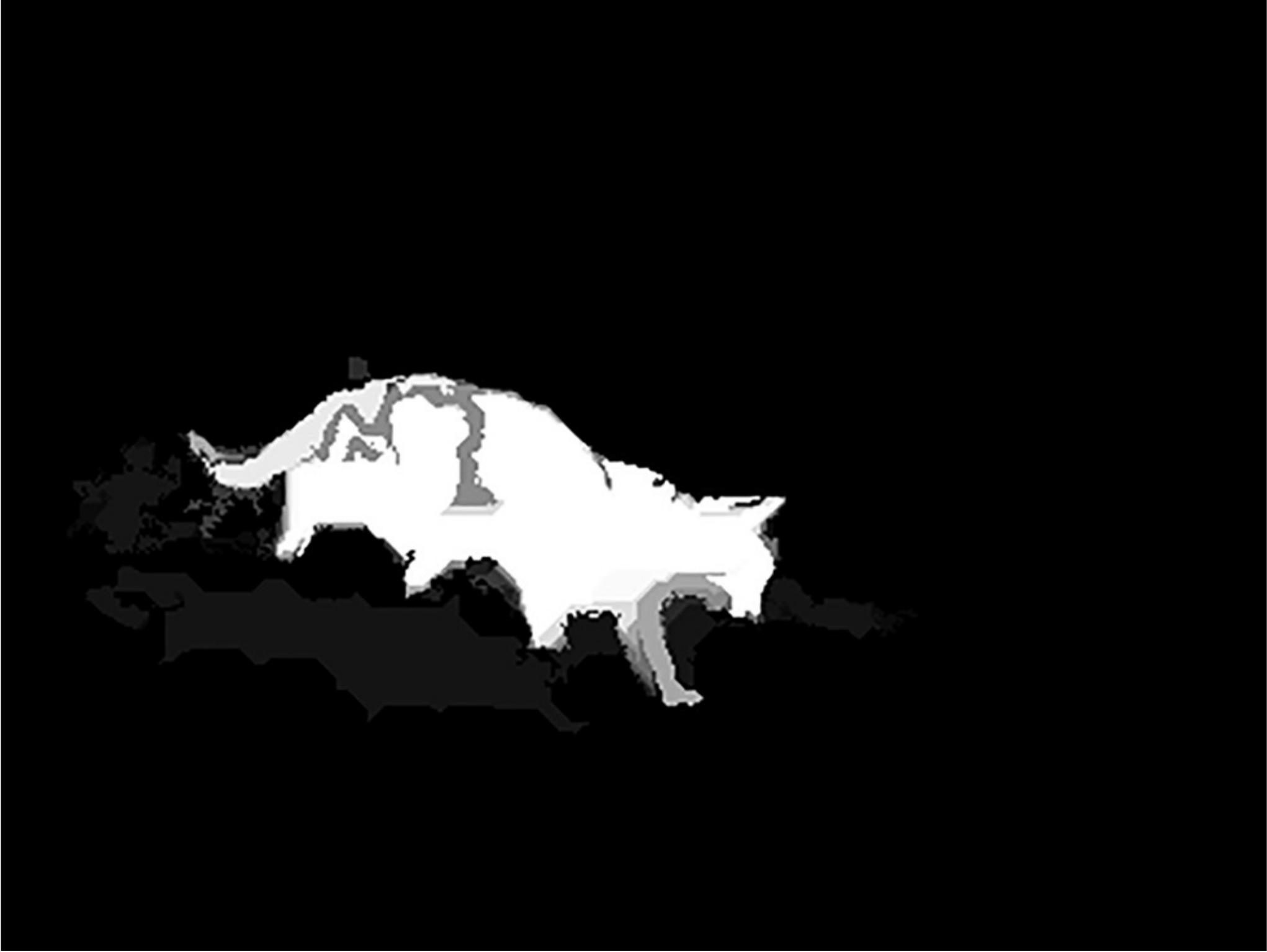}\label{fig-weighted-saliency-map-c}}\hspace{1pt}%
	\subfloat[]{\includegraphics[width=0.24\linewidth]{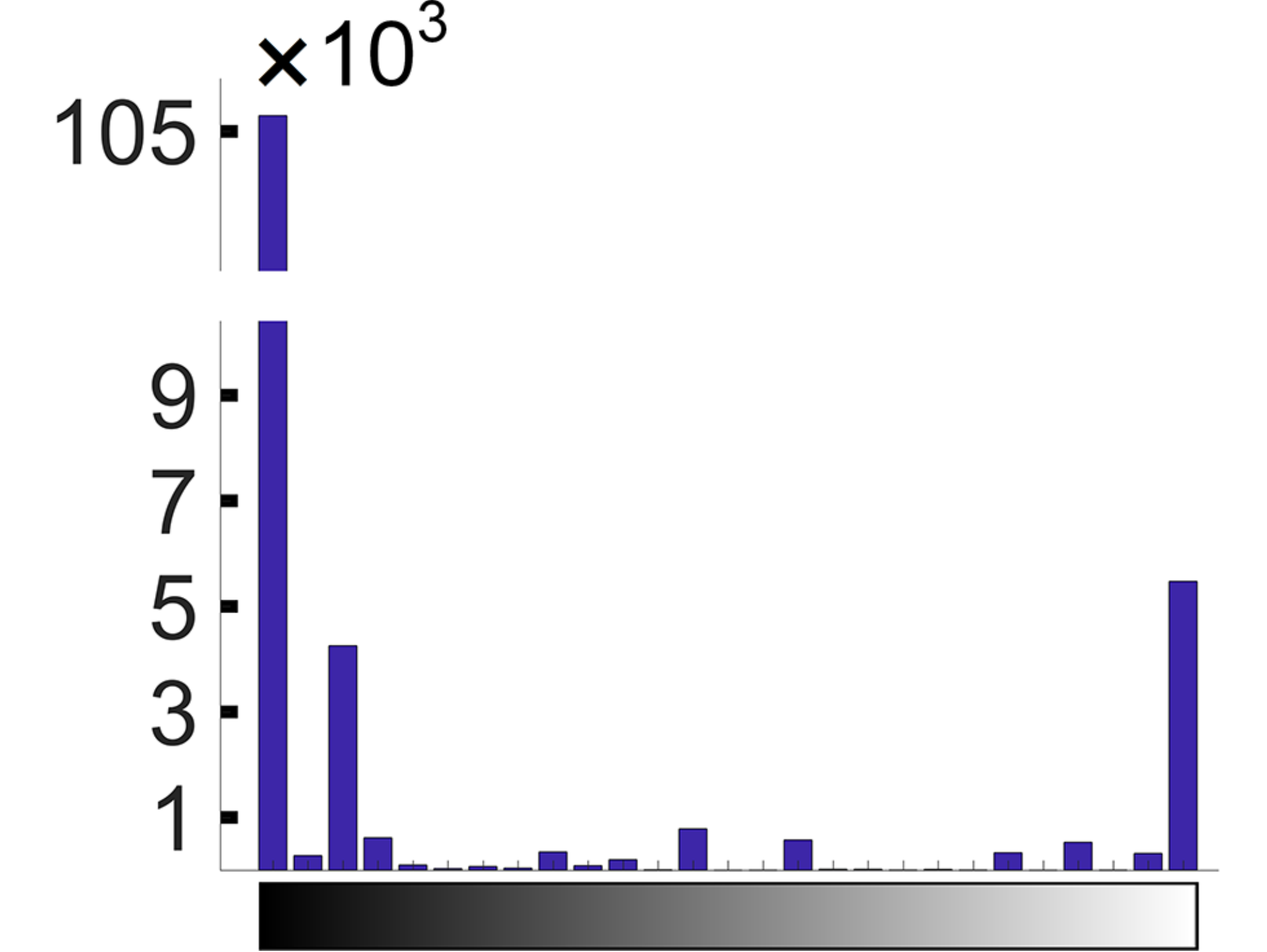}\label{fig-weighted-saliency-map-d}}%
	\hfill\null
	\vfill
	\hfill%
	\subfloat[]{\includegraphics[width=0.24\linewidth]{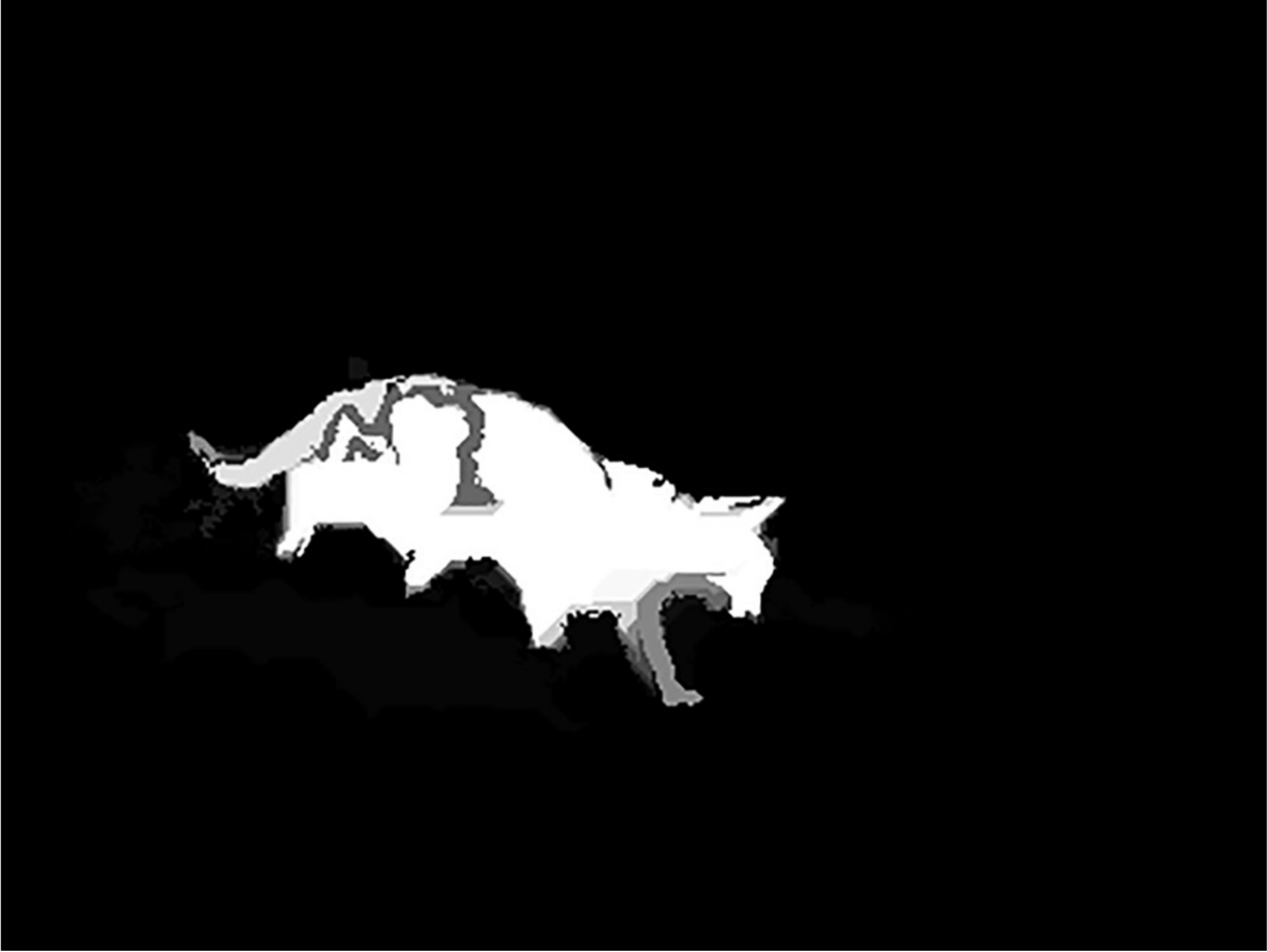}\label{fig-weighted-saliency-map-e}}\hspace{1pt}%
	\subfloat[]{\includegraphics[width=0.24\linewidth]{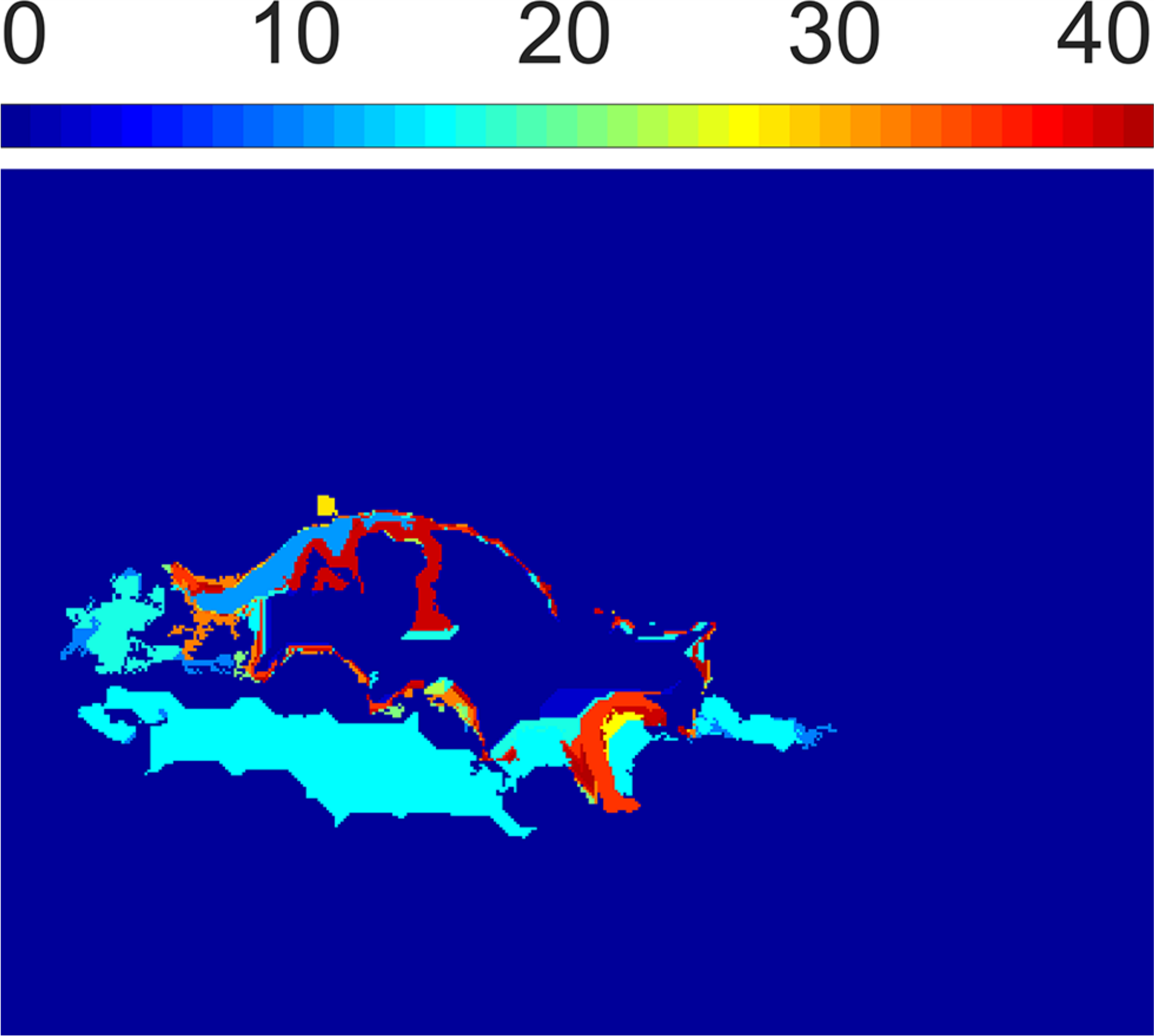}\label{fig-weighted-saliency-map-f}}\hspace{1pt}%
	\subfloat[]{\includegraphics[width=0.24\linewidth]{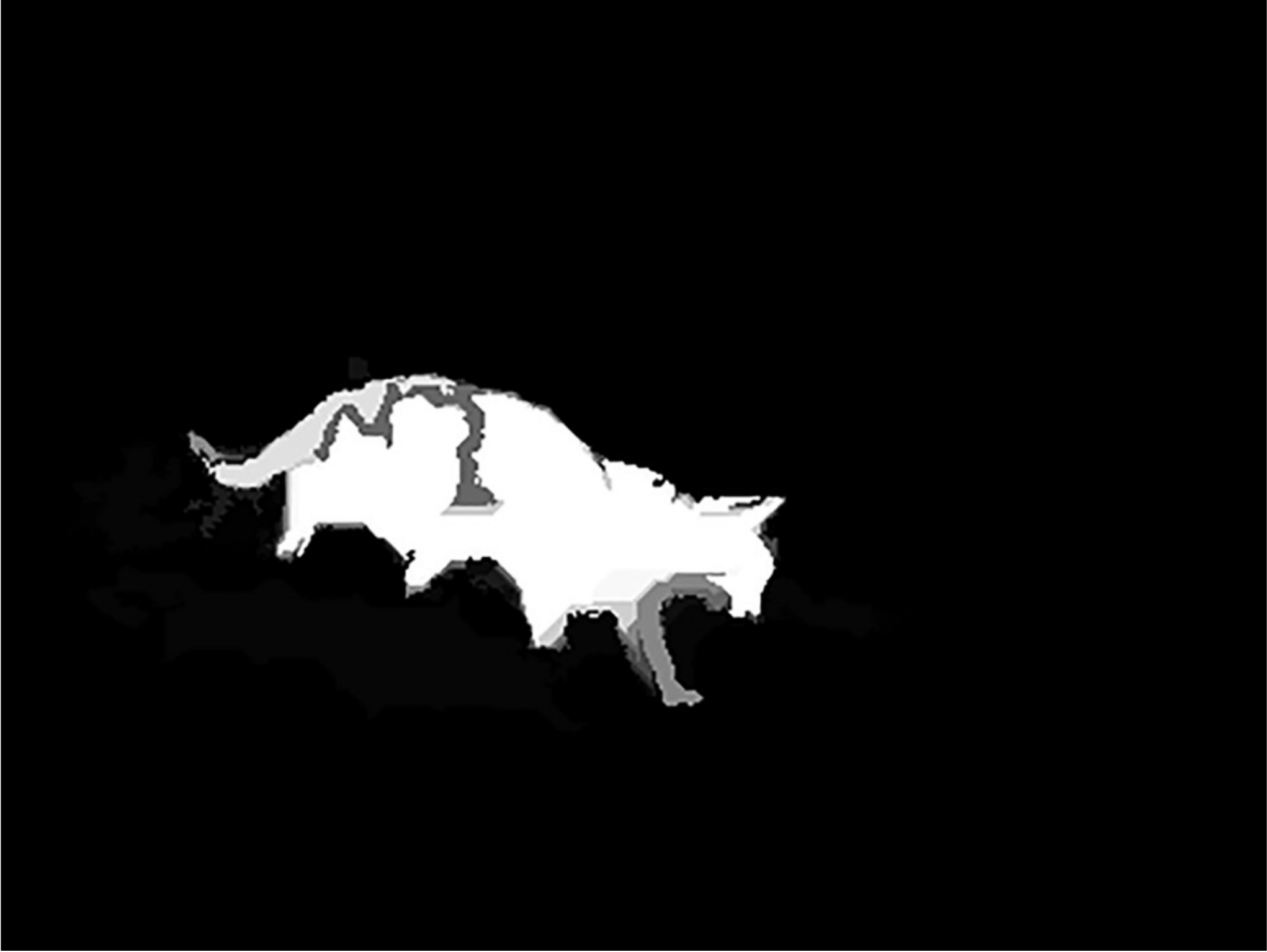}\label{fig-weighted-saliency-map-g}}\hspace{1pt}%
	\subfloat[]{\includegraphics[width=0.24\linewidth]{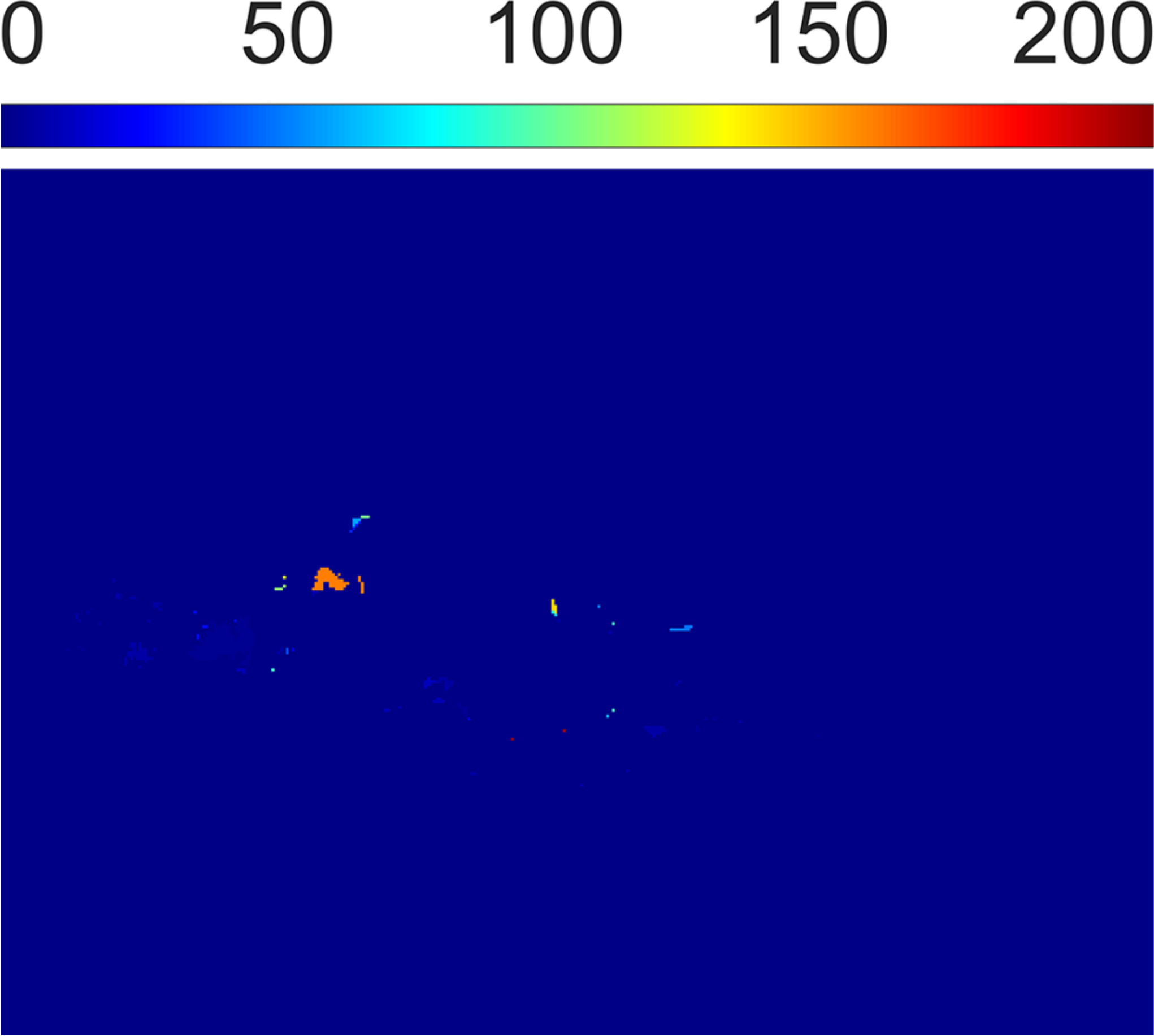}\label{fig-weighted-saliency-map-h}}%
	\hfill\null
\caption{Applying post-processing I to $\bar{A}_w$. (a) Weighted mean attention map $\bar{A}_w$. (b) Morphological reconstruction. (c) Normalization result, and (d) its histogram. (e) Intensity mapping result. (f) Difference between (c) and (e). (g) Weighted saliency map $S_w$. (h) Difference between (e) and (g) for the demonstration of hole-filling.}
\label{fig-weighted-saliency-map}
\end{figure}

By integrating two kinds of weights into 11 master attention maps and averaging them, we compute the weighted mean attention map $\bar{A}_w$ (see Fig.~\ref{fig-weighted-saliency-map-a}) as follows:
\begin{equation}
\bar{A}_w=\sum\limits_{i=1}^{11}{{w_i} \cdot {\texttt{\textbf{N}}}{\left(W_i \circ A_i\right)}} \,,
\end{equation}
where $\circ$ denotes the Hadamard product, and ${\texttt{\textbf{N}}}\left(\cdot\right)$ is the normalization function which sets the values in $\bar{A}_w$ to $[0,1]$. 

Figures~\ref{fig-weighted-saliency-map-b}--\ref{fig-weighted-saliency-map-e} illustrate the same post-processing procedure introduced in Section~\ref{sec:post-processing}. From Fig.~\ref{fig-weighted-saliency-map-h}, we can see that the hole-filling operation completes the closed dark regions inside the cat. Finally, we obtain the second saliency map, i.e., the weighted saliency map $S_w$ with the range $[0,255]$, as shown in Fig.~\ref{fig-weighted-saliency-map-g}.

\subsection{Combination}
\label{sec:combination}
\begin{figure}[t]
	\centering
	\hfill%
	\subfloat[]{\includegraphics[width=0.24\linewidth]{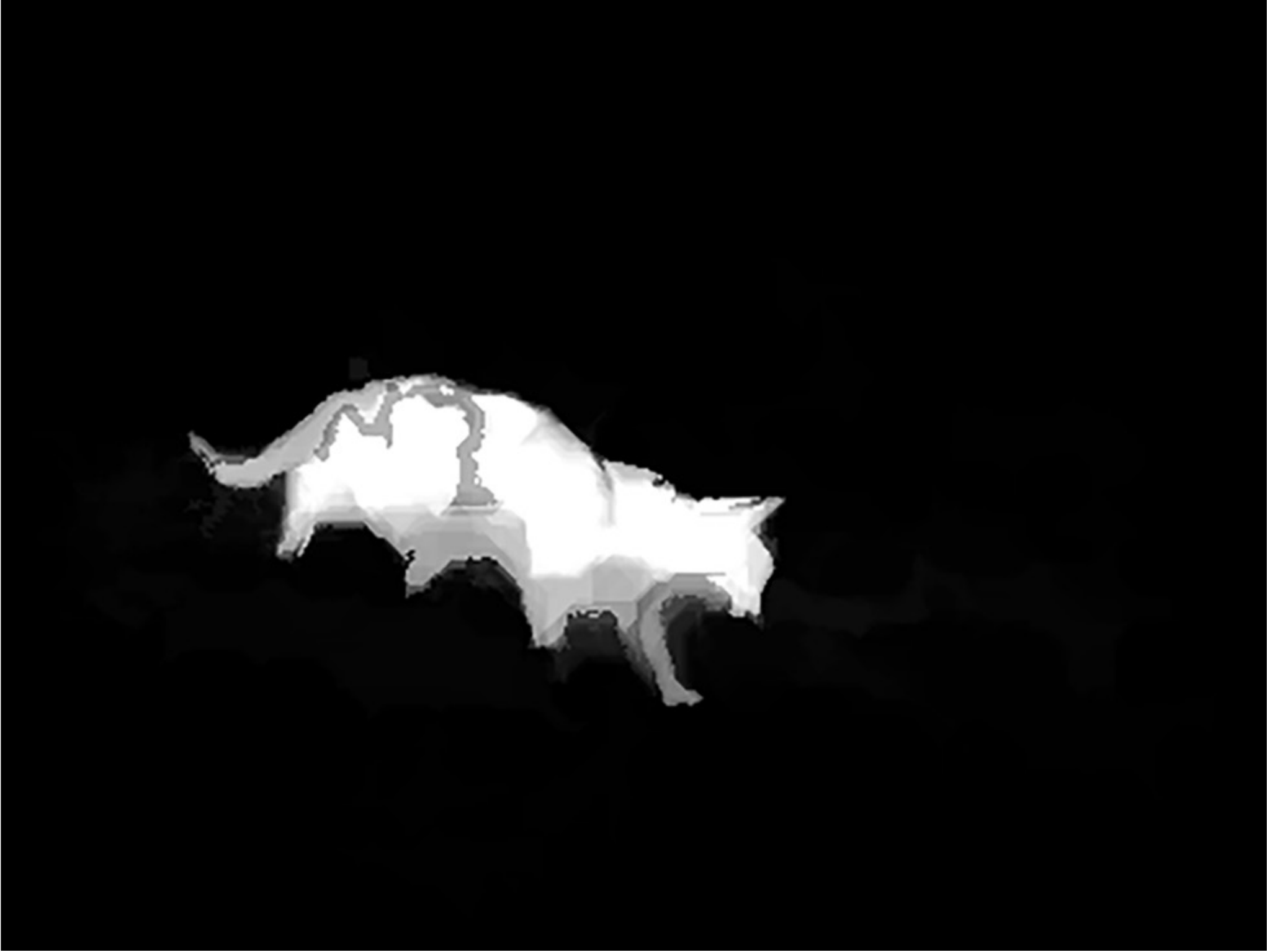}\label{fig-combination-a}}\hfill%
	\subfloat[]{\includegraphics[width=0.24\linewidth]{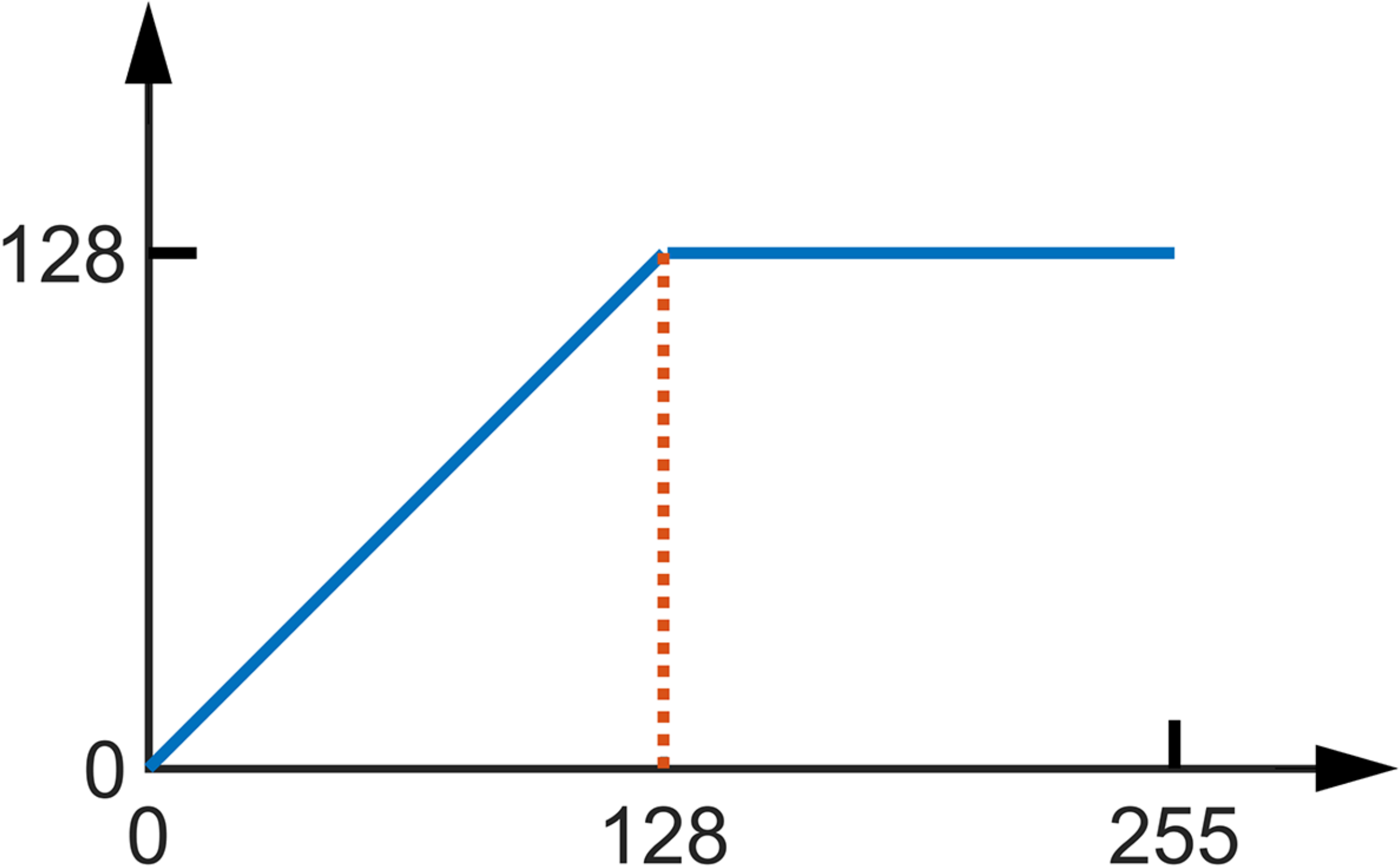}\label{fig-combination-b}}\hfill%
	\subfloat[]{\includegraphics[width=0.24\linewidth]{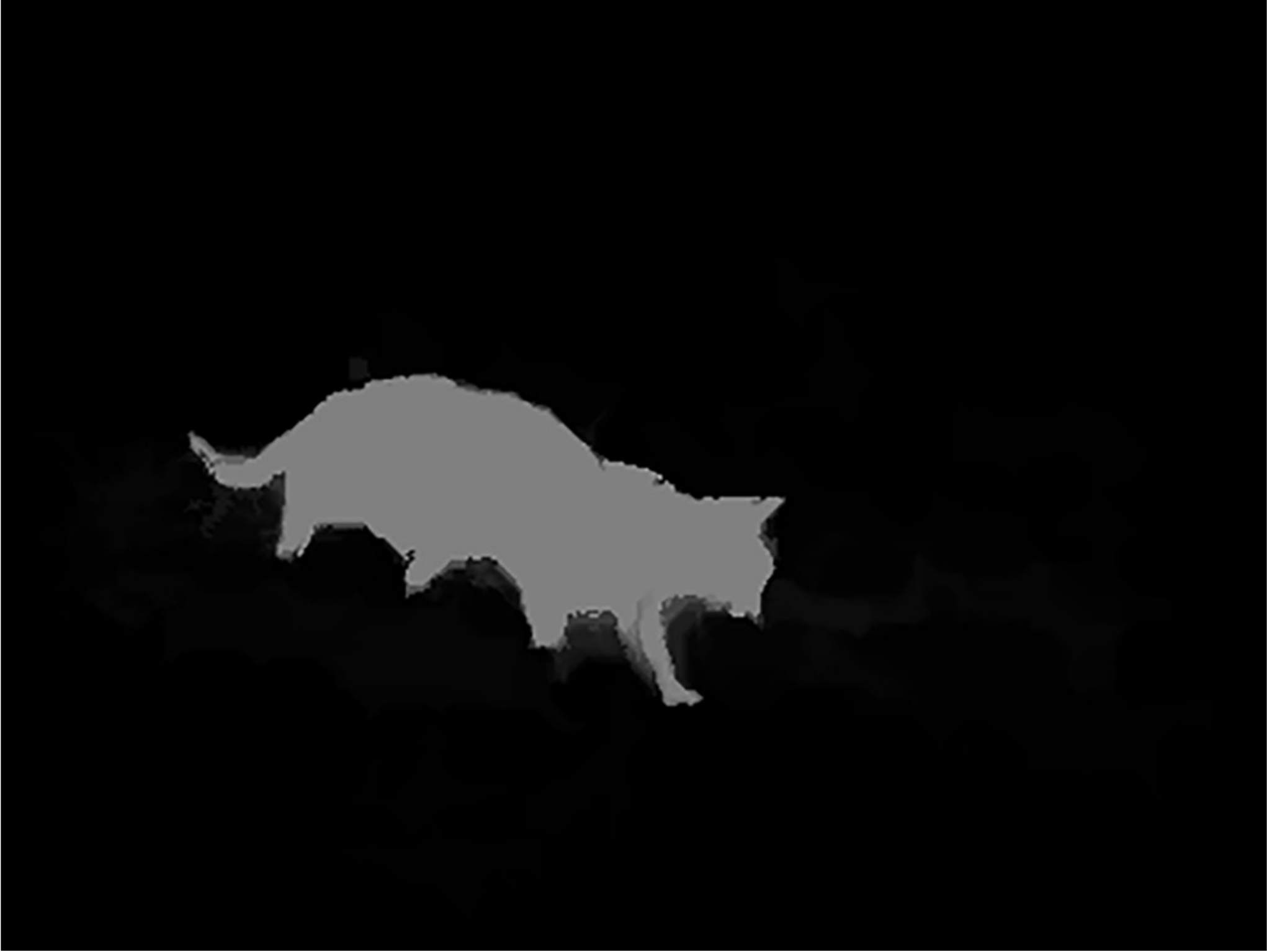}\label{fig-combination-c}}%
	\hfill\null
\caption{Combination. (a) Original output of averaging Figs.~\ref{fig-post-processing-I-h} and \ref{fig-weighted-saliency-map-g}. (b) Truncation curve. (c) Mean saliency map $\bar{S}$.}
\label{fig-combination}
\end{figure}

To couple with the saliency maps $S$ and $S_w$, we simply average them at the first step of the combination stage. The original output is illustrated in Fig.~\ref{fig-combination-a}. Considering that the use of saliency maps is to assist in salient object segmentation, the original combination result is obviously not ideal. First, for the purpose of eliminating the perceptually insignificant regions outside the cat, we perform an intensity mapping in the post-processing I, which simultaneously suppresses the inner saliency and subsequently results in an indeterminate object region in $S$. Second, in $S_w$ the salient object has a clear contour, but apparently shows a nonuniform intensity distribution. Third, the locations of the regions with higher saliency values are completely different between two saliency maps.

To address the above issues, a truncation operation is introduced to clip the original output. Intuitively, we wish the resultant salient object to have a uniform intensity distribution, which can be further highlighted by using a post-processing procedure. Since both $S$ and $S_w$ have been normalized to the range $[0,255]$, we define the improved mean saliency map $\bar{S}$ as
\begin{equation}
\bar{S} = \frac{\left[ S+S_w \right]_{0}^{255}}{2} \,,
\end{equation}
where $[\,\cdot\,]_{0}^{255}$ is the operator for truncating the inner to have values between 0 and 255.

As illustrated in Fig.~\ref{fig-combination-b}, the above definition causes a piecewise mapping, in which the values above 128 are clipped and the others stay unchanged. From Fig.~\ref{fig-combination-c}, we can see the resultant map $\bar{S}$ occupies the common salient parts between $S$ and $S_w$. Although the detected region has lower saliency, the whole region is uniform and clearly stands out of the background. This means that we also can perform a post-processing operation on $\bar{S}$ to refine its saliency.

\renewcommand{\algorithmicrequire}{\textbf{Input:}}
\renewcommand{\algorithmicensure}{\textbf{Output:}}
\begin{algorithm}[b]
\caption{post-processing II}
\label{algo-post-processing-II}
\begin{algorithmic}[1]
\Require mean saliency map $\bar{S}$
\Ensure final result $\widehat{S}$
\State{$\widehat{S} = {\texttt{\textbf{MAP}}}\,(\bar{S}, \vartheta_r, \vartheta_g)$}
\State{$\widehat{S} = {\texttt{\textbf{FILL}}}\,(\widehat{S})$}
\State{$\widehat{S} = {\texttt{\textbf{RESIZE}}}\,(\widehat{S})$}
\end{algorithmic}
\end{algorithm}

\begin{figure}[t]
	\centering
	\hfill%
	\subfloat[]{\includegraphics[width=0.24\linewidth]{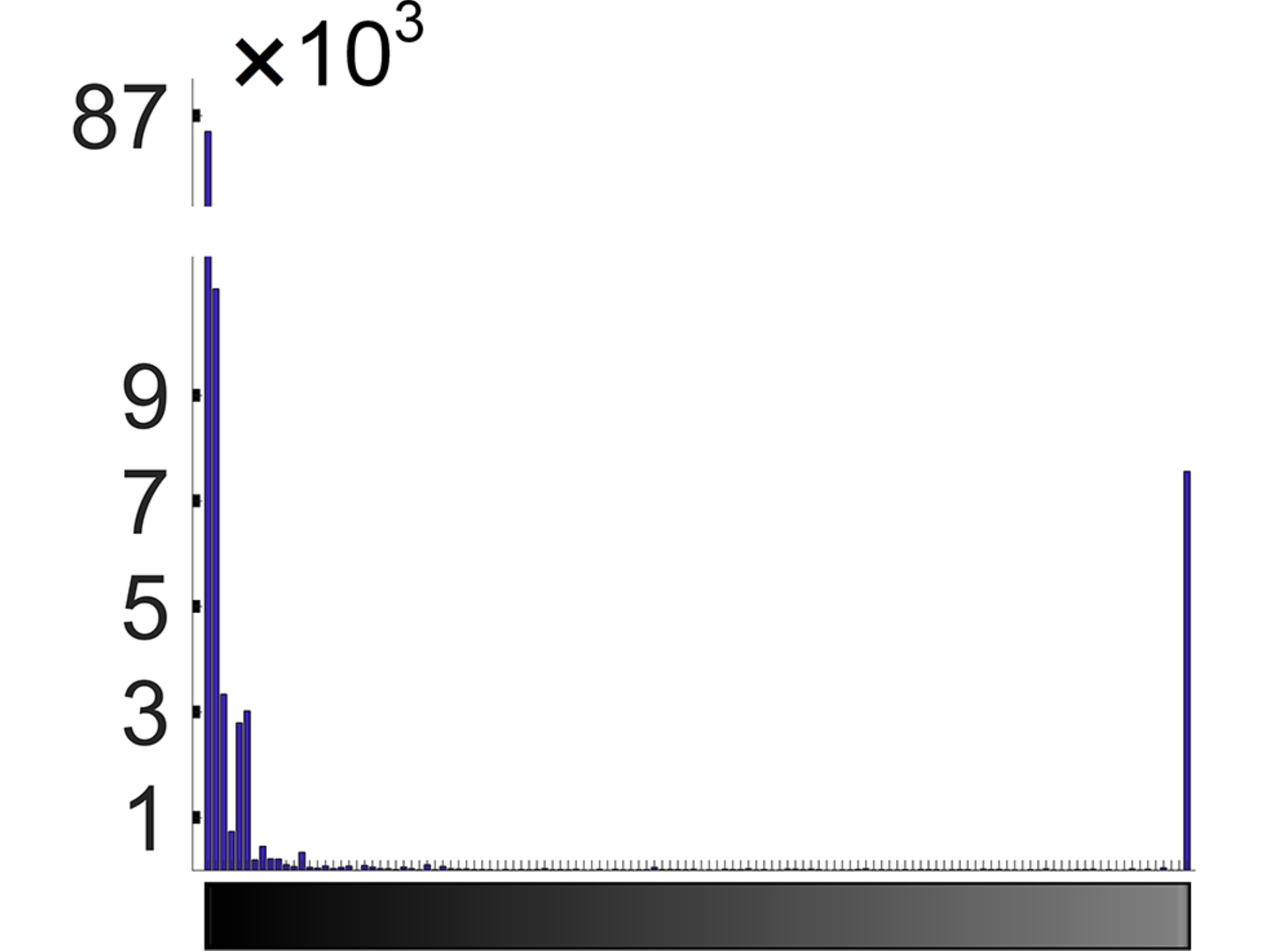}\label{fig-post-processing-II-a}}\hfill%
	\subfloat[]{\includegraphics[width=0.24\linewidth]{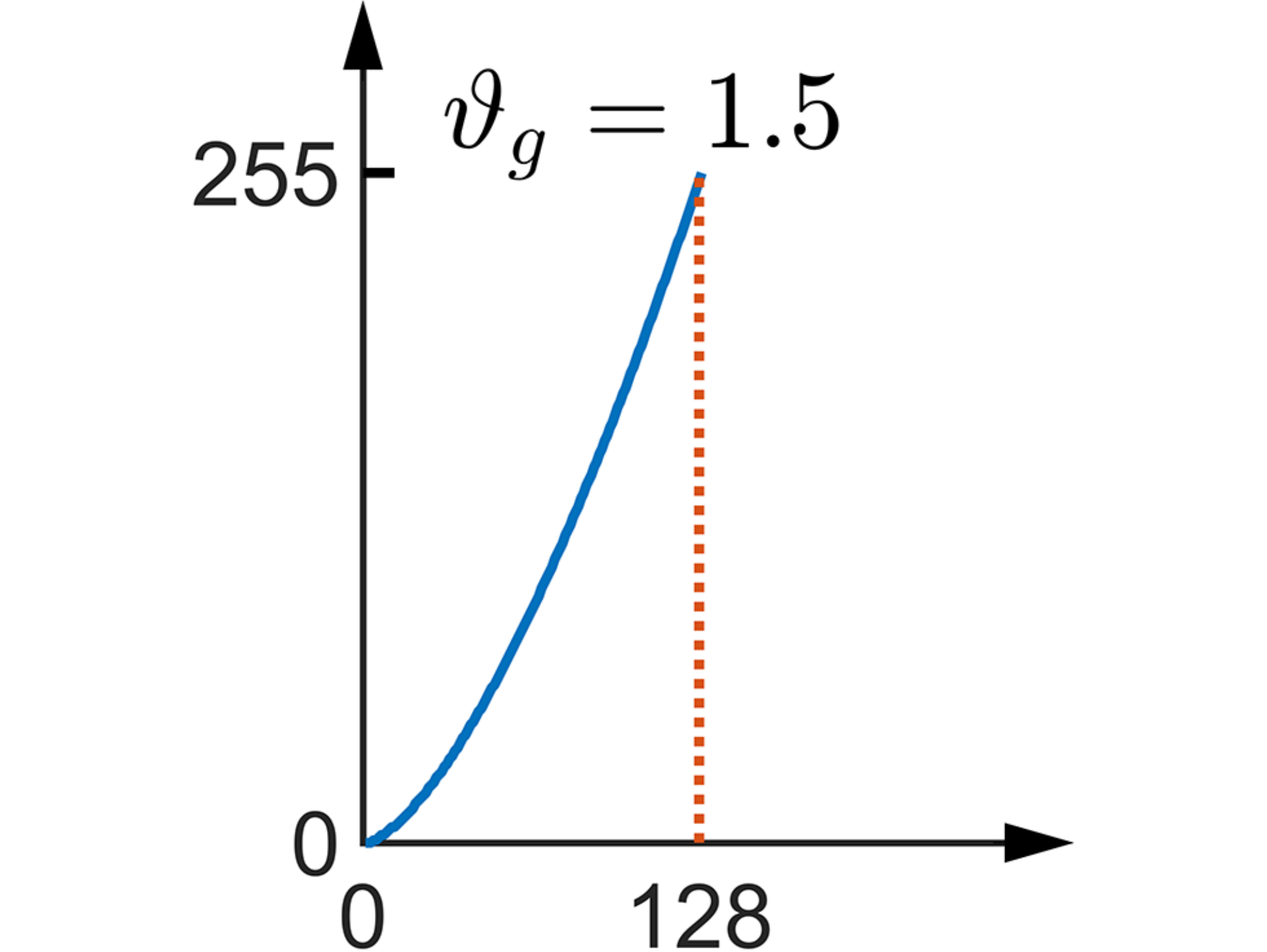}\label{fig-post-processing-II-b}}\hfill%
	\subfloat[]{\includegraphics[width=0.24\linewidth]{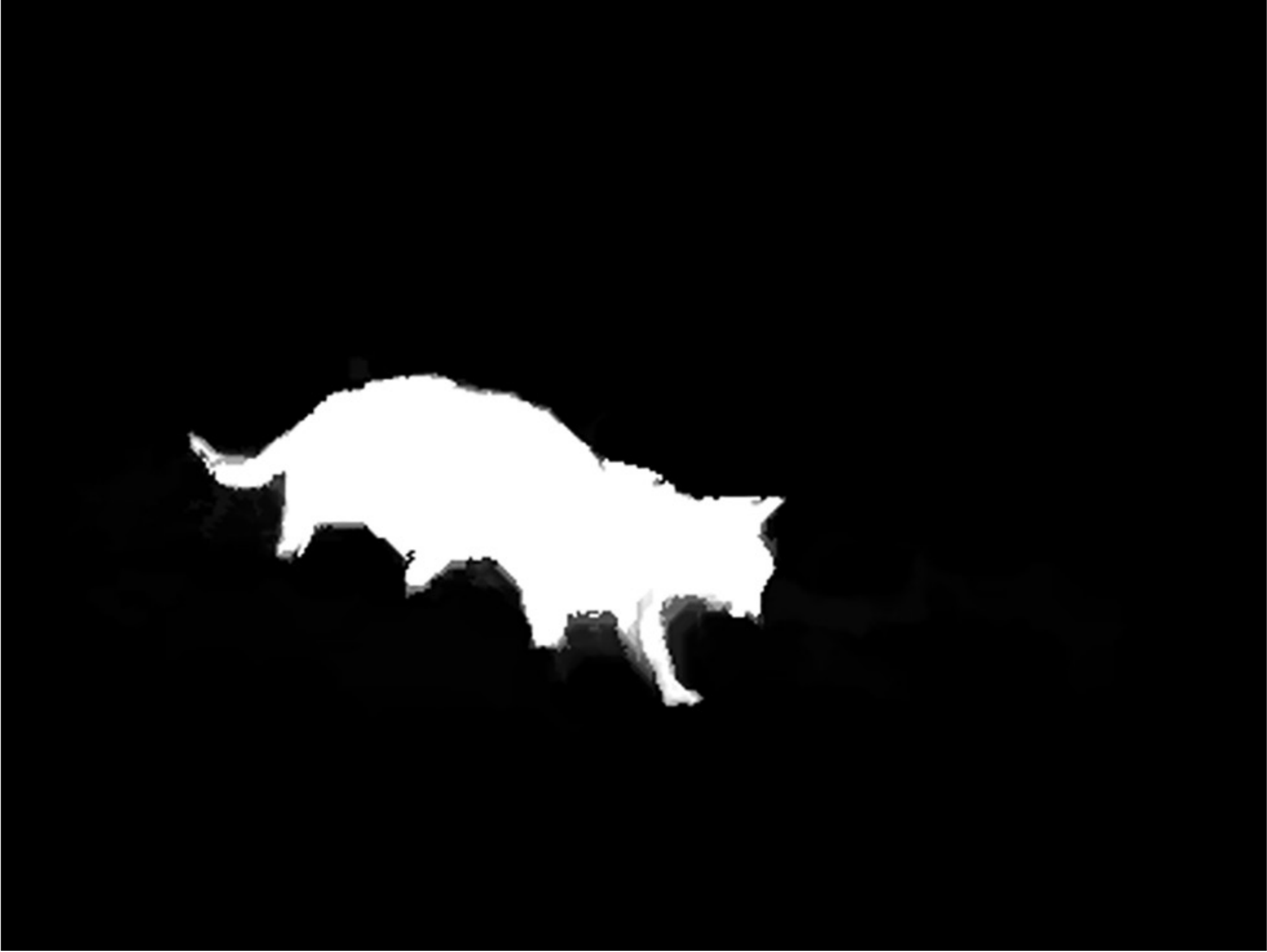}\label{fig-post-processing-II-c}}%
	\hfill\null
	\vspace{-2pt}
\caption{Post-processing II. (a) Histogram of Fig.~\ref{fig-combination-c}. (b) Intensity mapping curve. (c) Final result $\widehat{S}$.}
\label{fig-post-processing-II}
\end{figure}

A new post-processing procedure is summarized in Algorithm~\ref{algo-post-processing-II}. Compared with Algorithm~\ref{algo-post-processing-I}, the difference is that this procedure only includes two operations: intensity mapping and hole-filling. For the former operation, we use the same parameter settings as before. Figure~\ref{fig-post-processing-II-a} shows the histogram of the mean saliency map $\bar{S}$, the intensity mapping curve is illustrated in Fig.~\ref{fig-post-processing-II-b}. Note that different from Fig.~\ref{fig-post-processing-I-e}, here the intensity mapping curve maps the inputs in the range $[0,128]$ to the outputs in the range $[0,255]$. After filling all the small dark holes in the object region, we obtain the final saliency result $\widehat{S}$ of the proposed model, as shown in Fig.~\ref{fig-post-processing-II-c}. It can be seen that our model well suppresses the image background and uniformly highlights the foreground object. More importantly, for the future task of salient object segmentation, we can easily perform a thresholding operation on $\widehat{S}$ while generate more stable segmentation results over a wide range of thresholds.

\section{Experiments}
\label{sec:experiments}
We evaluate the proposed model with twenty-one saliency models including AC~\cite{ICVS2008/Achanta}, BMS~\cite{ICCV2013/Zhang}, CA~\cite{CVPR2010/Goferman}, COV~\cite{JOV2013/Erdem}, FES~\cite{SCIA2011/Tavakoli}, FT~\cite{CVPR2009/Achanta}, GC~\cite{ICCV2013/Cheng}, GU~\cite{ICCV2013/Cheng}, HC~\cite{CVPR2011/Cheng}, HFT~\cite{TPAMI2013/Li}, MSS~\cite{ICIP2010/Achanta}, PCA~\cite{CVPR2013/Margolin}, RC~\cite{CVPR2011/Cheng}, RPC~\cite{PONE2014/JingLou}, SEG~\cite{ECCV2010/Rahtu}, SeR~\cite{JOV2009/Seo}, SIM~\cite{CVPR2011/Murray}, SR~\cite{CVPR2007/Hou}, SUN~\cite{JOV2008/Zhang}, SWD~\cite{CVPR2011/Duan}, and TLLT~\cite{CVPR2015/Gong} on three benchmark data sets: ASD~\cite{CVPR2007/Liu,CVPR2009/Achanta}, ECSSD~\cite{CVPR2013/Yan,TPAMI2016/Shi}, and ImgSal~\cite{BMVC2011/Li,TPAMI2013/Li}. The used saliency maps of the above models are from:

\begin{itemize}
\item For BMS,\footnote{\url{http://cs-people.bu.edu/jmzhang/BMS/BMS.html}} HFT,\footnote{\url{http://www.escience.cn/people/jianli/DataBase.html}\label{HFT}} HS,\footnote{\url{http://www.cse.cuhk.edu.hk/leojia/projects/hsaliency/}} RPC,\footnote{\url{http://www.loujing.com/rpc-saliency/}} and TLLT\footnote{\url{http://www.escience.cn/people/chengong/Codes.html}} over all the data sets, we use the author-provided saliency maps, or run the authors' codes to obtain saliency maps.

\item For the AC, CA, FT, HC, RC, and SR models on the ASD data set, we directly use the saliency maps provided by Cheng et al.~\cite{CVPR2011/Cheng}.\footnote{\url{http://cg.cs.tsinghua.edu.cn/people/~cmm/Saliency/Index.htm}} For the remainder models on ASD, we retrieve the related saliency maps from the MSRA10K database~\cite{TPAMI2015/Cheng}.\footnote{\url{http://mmcheng.net/msra10k/}}

\item For the remainder models, we employ the implementation of the salient object detection benchmark published by Borji et al.~\cite{TIP2015/Borji-Benchmark}.\footnote{\url{http://mmcheng.net/salobjbenchmark/}} On the ECSSD data set, the saliency maps come directly from the author-provided results; on the ImgSal data set, we run the authors' source code to generate saliency maps.
\end{itemize}

\subsection{Data sets}
The popular ASD data set (a.k.a, MSRA1000) is a subset of MSRA5000~\cite{CVPR2007/Liu}.\footnote{\url{http://research.microsoft.com/en-us/um/people/jiansun/SalientObject/salient\_object.htm}} The original MSRA5000 data set contains 5000 images with the labeled rectangles from nine participants. Achanta et al.~\cite{CVPR2009/Achanta} consider that the use of saliency maps is for salient object segmentation, then derive ASD with 1000 images from MSRA5000. Instead of the user-drawn rectangles around salient regions used in~\cite{CVPR2007/Liu}, the ASD data set provides the object-contour based ground truth for more accurate comparisons of segmentation results.\footnote{\url{http://ivrl.epfl.ch/supplementary_material/RK_CVPR09/}}

The ECSSD data set is an extension of CSSD.\footnote{\url{http://www.cse.cuhk.edu.hk/leojia/projects/hsaliency/data set.html}} In order to represent more general situations of natural images than ASD, Yan et al. construct the CSSD data set, which contains 200 images with diversified patterns in both foreground and background~\cite{CVPR2013/Yan}. Subsequently, they extend CSSD to a larger data set named ECSSD, which includes 1000 structurally complex images and pixel-wise ground truth masks labeled by five helpers~\cite{TPAMI2016/Shi}.

In addition, we evaluate the proposed model on the ImgSal data set, which is designed for the detection of salient regions of different sizes~\cite{BMVC2011/Li,TPAMI2013/Li}.\textsuperscript{\ref{HFT}} The ImgSal contains 235 images collected using Google, and provides both region ground truth (human labeled) and fixation ground truth (by eye tracker). For region ground truth, the authors ask nineteen naive subjects to label the images in a random manner, and generate two kinds of labeling results for each image: binary map and probability map. In our experiments, we only use the binary maps for evaluating saliency detection results.

\subsection{Experimental Setup}
The common used metrics to evaluate salient object detection models are \textit{Precision-Recall} and \textit{$F_\beta$-measure}. For an input image, the resultant saliency map is a gray-scale image having integer values in the range $[0,255]$. So we can partition it to a binary map $M$ by using a threshold, then compute precision and recall by comparing $M$ with the corresponding ground truth $G$ as follows:
\begin{equation}
Precision=\frac{|M\cap G|}{|M|}\,, ~Recall=\frac{|M\cap G|}{|G|}\,,
\end{equation}
where $\left|\,\cdot\,\right|$ indicates the number of the foreground pixels. Moreover, to jointly evaluate precision and recall, the $F_\beta$-measure can be computed as
\begin{equation}
\label{eq:F-beta}
F_\beta=\frac{(1+\beta^2)\times Precision \times Recall}{\beta^2 \times Precision + Recall}\,,
\end{equation}
where $\beta^2$ is set to 0.3 for emphasizing the precision as suggested in~\cite{CVPR2009/Achanta}.

In our experiments, two binarization ways are used to partition saliency maps.

\textit{1) Fixed Thresholding:} We vary a threshold $T_f$ from 0 to 255 to compute the scores of precision, recall and $F_\beta$-measure. Besides plotting the Precision-Recall and $F_\beta$-measure curves, we report two statistics for quantitative evaluation, i.e., average $F_\beta$ score (denoted ``AvgF'') and maximum $F_\beta$ score (denoted ``MaxF'').

\textit{2) Adaptive Thresholding:} As presented in~\cite{CVPR2009/Achanta}, we use an adaptive threshold $T_a$ (cf. Eq.~\eqref{eq:T_a}) to partition $\widehat{S}$ and compute the scores of precision, recall and $F_\beta$-measure. Besides plotting Precision-Recall bars, we also report the $F_\beta$ score obtained by using $T_a$ (denoted ``AdpF''):
\begin{equation}
\label{eq:T_a}
T_a = \frac{2}{W \times H}\sum\limits_{x=1}^{W}{\sum\limits_{y=1}^{H}{\widehat{S}(x,y)}} \,,
\end{equation}
where $W$ and $H$ are the width and height of $\widehat{S}$ respectively, $\widehat{S}(x,y)$ is the saliency value at the coordinate $(x,y)$.

\subsection{Parameter Analysis}
\label{sec:parameter-analysis}
The proposed model includes five parameters: sample step $\delta$, kernel radius $\omega_c$ of closing operation, kernel radius $\omega_r$ of morphological reconstruction, saturation ratio $\vartheta_r$ and gamma $\vartheta_g$ of intensity mapping. To find the optimal parameter setting, we exploit the ``MaxF'' metric suggested in~\cite{TPAMI2004/Martin} to compare the saliency maps obtained using different parameter settings. After 256 $F_\beta$ scores have been computed by fixed thresholding, the maximum one is chosen as the best score for each group of parameter setting. In our experiments, the ranges of five parameters are: $\delta \in [4:4:40]$, $\omega_c \in [1:20]$, $\omega_r \in [1:20]$, $\vartheta_r \in [0.001:0.001:0.009] \cup [0.01:0.01:0.1]$, and $\vartheta_g \in [1.0:0.1:3.0]$, respectively.

\begin{figure}[b]
	\hfill%
	\includegraphics[width=0.47\linewidth]{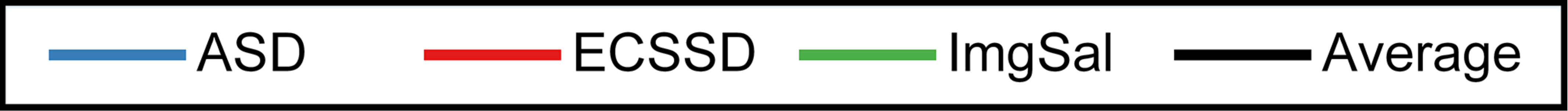}%
	\hfill\null
	\vfill
	\hfill%
	\subfloat[$\delta$]{\includegraphics[width=0.33\linewidth]{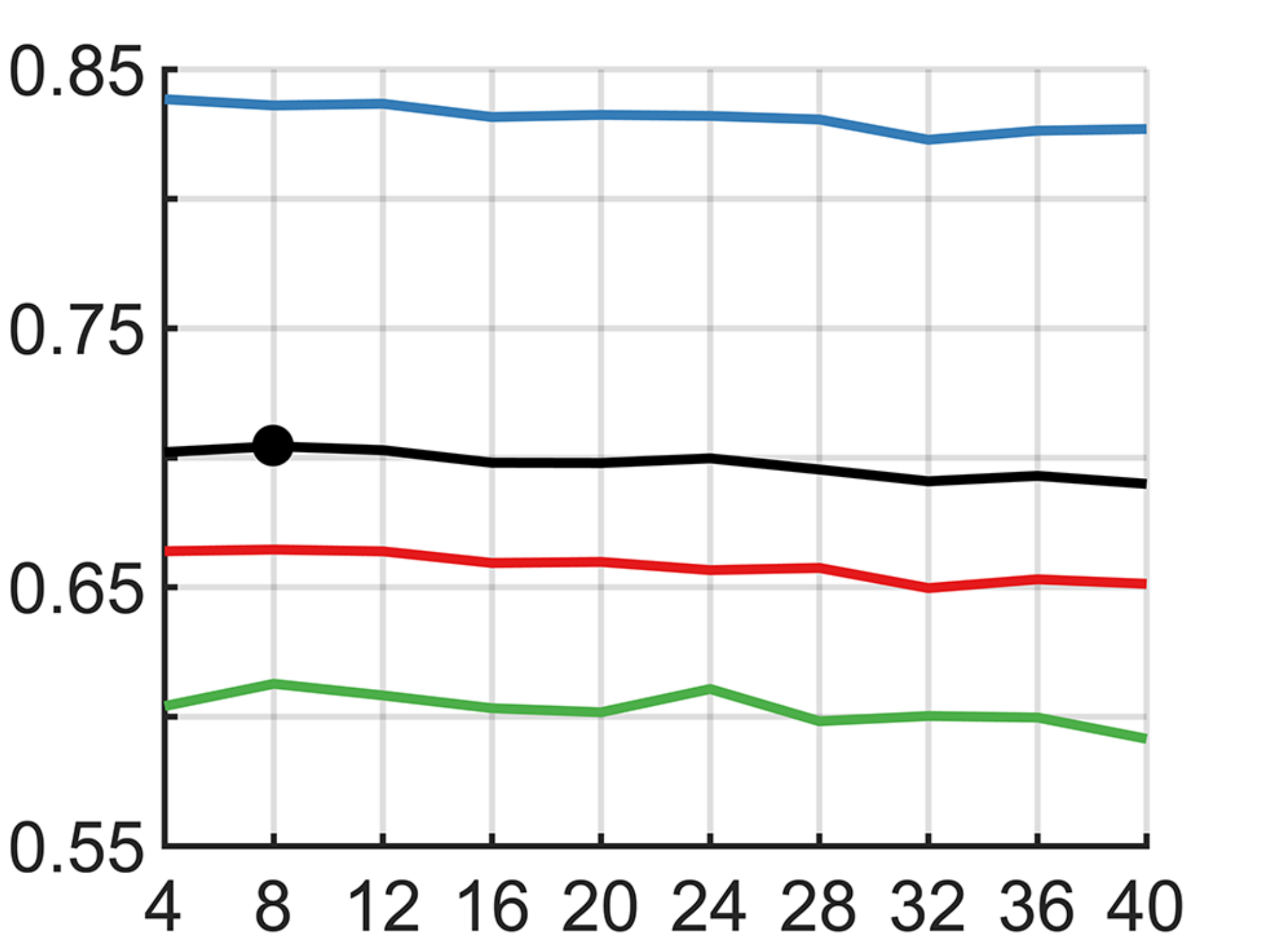}\label{fig-parameter-analysis-a}}\hfill%
	\subfloat[$\omega_c$]{\includegraphics[width=0.33\linewidth]{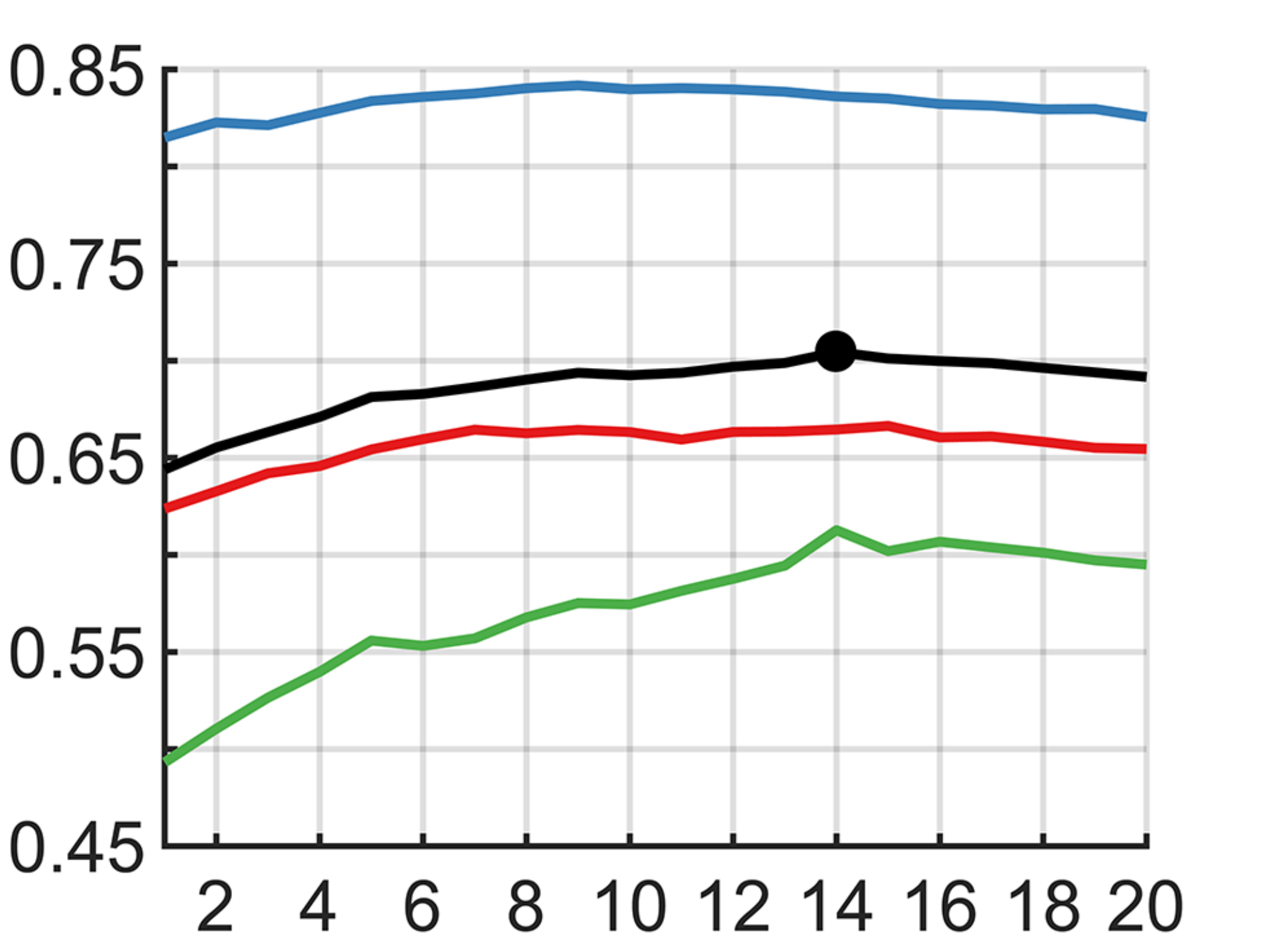}\label{fig-parameter-analysis-b}}\hfill%
	\subfloat[$\omega_r$]{\includegraphics[width=0.33\linewidth]{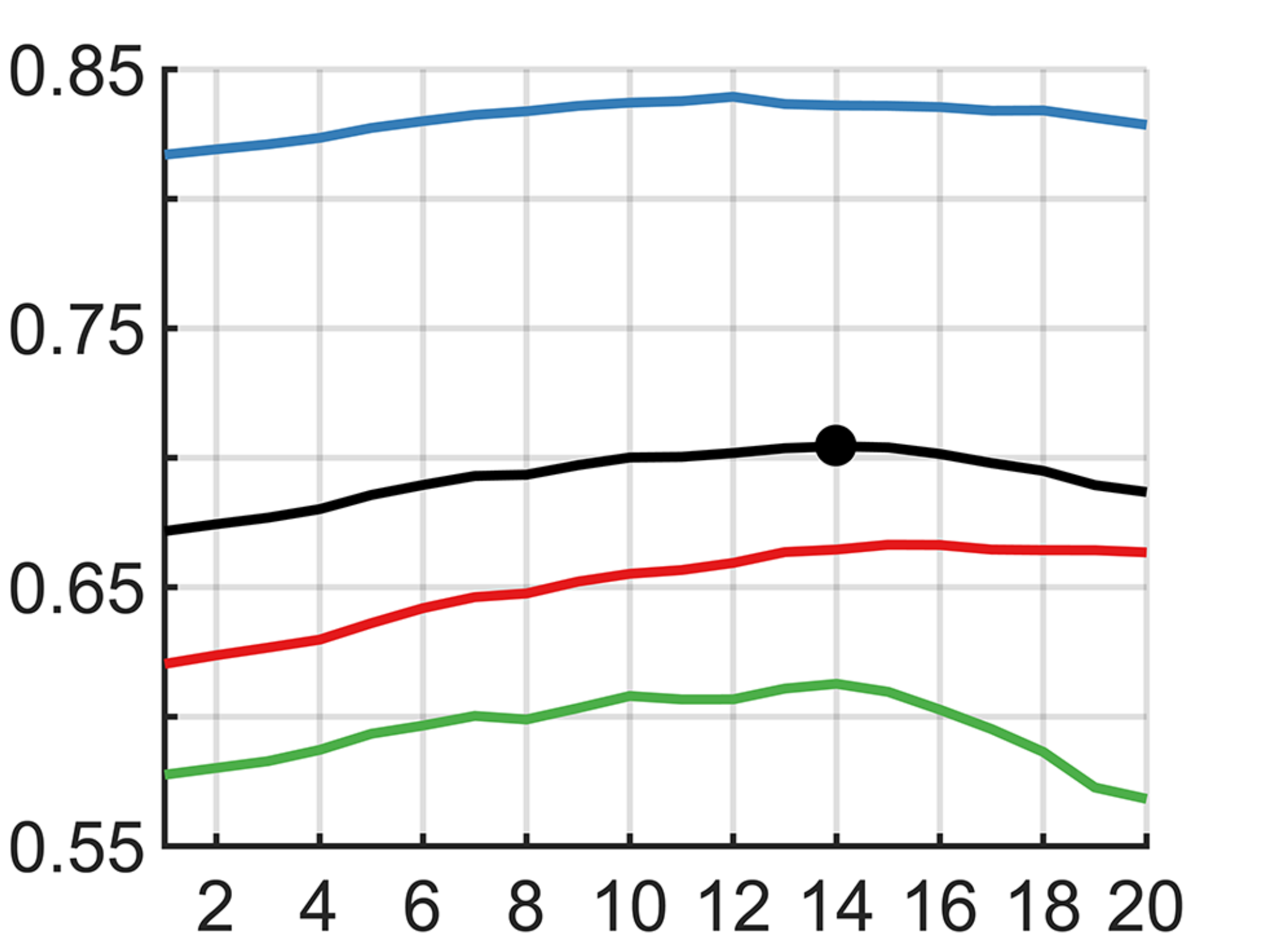}\label{fig-parameter-analysis-c}}%
	\hfill\null
	\vfill
	\hfill%
	\subfloat[$\vartheta_r$]{\includegraphics[width=0.33\linewidth]{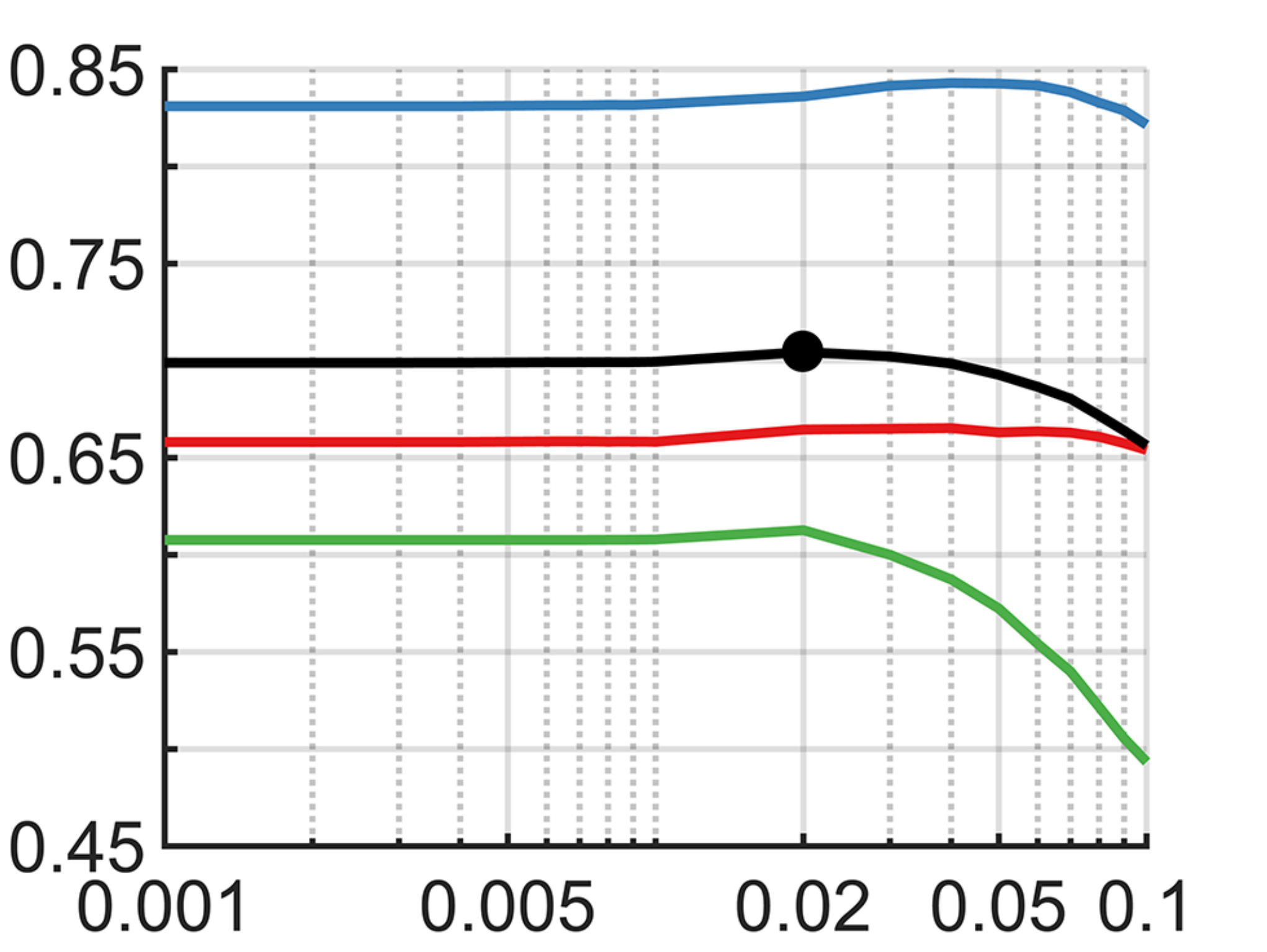}\label{fig-parameter-analysis-d}}%
	\subfloat[$\vartheta_g$]{\includegraphics[width=0.33\linewidth]{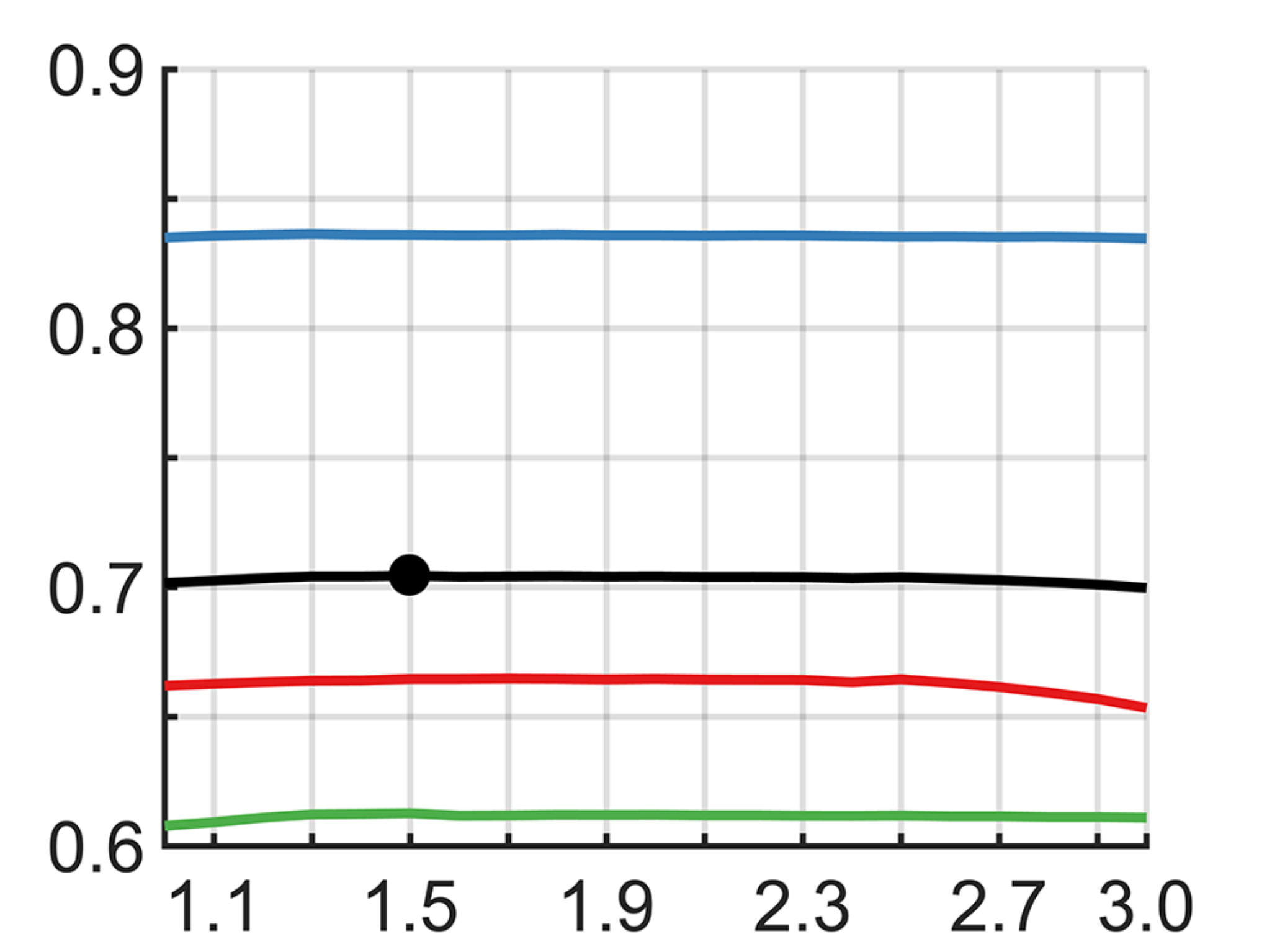}\label{fig-parameter-analysis-e}}%
	\hfill\null
\caption{Parameter analysis of the proposed model.}
\label{fig-parameter-analysis}
\end{figure}

Figure~\ref{fig-parameter-analysis} shows the influences of five parameters on the evaluation data sets. First, the proposed model is not sensitive to the parameter $\vartheta_g$, while varying it from 1.0 to 3.0 rarely changes the MaxF scores. Second, the parameters $\omega_c$, $\omega_r$, and $\vartheta_r$ have direct impacts on MaxF, especially on the ImgSal data set. Overall, each MaxF curve shows a slight upward trend as the parameter value increases, then starts to drop after the MaxF reached the summit. Compared with ASD or ECSSD, the influences of the above three parameters are more apparent on the ImgSal data set. Third, the sample step $\delta$ does not significantly impact on MaxF, all the curves do not clearly show the unimodal distributions. However, the runtime of the proposed model is directly influenced by the sample step. As the value of $\delta$ decreases, it typically results in lower speed performance.

Based on the diversity of three data sets, we further use the average MaxF metric to determine the optimal parameter values. After three MaxF curves of each parameter have been obtained, we simply average them and choose the location of the maximum as the optimal value of each parameter. The black curves in Fig.~\ref{fig-parameter-analysis} (indicated by ``Average'') exhibit the trends of five parameters. The optimal values of five parameters are reported in Table~\ref{tab-optimal-parameter}.

\begin{table}[tbp]
\renewcommand{\arraystretch}{1.2}
\caption{Optimal parameter values}
\label{tab-optimal-parameter}
\centering
	\begin{tabular}{|c||c|}
	\hline
	Parameter & Optimal value \\
	\hline
	\hline
	$\delta$ & 8 \\
	$\omega_c$ & 14 \\
	$\omega_r$ & 14 \\
	$\vartheta_r$ & 0.02 \\
	$\vartheta_g$ & 1.5 \\
	\hline
	\end{tabular}
\end{table}

\subsection{Results}
\begin{figure}[t]
\centering
	\hfill%
	\includegraphics[width=0.999\linewidth]{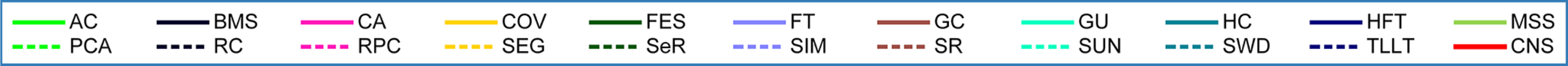}%
	\hfill\null\\
	\hfill%
	\hspace{-2em}%
	\rotatebox{90}{\captionsetup[subfigure]{font=small,labelformat=empty,margin=0pt}%
	\subfloat[{\scriptsize ASD}]{\includegraphics[width=0.2\linewidth,height=0.01cm]{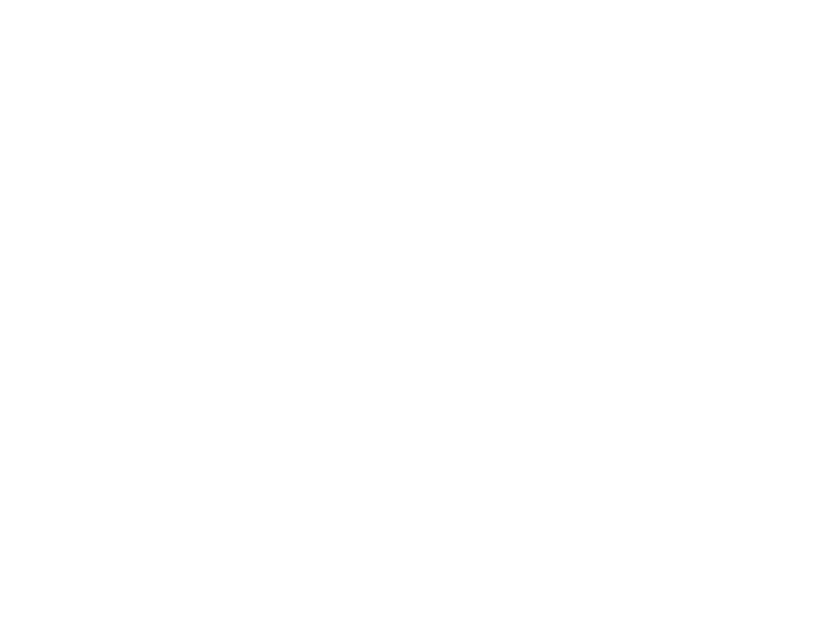}}}%
	\subfloat{\includegraphics[height=0.2\linewidth]{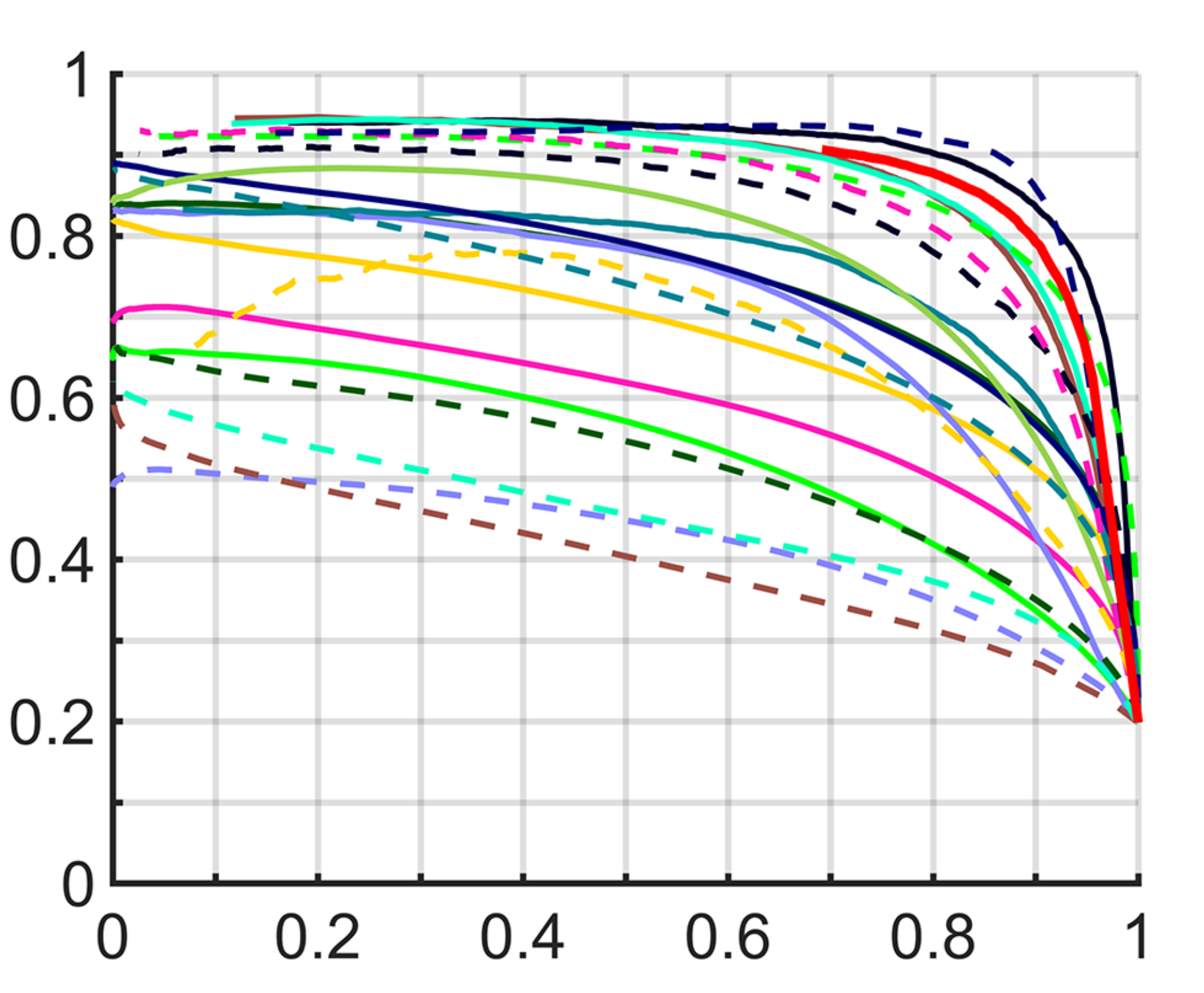}}\hfill%
	\subfloat{\includegraphics[height=0.2\linewidth]{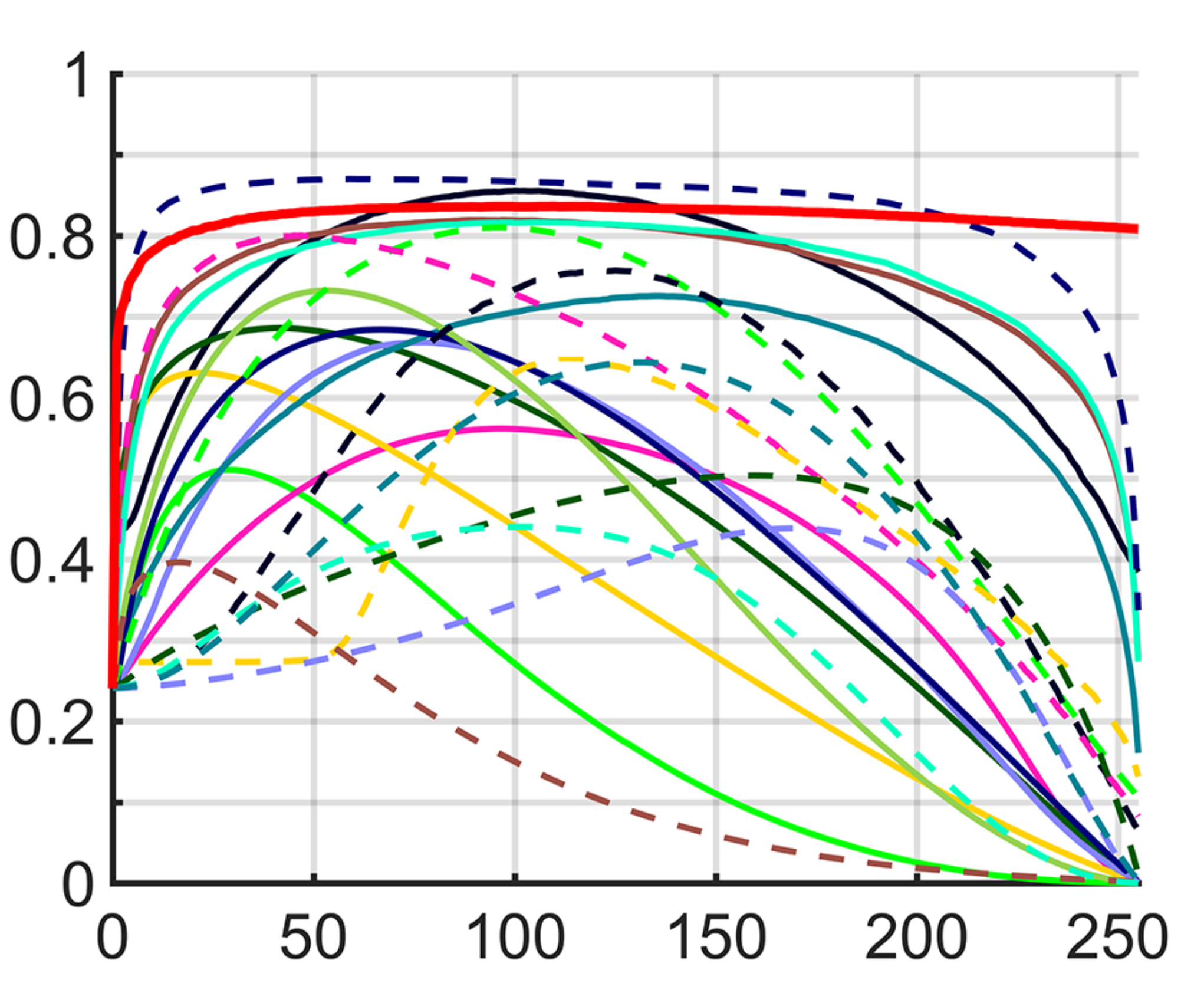}}\hfill%
	\subfloat{\includegraphics[height=0.2\linewidth]{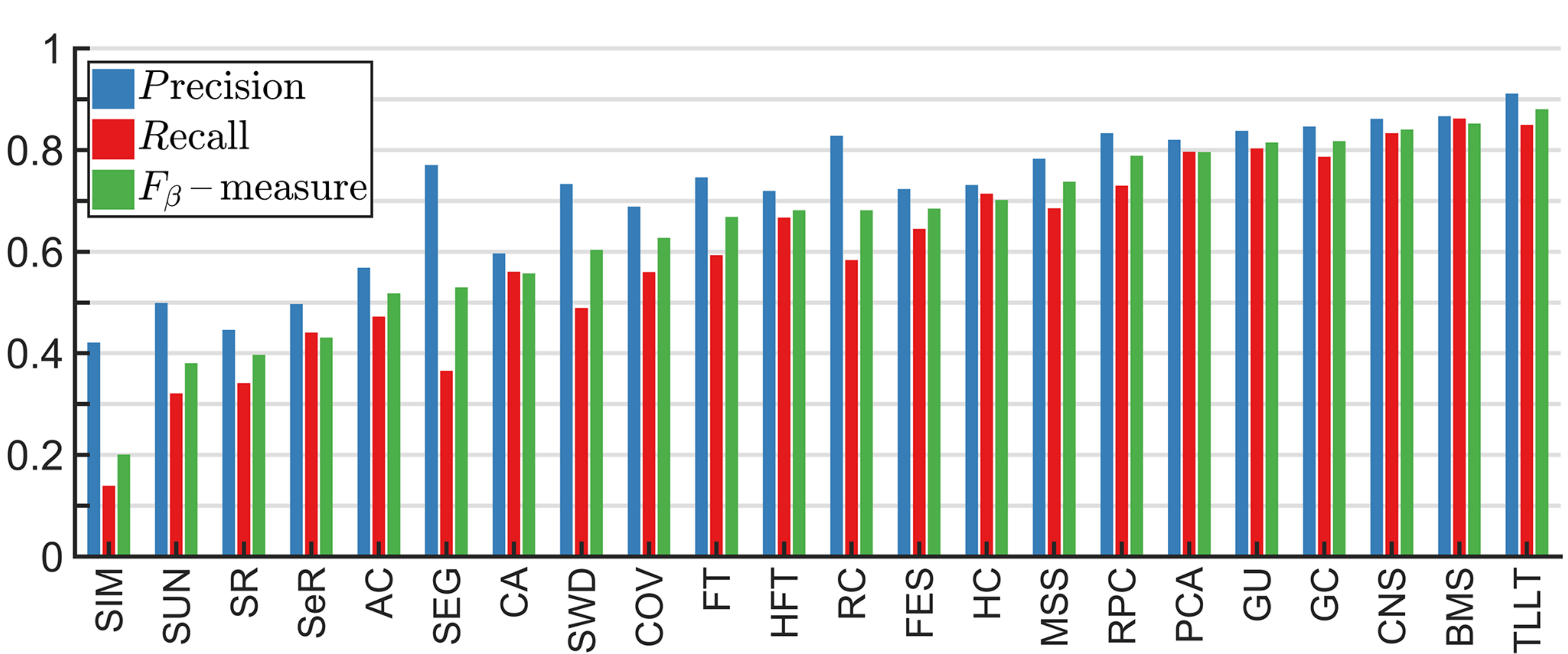}}%
	\hfill\null\\
	\hfill%
	\hspace{-2em}%
	\rotatebox{90}{\captionsetup[subfigure]{font=small,labelformat=empty,margin=0pt}%
	\subfloat[{\scriptsize ECSSD}]{\includegraphics[width=0.2\linewidth,height=0.01cm]{fig-blank}}}%
	\subfloat{\includegraphics[height=0.2\linewidth]{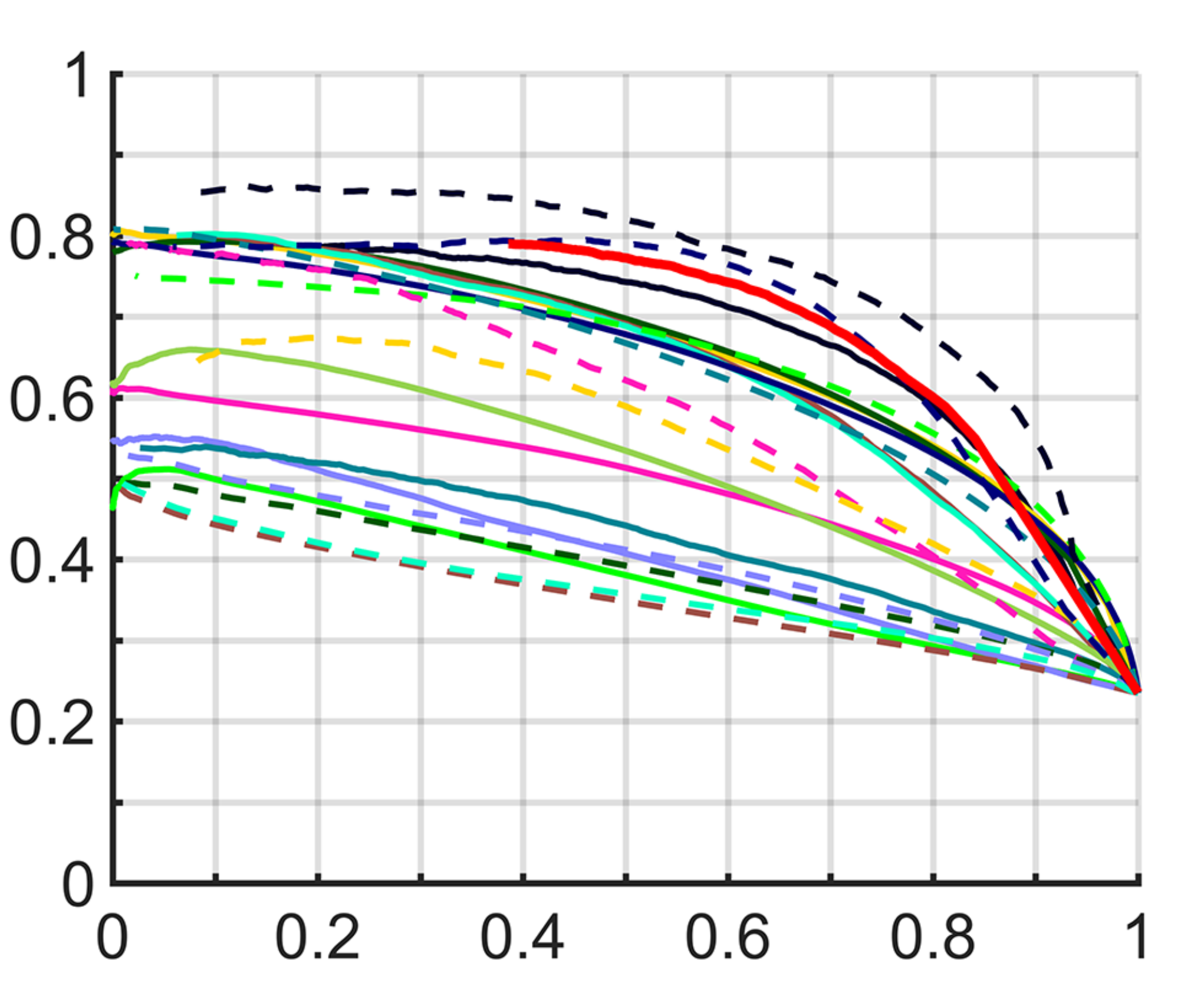}}\hfill%
	\subfloat{\includegraphics[height=0.2\linewidth]{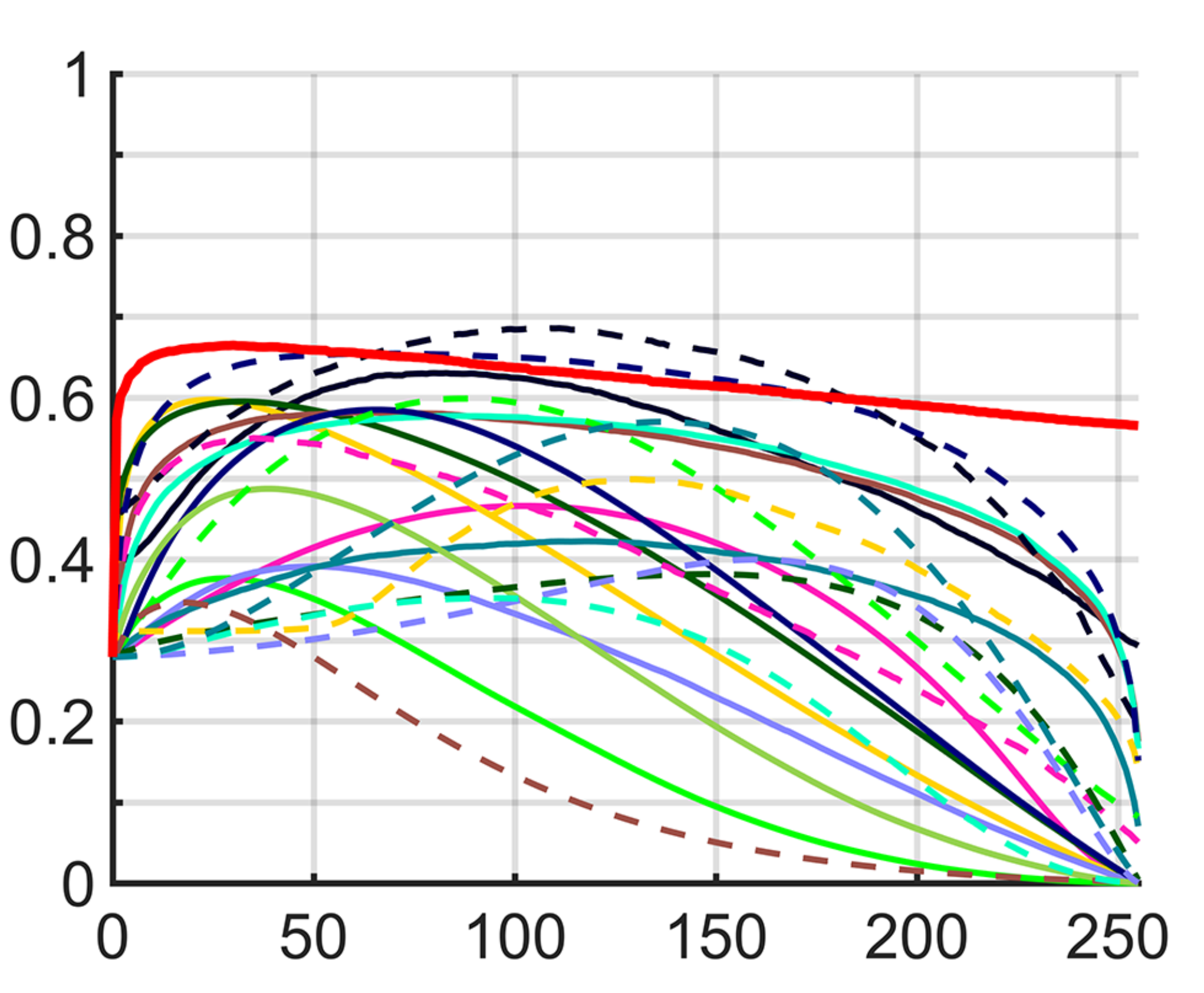}}\hfill%
	\subfloat{\includegraphics[height=0.2\linewidth]{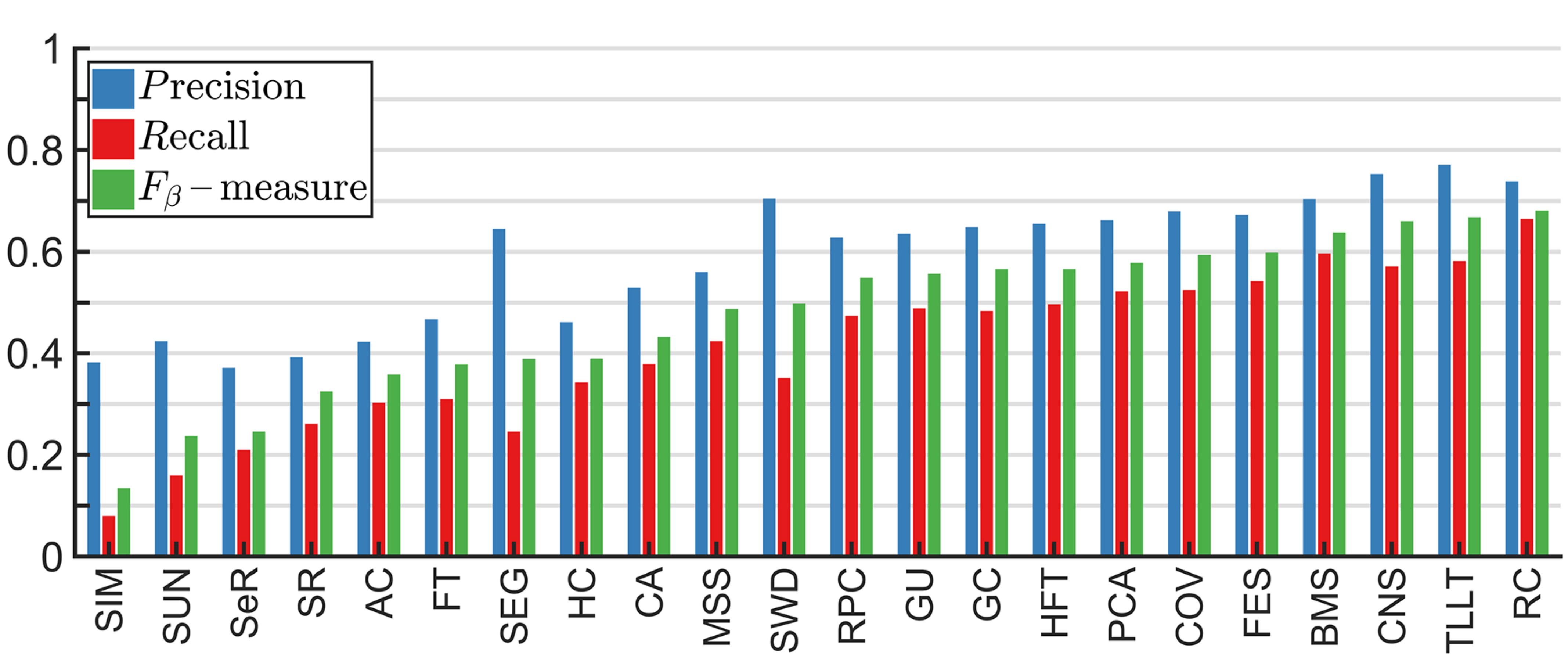}}%
	\hfill\null\\
	\hfill%
	\hspace{-2em}%
	\rotatebox{90}{\captionsetup[subfigure]{font=small,labelformat=empty,margin=0pt}%
	\subfloat[{\scriptsize ImgSal}]{\includegraphics[width=0.2\linewidth,height=0.01cm]{fig-blank}}}%
	\setcounter{subfigure}{0}
	\subfloat[]{\includegraphics[height=0.2\linewidth]{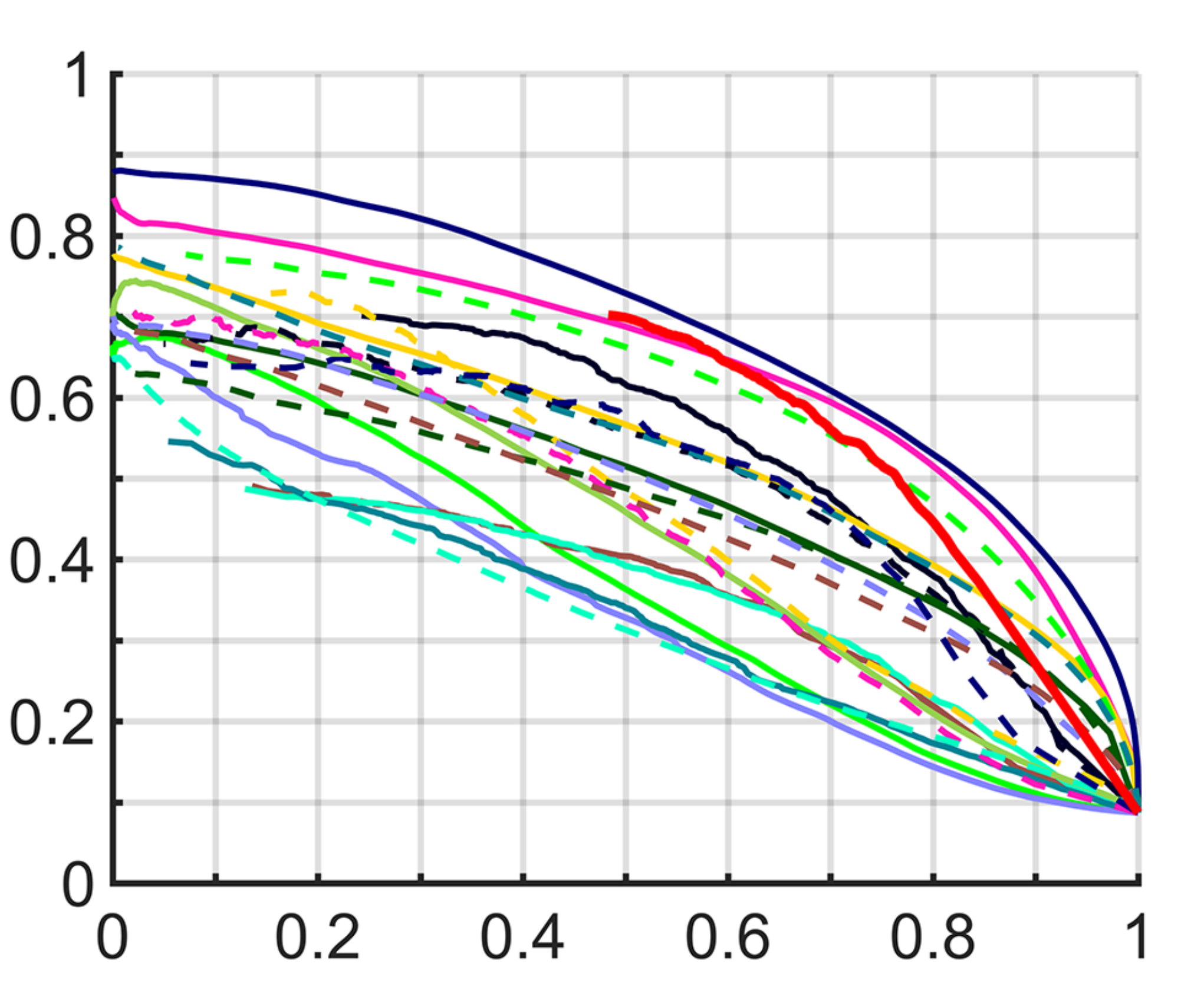}\label{fig-comparison-prf-a}}\hfill%
	\subfloat[]{\includegraphics[height=0.2\linewidth]{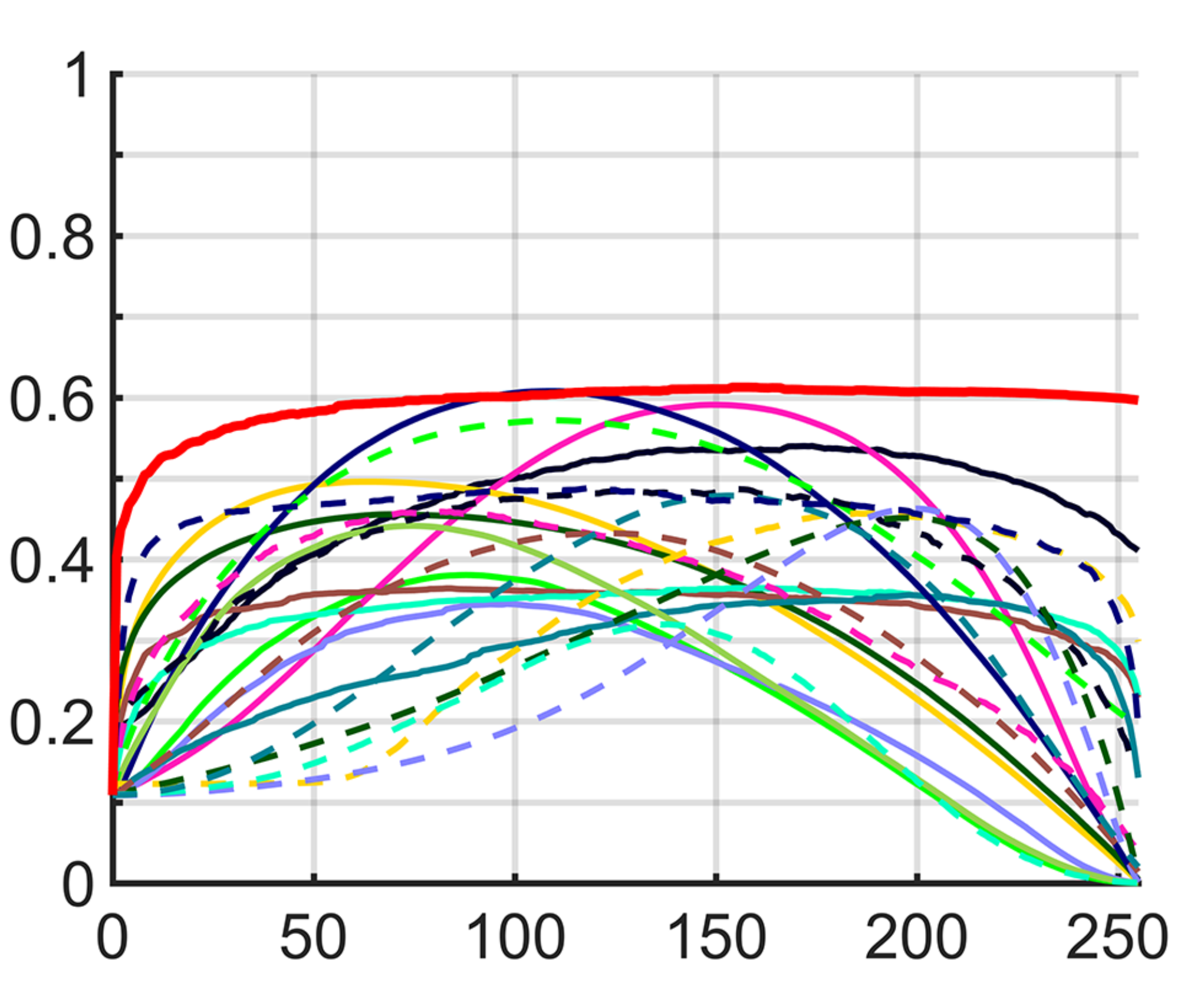}\label{fig-comparison-prf-b}}\hfill%
	\subfloat[]{\includegraphics[height=0.2\linewidth]{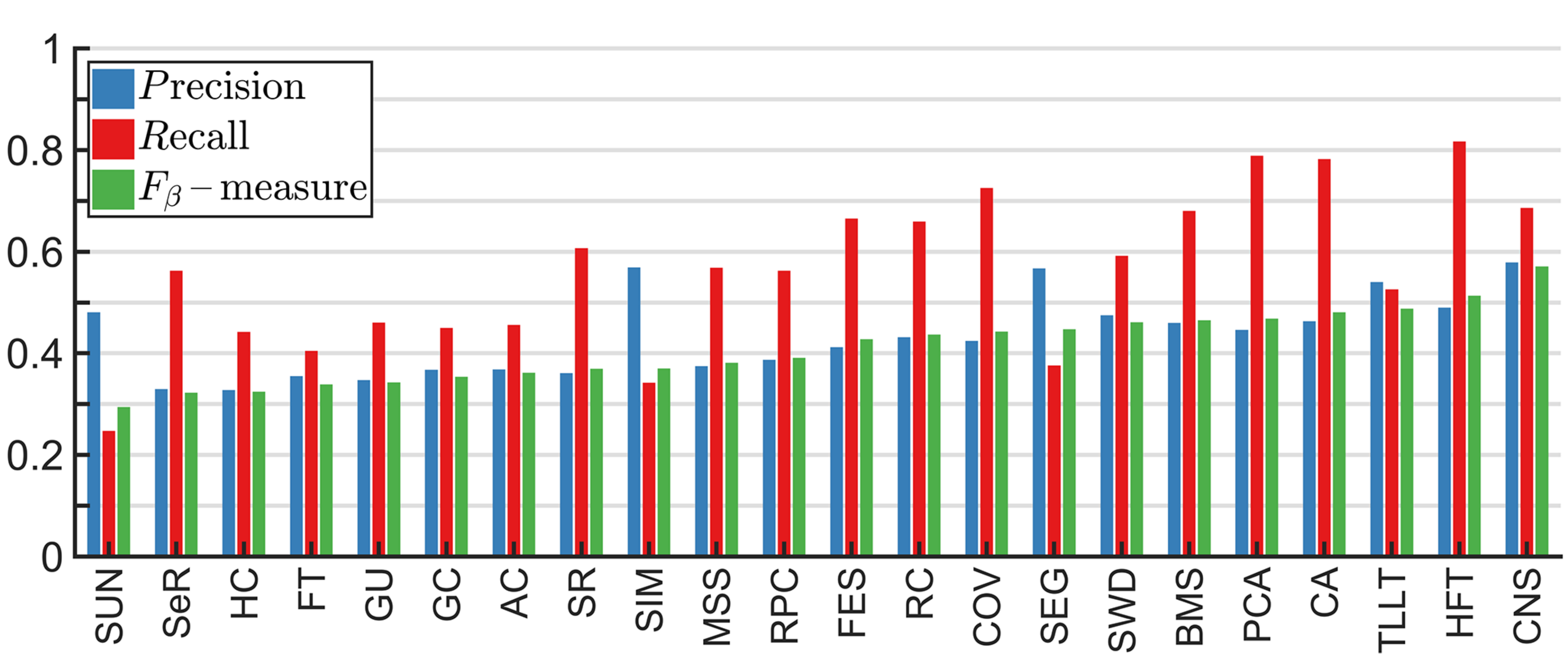}\label{fig-comparison-prf-c}}%
	\hfill\null
\caption{Performance of the proposed model compared with twenty-one saliency models on ASD~\cite{CVPR2007/Liu,CVPR2009/Achanta}, ECSSD~\cite{CVPR2013/Yan,TPAMI2016/Shi}, and ImgSal~\cite{BMVC2011/Li,TPAMI2013/Li}, respectively. (a) Precision-Recall curves. (b) $F_\beta$-measure curves. (c) Precision-Recall bars.}
\label{fig-comparison-prf}
\end{figure}

We present the statistical comparison results of the proposed model compared with twenty-one saliency models. Figures~\ref{fig-comparison-prf-a} and \ref{fig-comparison-prf-b} show the precision-recall and $F_\beta$-measure curves produced by fixed thresholding. The precision-recall bars generated by utilizing the adaptive threshold $T_a$ are presented in Fig.~\ref{fig-comparison-prf-c}. More quantitative details are given in Fig.~\ref{fig-comparison}.

Due to the intensity mapping used in the post-processing procedure, the resultant curves of our model clearly present two noticeable characteristics: one is that the recall scores span a more narrow range; the other is that the $F_\beta$-measure curves tend to be more flat after they rapidly reach the summits. Although having some disadvantages in the precision, our model has higher $F_\beta$ scores, especially on the ECSSD and ImgSal data sets. The crucial advantage of our model indeed is associated with the essential task of salient object detection, which is to solve a salient foreground segmentation problem~\cite{TIP2015/Borji-Benchmark}.

A good salient object detection model should generate accurate saliency maps with evenly highlighted foregrounds and thoroughly suppressed backgrounds. Then an easy way to extract salient objects is to binarize the saliency maps with a single fixed threshold. However, this threshold is quite difficult to determine automatically. In practice, we usually use the maximum $F_\beta$ score (i.e., MaxF) to evaluate the performance of a saliency model, and choose the location of the MaxF as the optimal segmentation threshold~\cite{TPAMI2004/Martin}. Suppose that a saliency map is the same as its ground truth mask, the $F_\beta$-measure curve would be a horizontal line. Contrarily, if the $F_\beta$-measure curve is a horizontal line, we can obtain the identical segmentation results at any threshold in $[0,255]$. Therefore, for two models having the same MaxF, we prefer to select the one which produces a more flat $F_\beta$-measure curve. This means that the segmentation results would be more stable (that is, virtually unchanged) over a wide range of thresholds.

\begin{figure}[t]
\centering
	\includegraphics[width=\linewidth]{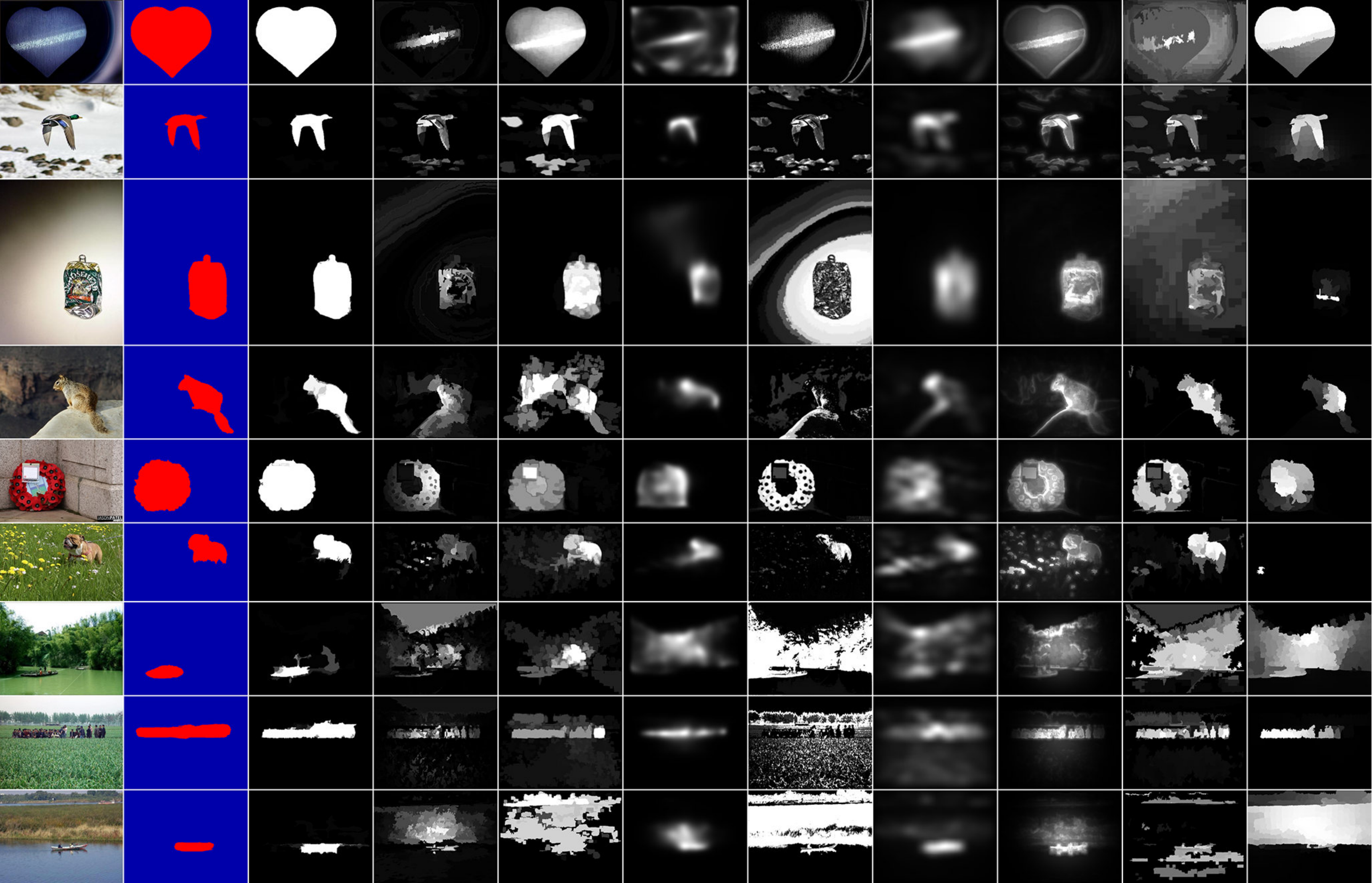}\\%
	\vspace{-10pt}
	\subfloat[]{\includegraphics[width=0.0909\linewidth,height=0.01cm]{fig-blank}}%
	\subfloat[]{\includegraphics[width=0.0909\linewidth,height=0.01cm]{fig-blank}}%
	\subfloat[]{\includegraphics[width=0.0909\linewidth,height=0.01cm]{fig-blank}}%
	\subfloat[]{\includegraphics[width=0.0909\linewidth,height=0.01cm]{fig-blank}}%
	\subfloat[]{\includegraphics[width=0.0909\linewidth,height=0.01cm]{fig-blank}}%
	\subfloat[]{\includegraphics[width=0.0909\linewidth,height=0.01cm]{fig-blank}}%
	\subfloat[]{\includegraphics[width=0.0909\linewidth,height=0.01cm]{fig-blank}}%
	\subfloat[]{\includegraphics[width=0.0909\linewidth,height=0.01cm]{fig-blank}}%
	\subfloat[]{\includegraphics[width=0.0909\linewidth,height=0.01cm]{fig-blank}}%
	\subfloat[]{\includegraphics[width=0.0909\linewidth,height=0.01cm]{fig-blank}}%
	\subfloat[]{\includegraphics[width=0.0909\linewidth,height=0.01cm]{fig-blank}}%
\caption{Visual comparison of salient object detection results. Top three rows, middle three rows, and bottom three rows are from ASD~\cite{CVPR2007/Liu,CVPR2009/Achanta}, ECSSD~\cite{CVPR2013/Yan,TPAMI2016/Shi}, and ImgSal~\cite{BMVC2011/Li,TPAMI2013/Li}, respectively. (a) Input images, and (b) ground truth masks. Saliency maps produced by using (c) the proposed CNS model, (d) RPC~\cite{PONE2014/JingLou}, (e) BMS~\cite{ICCV2013/Zhang}, (f) FES~\cite{SCIA2011/Tavakoli}, (g) GC~\cite{ICCV2013/Cheng}, (h) HFT~\cite{TPAMI2013/Li}, (i) PCA~\cite{CVPR2013/Margolin}, (j) RC~\cite{CVPR2011/Cheng}, and (k) TLLT~\cite{CVPR2015/Gong}.}
\label{fig-visual-comparison}
\end{figure}

Figure~\ref{fig-visual-comparison} shows a visual comparison of the saliency maps generated by different models. For these example images, our model generates more accurate saliency maps, which are very close to the corresponding ground truth masks. The salient regions detected by our model have uniform intensities and well-defined boundaries, which result in a simple thresholding for the subsequent salient object segmentation.

In Fig.~\ref{fig-comparison}, we report the quantitative statistics of the three evaluation metrics discussed earlier. The baseline scores, indicated by ``Average'', are simply the average of evaluation scores. With respect to AvgF, TLLT ranks the first on ASD. The proposed model outperforms all the others on ECSSD and ImgSal. Obviously, this is mainly owed to more flat $F_\beta$-measure curves in a wide range.

However, on the ASD and ECSSD data sets, our model has some disadvantages in terms of MaxF and AdpF. For the MaxF metric, TLLT performs the best on the ASD data set, and ranks the third on ECSSD. For the AdpF metric, TLLT also ranks the first on ASD, while on ECSSD the RC model performs the best. However, our model is among top three models in terms of both MaxF and AdpF on these two data sets. On the ImgSal data set, our model again outperforms all the others with large margins. Moreover, compared with ASD and ECSSD, the average performances of all the models are lower on ImgSal. It means that this data set is more challenging because the images collected in it contain salient regions of different sizes.

Finally, on average, the proposed model performs the best over all the compared models. Besides, the best two models are TLLT and BMS. The MaxF scores of nine models are lower than the average score. The top five worst models are SUN, SR, AC, SIM, and SeR. Except AC, the other four are eye fixation prediction models, which have no advantages for salient object detection because the output saliency maps are blurred and sparse. But this does not necessarily mean that the eye fixation prediction models are not suitable for detecting salient objects. For example, the BMS model is initially designed for the task of eye fixation prediction. We can see that on average it ranks the third and performs better than most of the salient object detection models evaluated in our experiments.

\begin{figure}[t]
\renewcommand{\arraystretch}{1.2}
\centering
\small
	\begin{tabular}{|c|l||ccc|ccc|ccc||ccc|}
	\hline	
	\multirow{2}{*}{\#} & \multicolumn{1}{c||}{\multirow{2}{*}{Model}}  & 
	\multicolumn{3}{c|}{ASD~\cite{CVPR2007/Liu,CVPR2009/Achanta}} & \multicolumn{3}{c|}{ECSSD~\cite{CVPR2013/Yan,TPAMI2016/Shi}} & \multicolumn{3}{c||}{ImgSal~\cite{BMVC2011/Li,TPAMI2013/Li}} & \multicolumn{3}{c|}{Average}\\
	\cline{3-5} \cline{6-8} \cline{9-11} \cline{12-14}
	& & AvgF & MaxF & AdpF & AvgF & MaxF & AdpF & AvgF & MaxF & AdpF & AvgF & MaxF & AdpF\\
	\hline
	\hline
1 & AC~\cite{ICVS2008/Achanta} & .2139  & .5107  & .5174  & .1688  & .3766  & .3575  & .2298  & .3807  & .3611  & .2042  & .4227  & .4120 \\
2 & BMS~\cite{ICCV2013/Zhang} & .7285  & {\color{green}.8555}  & {\color{green}.8515}  & .5214  & .6302  & .6370  & {\color{green}.4605}  & .5401  & .4646  & {\color{blue}.5701}  & {\color{green}.6753}  & {\color{blue}.6510} \\
3 & CA~\cite{CVPR2010/Goferman} & .4043  & .5615  & .5569  & .3403  & .4661  & .4314  & .3913  & {\color{blue}.5910}  & .4801  & .3786  & .5395  & .4895 \\
4 & COV~\cite{JOV2013/Erdem} & .3413  & .6305  & .6264  & .3347  & .5973  & .5931  & .3485  & .4960  & .4419  & .3415  & .5746  & .5538 \\
5 & FES~\cite{SCIA2011/Tavakoli} & .4484  & .6859  & .6840  & .3762  & .5951  & .5976  & .3371  & .4557  & .4268  & .3872  & .5789  & .5695 \\
6 & FT~\cite{CVPR2009/Achanta} & .4342  & .6681  & .6677  & .2419  & .3915  & .3775  & .2234  & .3451  & .3380  & .2998  & .4682  & .4611 \\
7 & GC~\cite{ICCV2013/Cheng} & {\color{blue}.7474}  & .8193  & .8169  & .5118  & .5814  & .5652  & .3381  & .3642  & .3531  & .5324  & .5883  & .5784 \\
8 & GU~\cite{ICCV2013/Cheng} & .7454  & .8164  & .8141  & .5103  & .5774  & .5558  & .3339  & .3646  & .3419  & .5299  & .5862  & .5706 \\
9 & HC~\cite{CVPR2011/Cheng} & .6113  & .7255  & .7009  & .3642  & .4224  & .3894  & .2849  & .3561  & .3238  & .4202  & .5013  & .4714 \\
10 & HFT~\cite{TPAMI2013/Li} & .4526  & .6839  & .6806  & .3739  & .5849  & .5652  & .4254  & {\color{green}.6079}  & {\color{green}.5129}  & .4173  & .6255  & .5862 \\
11 & MSS~\cite{ICIP2010/Achanta} & .4116  & .7321  & .7369  & .2543  & .4873  & .4864  & .2656  & .4415  & .3807  & .3105  & .5536  & .5347 \\
12 & PCA~\cite{CVPR2013/Margolin} & .5884  & .8101  & .7953  & .4252  & .5987  & .5778  & .4415  & .5718  & .4679  & .4850  & .6602  & .6137 \\
13 & RC~\cite{CVPR2011/Cheng} & .5192  & .7570  & .6809  & {\color{blue}.5766}  & {\color{red}.6860}  & {\color{red}.6801}  & .4048  & .4871  & .4365  & .5002  & .6434  & .5992 \\
14 & RPC~\cite{PONE2014/JingLou} & .5762  & .8002  & .7880  & .3757  & .5499  & .5479  & .3400  & .4598  & .3907  & .4306  & .6033  & .5755 \\
15 & SEG~\cite{ECCV2010/Rahtu} & .4305  & .6485  & .5288  & .3840  & .4990  & .3883  & .3096  & .4569  & .4470  & .3747  & .5348  & .4547 \\
16 & SeR~\cite{JOV2009/Seo} & .3975  & .5037  & .4300  & .3179  & .3818  & .2452  & .2855  & .4513  & .3216  & .3336  & .4456  & .3323 \\
17 & SIM~\cite{CVPR2011/Murray} & .3162  & .4384  & .2002  & .3080  & .3998  & .1342  & .2497  & .4626  & .3698  & .2913  & .4336  & .2347 \\
18 & SR~\cite{CVPR2007/Hou} & .1435  & .3964  & .3964  & .1275  & .3469  & .3246  & .3006  & .4324  & .3687  & .1905  & .3919  & .3632 \\
19 & SUN~\cite{JOV2008/Zhang} & .2916  & .4402  & .3803  & .2442  & .3522  & .2365  & .1764  & .3198  & .2937  & .2374  & .3708  & .3035 \\
20 & SWD~\cite{CVPR2011/Duan} & .4399  & .6434  & .6033  & .4074  & .5700  & .4971  & .3016  & .4787  & .4605  & .3830  & .5640  & .5203 \\
21 & TLLT~\cite{CVPR2015/Gong} & {\color{red}.8270}  & {\color{red}.8699}  & {\color{red}.8799}  & {\color{green}.5832}  & {\color{blue}.6543}  & {\color{green}.6671}  & {\color{blue}.4512}  & .4878  & {\color{blue}.4874}  & {\color{green}.6205}  & {\color{blue}.6707}  & {\color{green}.6781} \\
	\hline
22 & CNS & {\color{green}.8204}  & {\color{blue}.8361}  & {\color{blue}.8398}  & {\color{red}.6191}  & {\color{green}.6645}  & {\color{blue}.6593}  & {\color{red}.5902}  & {\color{red}.6127}  & {\color{red}.5702}  & {\color{red}.6765}  & {\color{red}.7044}  & {\color{red}.6898} \\	
	\hline
	\hline
	\multicolumn{2}{|c||}{Average} & .4950  & .6742  & .6444  & .3803  & .5188  & .4779  & .3404  & .4620  & .4109  & .4052  & .5517  & .5111 \\
	\hline
	\end{tabular}
\caption{Statistics of average $F_\beta$ (AvgF), maximum $F_\beta$ (MaxF), and $F_\beta$ using adaptive threshold (AdpF) on three evaluation data sets. The top three scores under each metric are highlighted in red, green, and blue, respectively. See the text for details.}
\label{fig-comparison}
\end{figure}

\subsection{Discussions}
\begin{figure}[t]
	\hspace{-2em}\hfill%
	\rotatebox{90}{\captionsetup[subfigure]{font=small,labelformat=empty,margin=0pt}%
	\subfloat[{\scriptsize CNS}]{\includegraphics[width=0.12\linewidth,height=0.01cm]{fig-blank}}%
	\subfloat[{\scriptsize CN}]{\includegraphics[width=0.12\linewidth,height=0.01cm]{fig-blank}}%
	\subfloat[{\scriptsize GT}]{\includegraphics[width=0.12\linewidth,height=0.01cm]{fig-blank}}%
	\subfloat[{\scriptsize Input}]{\includegraphics[width=0.12\linewidth,height=0.01cm]{fig-blank}}}%
	\subfloat{\includegraphics[width=0.96\linewidth]{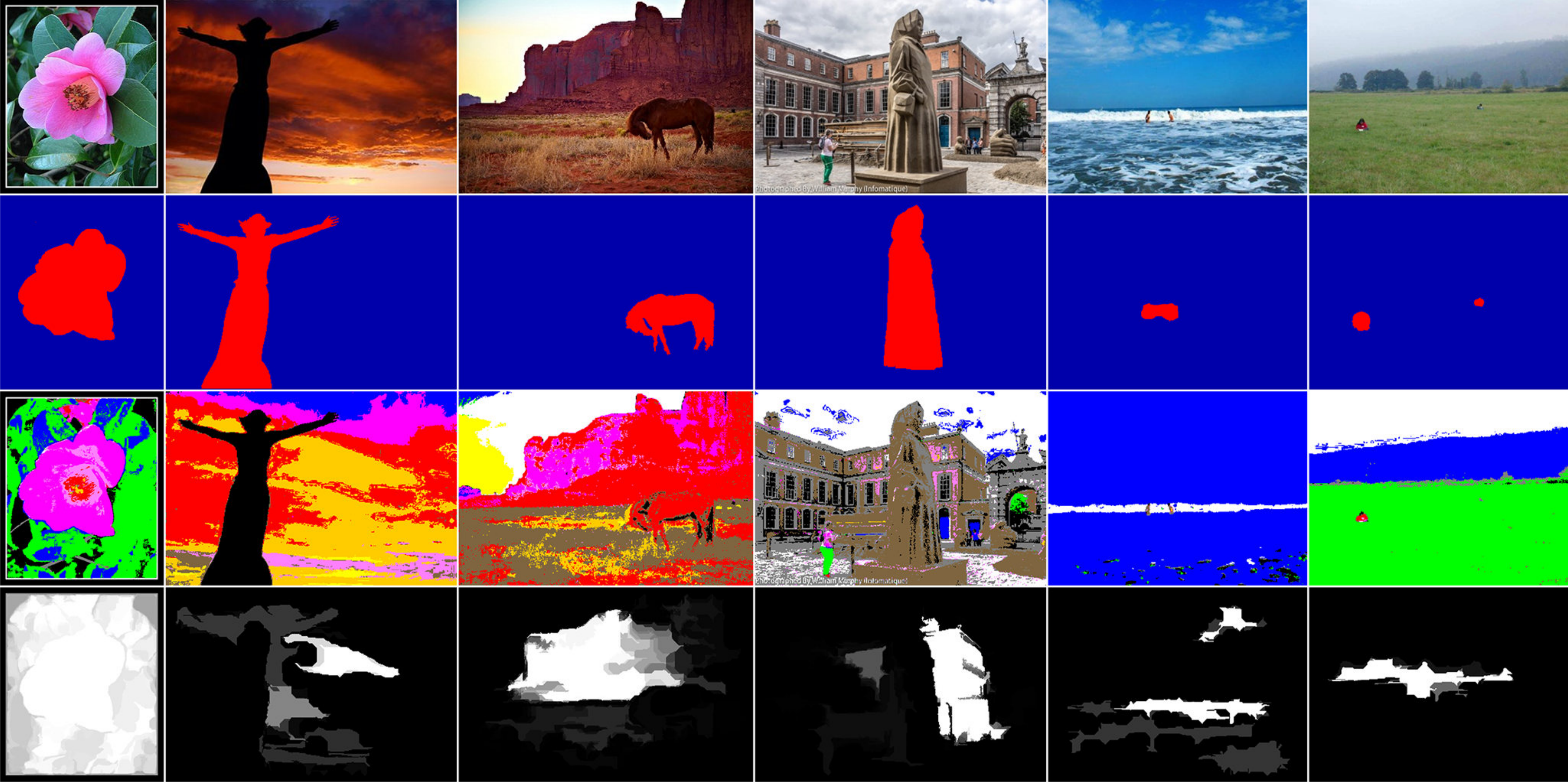}}%
	\hfill\null\vspace{-10pt}\\
	\subfloat{\includegraphics[width=0.03\linewidth,height=0.01cm]{fig-blank}\setcounter{subfigure}{0}}%
	\subfloat[]{\includegraphics[width=0.1\linewidth,height=0.01cm]{fig-blank}\label{fig-hard-case-a}}%
	\subfloat[]{\includegraphics[width=0.18\linewidth,height=0.01cm]{fig-blank}\label{fig-hard-case-b}}%
	\subfloat[]{\includegraphics[width=0.18\linewidth,height=0.01cm]{fig-blank}\label{fig-hard-case-c}}%
	\subfloat[]{\includegraphics[width=0.18\linewidth,height=0.01cm]{fig-blank}\label{fig-hard-case-d}}%
	\subfloat[]{\includegraphics[width=0.16\linewidth,height=0.01cm]{fig-blank}\label{fig-hard-case-e}}%
	\subfloat[]{\includegraphics[width=0.16\linewidth,height=0.01cm]{fig-blank}\label{fig-hard-case-f}}%
\caption{Hard image cases. Left two columns, middle two columns, and right two columns are from ASD~\cite{CVPR2007/Liu,CVPR2009/Achanta}, ECSSD~\cite{CVPR2013/Yan,TPAMI2016/Shi}, and ImgSal~\cite{BMVC2011/Li,TPAMI2013/Li}, respectively. Input: input images. GT: ground truth masks. CN: color name images. CNS: salient object detection results of the proposed model.}
\label{fig-hard-case}
\end{figure}

Although the proposed model performs well on the evaluation data sets, it does fail in some cases. These failures are mainly caused by three visual attributes implicitly used for identifying salient objects: location, color, and size. Figure~\ref{fig-hard-case} shows several hard image cases collected from the evaluation data sets. The third row are the color name images annotated by using the RGB colors given in Table~\ref{table:color-names}.

\begin{itemize}
\item \textit{Location:} The key idea of BMS is the Gestalt principle based surroundedness, thus the salient regions connected to the image borders would be masked out in the generation of attention maps, as shown in Fig.~\ref{fig-hard-case-b}.

\item \textit{Color:} The proposed model originates from BMS, and exploits eleven color name channels for figure-ground segregation. Sometimes, the foreground objects do not directly touch the image borders, but may have very similar colors to the backgrounds. For example, in the third rows of Figs.~\ref{fig-hard-case-c} and \ref{fig-hard-case-d}, the RGB colors of the manually labeled salient objects (the horse and the statue) and some background regions (e.g., the valley and the plinth) are almost the same. While salient objects and image borders are connected by background elements, the salient objects are always removed in the generation of attention maps. Moreover, the color statistics based global contrast is introduced in the proposed model. The color similarities between foreground regions and background elements impact the ability of literally popping out salient objects (cf. Figs.~\ref{fig-hard-case-c} and \ref{fig-hard-case-d}).

\item \textit{Size:} In the proposed model, some morphological operations, including closing and reconstruction, are used to compute saliency maps. The influences of the parameters $\omega_c$ and $\omega_r$ have been presented in Fig.~\ref{fig-parameter-analysis}. These parameters have a substantial impact on the outputs of our model, especially on the ImgSal data set. As Figs.~\ref{fig-hard-case-e} and \ref{fig-hard-case-f} show, the manually labeled regions are eroded because the morphological structures are larger than the sizes of these regions.

\item Another hard case is caused by the thin artificial borders around some test images, as illustrated in Fig.~\ref{fig-hard-case-a}. When doing the clear-border operation on boolean maps, the proposed model will regard the inner area as a whole region which is surrounded by an enclosed boundary, and does not set any of the foreground pixels to 0. Such a processing mechanism leaves unchanged background elements inside the artificial borders, and results in the failure of figure-ground segregation.
\end{itemize}

Clearly, the proposed model focuses on the bottom-up image processing technique, and only exploits some low-level image features. Therefore, it fails to highlight the regions that have similar colors to their surroundings. One way to tackle this issue is to invoke more complex visual features. Second, under the definition of surroundedness, the regions connected to the image borders are not enclosed by any complete outer contour. This results in the absence of object-level information in the attention map computation. The above problem can be solved by invoking some background priors and top-down cues. Finally, the proposed model works well for detecting large salient objects, but is not suitable for small ones. It would be interesting to adopt a multi-scale strategy or automatically seek the optimal scale for the detection of different sizes of salient objects.

\section{Conclusions}
\label{sec:conclusions}
Throughout this paper, we present a salient object detection model based on color name space. Considering the outstanding contribution of color contrast for saliency detection, a unified framework is constructed to overcome the limitation of the boolean map based saliency. By exploiting several visual features with respect to linguistic color names, we suggest that the model of fusing color attributes provides superior performance over that only based on the surroundedness cue. Moreover, we propose an improved post-processing procedure to uniformly smooth and highlight salient regions, so that the detected salient objects have high and constant intensity levels for the convenience of object segmentation. Experimental results indicate the performance improvement of the proposed model on three evaluation data sets.

With regard to future work, first, we intend to invoke a background measure to handle the salient objects that heavily connected to the image borders. Second, it would be interesting to incorporate more visual features and top-down cues to solve the problem of color confusion between foreground regions and backgrounds. Third, for the morphological structures used in the proposed model, only a fixed value is chosen as the optimal kernel radius, which results in the loss of small salient objects. We have noted that an adaptive radius can effectively address this issue. How to automatically determine the radius size is left to future investigation. Finally, the current version of our MATLAB code is implemented for the purpose of academic research. We further plan to optimize the code to improve the speed performance of the proposed model.

\begin{acknowledgements}
J. Lou is supported by the Changzhou Key Laboratory of Industrial Internet and Data Intelligence (No.~CM20183002) and the QingLan Project of Jiangsu Province (2018). The work of L. Chen, F. Xu, W. Zhu, and M. Ren is supported by the National Natural Science Foundation of China (Nos.~61231014 and 61727802). H. Wang is supported by the National Defense Pre-research Foundation of China (No.~9140A01060115BQ02002) and the National Natural Science Foundation of China (No.~61703209). Q. Xia is supported by the National Natural Science Foundation of China (No.~61403202) and the China Postdoctoral Science Foundation (No.~2014M561654). The authors thank Andong Wang and Haiyang Zhang for helpful discussions regarding this manuscript.
\end{acknowledgements}

\bibliographystyle{spmpsci}      
\bibliography{Manuscript}   

\end{document}